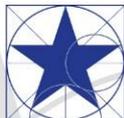

## PSL
UNIVERSITÉ PARIS

## THÈSE DE DOCTORAT
### DE L'UNIVERSITÉ PSL
Préparée à l'Université Paris-Dauphine

# Learning off-road maneuver plans for autonomous vehicles

Soutenue par
**Kevin Osanlou**
Le 27/05/2021

Ecole doctorale n° ED 543
**Ecole doctorale SDOSE**

Spécialité
**Informatique**

Composition du jury :

| | |
|---|---|
| Aurélie, BEYNIER<br>HDR, Sorbonne Université | *Rapporteur* |
| Ludovic, DENOYER<br>Pr, Sorbonne Université<br>Détaché chez Facebook France | *Rapporteur* |
| Rachid, ALAMI<br>Directeur de recherche, LAAS - CNRS | *Président du jury*<br>*Examinateur* |
| François, CHARPILLET<br>Directeur de recherche, LORIA - INRIA | *Examinateur* |
| Christophe, GUETTIER<br>Dr, Safran | *Examinateur* |
| Jeremy, FRANK<br>Research Lead, P&S Group - NASA ARC | *Examinateur* |
| Éric, JACOPIN<br>HDR, CREC Saint-Cyr | *Directeur de thèse* |
| Tristan, CAZENAVE<br>Pr, Université Paris-Dauphine | *Directeur de thèse* |

**Dauphine** | PSL
UNIVERSITÉ PARIS

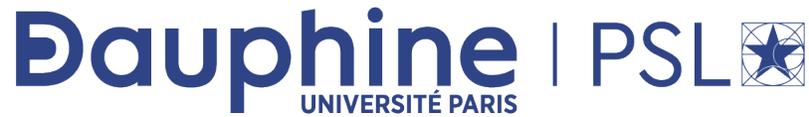

# LEARNING OFF-ROAD MANEUVER PLANS FOR AUTONOMOUS VEHICLES

by

Kevin Osanlou

A dissertation submitted to Paris-Dauphine University

in conformity with the requirements for the degree of

Doctor of Philosophy in Computer Science

In the field of Artificial Intelligence for Decision-Making

Supervisor: Tristan Cazenave

Co-supervisors: Eric Jacopin, Christophe Guettier

Paris, France

March, 2021

# Acknowledgments

This thesis is part of a collaboration between company Safran Electronics & Defense, laboratory LAMSADE from Paris-Dauphine University and laboratory CREC from Écoles de Saint-Cyr Coëtquidan. I would like to thank every entity involved in this collaboration. I also thank the Association Nationale de la Recherche et de la Technologie (ANRT), which partially funded this thesis along with Safran Electronics & Defense. I would like to deeply thank my thesis supervisor, professor Tristan Cazenave, for the opportunity he gave me to carry out this thesis research under his supervision and the precious insights he gave me. I would like to thank Eric Jacopin who co-supervised this thesis and provided precious feedback and advice. I would like to thank Christophe Guettier who gave me various ideas to explore which led to several contributions in the thesis. I would like to thank Andrei Bursuc who offered me helpful insight on various occasions and helped me write publications. I would like to thank Jeremy Frank for the internship opportunity he offered me at NASA, his insights and feedback. I also thank all members of the committee of this thesis for the interest they shared in my works: Ludovic Denoyer, Aurélie Beynier, Rachid Alami and François Charpillet.

# Résumé Français

A French summary is provided here. Please move to Table of Contents for English.

Dans cette thèse, nous étudions les bénéfices que l'apprentissage automatique peut apporter à des tâches de planification automatique, un secteur de plus en plus important dans un monde où l'autonomie des systèmes ne cesse de croître. Dans ce résumé, nous présentons deux contributions majeures de cette thèse. Dans un premier temps, nous considérons un scénario dans lequel un Véhicule Autonome (VA), envoyé sur un terrain dans lequel des inondations ont eu lieu, est chargé d'acheminer des vivres aux territoires les plus touchés. Nous proposons une formalisation du problème en tant que problème de planification d'itinéraire. Dans un second temps, nous considérons un scénario dans lequel le VA est dans un convoi militaire, et doit effectuer des manœuvres synchronisées avec un Véhicule Partenaire (VP). Pour cela, le VA doit calculer une stratégie réactive d'exécution de manœuvres qui prend en compte l'incertitude liée aux manœuvres du VP. Nous formalisons le problème en planification dynamique de tâches. Dans le corpus principal (en anglais), nous proposons également des contributions supplémentaires avec l'utilisation de recherche locale pour la planification d'itinéraire, ainsi que de l'optimisation multi-critère.

## Planification d'Itinéraire sous Contraintes

### Contexte et Formalisation du Problème

Dans cette section, nous considérons une problématique dans lequel un VA se trouve dans un terrain victime d'inondations. Le VA se situe à une position initiale sur le terrain, et une mission lui est assignée. Pour la satisfaire, il doit acheminer des vivres à certaines positions en manque, puis se rendre à une position finale pour terminer la mission. Le terrain étant endommagé par l'inondation, le VA emprunte des chemins "off-road". Nous représentons le terrain par un graphe connecté non orienté $\mathcal{G} = (\mathcal{V}, \mathcal{E})$: les nœuds du graphe représentent les diverses positions sur le terrain, et les arrêtes les chemins off-road qui lient ces positions et que le VA peut emprunter pour se déplacer. Le poids d'une arrête représente un coût que le VA doit payer pour traverser l'arrête: nous considérons ici que celui-ci est la distance du chemin off-road correspondant à l'arrête. La Figure 1 illustre un tel graphe. Nous considérons des scénarios VA représentés par des graphes non orientés dans les expérimentations de cette thèse, néanmoins les méthodes de résolution proposées pour ce problème sont applicables également à des graphes orientés, seul l'heuristique utilisée (le type de graph neural network) doit être modifiée pour prendre en compte les directions des arrêtes.

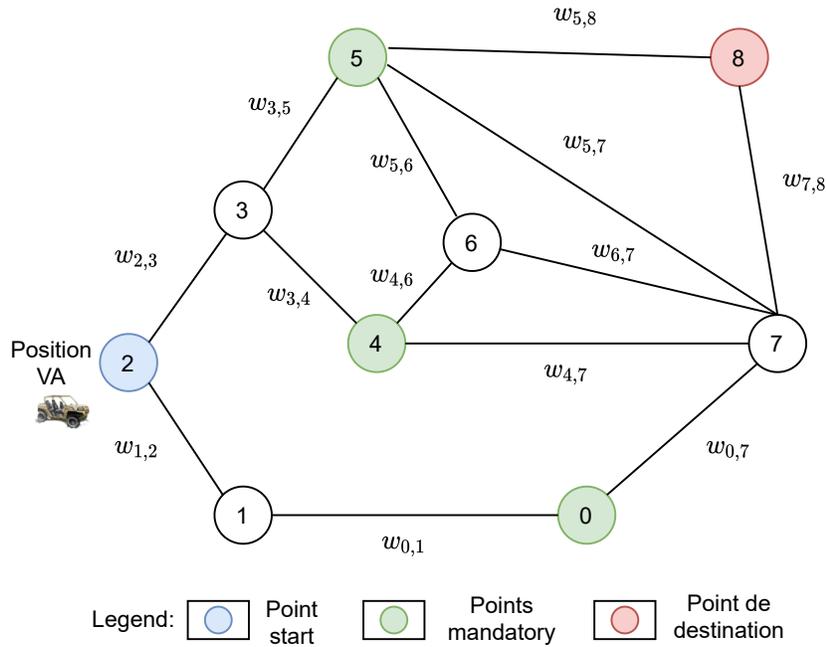

**Figure 1: Représentation graphique du scénario.** Le terrain est représenté par un graphe $\mathcal{G} = (\mathcal{V}, \mathcal{E})$. Chaque arrête $e_{i,j}$ a un poids $w_{i,j}$ qui correspond au coût à payer pour traverser l'arrête. Le VA est situé sur le point start en bleu, doit passer par les points obligatoires en vert pour acheminer des vivres, puis se rendre au point destination en rouge pour terminer la mission.

Une mission pour le VA est caractérisée par 3 éléments: un nœud de départ où se situe le VA; des nœuds *mandatory* (*i.e.* obligatoires) que le VA doit visiter au moins une fois; et un nœud de destination où le VA doit se rendre pour terminer la mission après avoir visité tous les nœuds mandatory. Aucun ordre de visite n'est imposé sur les nœuds mandatory. Le VA doit satisfaire les objectifs de cette mission tout en minimisant la distance totale parcourue. Nous montrons par la suite que le problème est NP-difficile: il s'agit d'un variant du Travelling Salesman Problem (TSP) avec des points de départ et arrivée différents. Mathématiquement, une mission correspond à une instance $I$ de ce type de problème défini tel que:

$$I = (start, dest, M)$$

où:

- $start \in \mathcal{V}$ correspond au nœud de départ de la mission dans le graphe.

- $dest \in \mathcal{V}$ correspond au nœud de destination de la mission dans le graphe.

- $M \subset \mathcal{V}$ correspond à un ensemble de nœuds mandatory distincts qui doivent être visités au moins une fois dans la mission, sans ordre de visite spécifique.

Un chemin solution satisfait les objectifs de la mission, *i.e.* débute du nœud de départ, visite tous les nœuds obligatoires au moins une fois, et se termine au point de destination. Un chemin solution $\pi$ est donc une liste ordonnée de nœuds $\pi = \{start, v_1, ..., v_q, dest\}$, où les $v_i$ sont des nœuds du graphe, tel que $M \subset \pi$. Le coût (distance) de $\pi$ est la somme de la distance de chaque arrête incluse: $d_{start,v_1} + d_{v_1,v_2} + ... + d_{v_q,dest}$ où $d_{v,v'}$ est le poids de l'arrête qui relie $v$ et $v'$. Un chemin solution $\pi^*$ est optimal s'il n'existe aucun autre chemin solution avec un coût moins élevé.

De plus, deux phases existent: une phase **hors-ligne**, pendant laquelle le VA n'est pas sur un terrain. Pendant cette phase, le VA doit être 'préparé' à un terrain particulier. Le graphe du terrain en question étant connu d'avance, le système de planification du VA doit être adapté à celui-ci pour y être plus performant. Cette phase est supposée aussi longue que nécessaire. Une autre phase, **en-ligne**, correspond aux situations pour lesquelles le VA est présent sur ce terrain. Lors de cette phase, le VA peut recevoir à tout moment une mission différente, et doit trouver un chemin solution optimal à la mission aussi vite que possible afin de l'emprunter et accomplir la mission.

Par la suite, le mot 'instance' fera référence à une instance du problème (c'est à dire une mission particulière). En posant $|V| = n$, le nombre d'instances existant pour le graphe $\mathcal{G}$ est donné par:

$$C_n = 2^{n-2}n(n-1) \tag{1}$$

où $n(n-1)$ correspond au choix du nœud de départ puis du nœud de destination; $2^{n-2}$ au nombre de combinaisons possibles de nœuds obligatoires parmi les $n-2$ nœuds restants. Le nombre d'instances pour un graphe avec $|\mathcal{V}| = n = 20$ est égal à $99\,614\,720$. Par conséquent, calculer un chemin solution optimal pour toutes les instances existantes dans la phase hors-ligne afin de pouvoir utiliser ces solutions en-ligne devient trop difficile même avec un grand temps de préparation disponible. Nous proposons à la place d'entraîner un modèle d'apprentissage automatique dans la phase hors-ligne afin d'apprendre la topologie des instances du graphe, et d'utiliser ce modèle en-ligne pour accélérer la résolution d'instances.

## Méthode de Résolution

Résoudre $I = (start, dest, M)$ dans le graphe $\mathcal{G}$ de manière optimale revient à trouver un chemin $\pi^*$, *i.e.* une séquence de nœuds qui débute à $start$, se termine à $dest$, minimise le poids total des arrêtes incluses dans $\pi^*$ et satisfait les contraintes de visites obligatoires $M \subset \pi^*$. L'ordre de visite n'est pas imposé pour $M$. Si c'était le cas, le problème pourrait être résolu en temps polynomial en calculant de manière consécutive les paires de plus courts chemins depuis le nœud start jusqu'au $1^{er}$ nœud obligatoire, du $1^{er}$ nœud obligatoire jusqu'au $2^{eme}$ et ainsi de suite jusqu'au nœud de destination. Ici, trouver un chemin solution optimal revient à trouver un ordre de visite optimal des nœuds obligatoires. Pour ce faire, dans un premier temps, nous calculons les plus courts chemins entre toutes les paires de nœuds $\langle v_i, v_j \rangle$ dans $\mathcal{G}$ avec l'algorithme de Dijkstra, ainsi que le coût associé à chaque plus court chemin. Le chemin solution associé à un ordre de visite des nœuds obligatoires $o = (m_1, m_2, m_3, ..., m_{q-1}, m_q) \mid m_i \in M \;\forall i = 1, 2, ..., q$ peut être reconstruit en concaténant les plus courts chemins de $start$ à $m_1$, de $m_1$ à $m_2$, de $m_2$ à $m_3$, ... , de $m_{q-1}$ à $m_q$, et de $m_q$ à $dest$. Le coût total est la somme du coût de chaque plus court chemin utilisé pour la reconstruction.

Afin de déterminer un ordre de visite optimal des nœuds obligatoires, nous définissons un arbre de recherche. Pour une instance $I = (start, dest, M)$, la racine de cet arbre est le nœud $start$, chaque feuille de l'arbre est le nœud $dest$, et chaque nœud intermédiaire de l'arbre (à ne pas confondre avec les nœuds du graphe) est un nœud obligatoire dans $M$, de telle sorte que chaque chemin possible de la racine de l'arbre jusqu'à n'importe quelle feuille de l'arbre représente un ordre de visite unique des nœuds obligatoires dans $M$. Le coût de transition d'un nœud parent de l'arbre $v$ à un nœud fils $v'$ correspond au coût du plus court chemin dans $\mathcal{G}$ de $v$ à $v'$. Nous appelons cet arbre l'arbre de recherche mandatory. Celui-ci présente une structure similaire au Travelling Salesman Problem (TSP), mais diffère de par le choix du point de départ et d'arrivée qui sont différents. Il s'agit d'un problème NP-difficile.

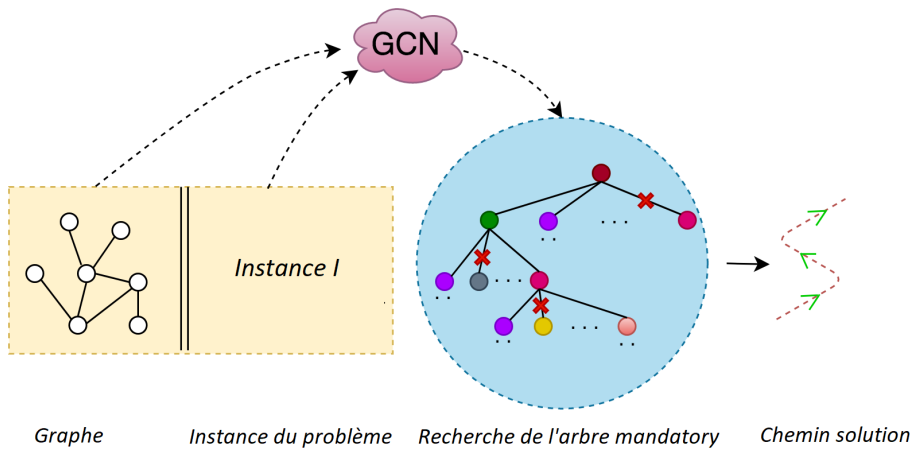

**Figure 2: Illustration du procédé général proposé.**

Afin de résoudre $I$ de manière optimale, nous proposons d'explorer cet arbre et de rechercher le chemin de la racine jusqu'à une feuille qui présente le coût de transition total (somme des coûts de transition entre chaque nœud du chemin dans l'arbre) le plus faible. Pour ce faire, nous proposons d'entraîner un Graph Convolutional Network (GCN) dans la phase hors-ligne à prédire un ordre de visite des nœuds obligatoires pour toute instance. Dans la phase en-ligne, pour résoudre une instance $I$, ce GCN est utilisé pour obtenir l'ordre de visite suggéré des nœuds obligatoires. Le coût de cet ordre suggéré est utilisé ensuite pour initialiser l'upper bound (borne sup) d'un algorithme Branch & Bound (B&B) qui recherche une solution optimale de visite des nœuds obligatoires dans l'arbre de recherche mandatory. L'introduction de cet upper bound initiale permet d'accélérer la recherche B&B. Le chemin optimal est ensuite reconstruit à partir de la solution optimale trouvée par la recherche B&B. Ce procédé est illustré dans la Figure 2.

## Construction de Chemin avec un GCN

Dans cette section, nous proposons dans un premier temps un encodage d'instances en vecteurs. Ensuite nous proposons une architecture GCN capable de prendre ces vecteurs en entrée et de prédire en sortie un nœud obligatoire à visiter. Enfin, nous définissons une utilisation récursive du GCN qui permet de définir un ordre de visite complet des nœuds obligatoires pour une instance donnée.

Pour toute instance $I = (start, dest, M)$, nous définissons un encodage qui associe à $I$ un vecteur $\mathbf{x}$ constitué de triplets d'attributs pour chaque nœud dans $\mathcal{G}$. Le vecteur **x** est donc constitué d'un total de $3\,|\mathcal{V}|$ attributs. Les 3 attributs pour chaque nœud $v_j \in \mathcal{V}$ sont:

- Un attribut *start*:

$$st_j = \begin{cases} 1, & \text{si } v_j \text{ est le nœud de départ } start \text{ de } I \\ 0, & \text{sinon} \end{cases}$$

- Un attribut *end*:

$$ed_j = \begin{cases} 1, & \text{si } v_j \text{ est le nœud de destination } dest \text{ de } I \\ 0, & \text{sinon} \end{cases}$$

- Un attribut *mandatory* :

$$my_j = \begin{cases} 1, & \text{si } v_j \text{ est un nœud obligatoire de } M \text{ dans } I \\ 0, & \text{sinon} \end{cases}$$

Le vecteur **x** est la concaténation de ces attributs:

$$\mathbf{x} = (st_1, ed_1, my_1, st_2, ed_2, my_2, ..., st_{|\mathcal{V}|}, ed_{|\mathcal{V}|}, my_{|\mathcal{V}|}) \tag{2}$$

Notre réseau de neurone est un GCN et non un Multi-Layer Perceptron (MLP). Nous justifions ce choix par le fait que ceci nous permet une réduction significative de l'espace des données en entrée: l'architecture GCN permet de prendre la matrice d'adjacence $A$ de $\mathcal{G}$ en compte directement dans ses convolutions graphes. Ceci n'est pas le cas pour un MLP qui est composé de couches entièrement connexes (fully-connected). Ainsi, l'utilisation d'un MLP implique la prise en compte de la connectivité du graphe $\mathcal{G}$ dans les inputs $\mathbf{x}_i$, ce qui augmenterait la taille des données en entrée de $|\mathcal{V}|^2$ et complexifierait le modèle à entraîner.

L'architecture de notre GCN utilise les couches GCN de (Kipf and Welling, 2017). Ces couches GCN possèdent la loi de propagation suivante de la couche $l$ à $l+1$, *i.e.* pour calculer la matrice des attributs $H^{(l+1)}$ depuis la matrice des attributs $H^{(l)}$:

$$H^{(l+1)} = \sigma\left( \tilde{D}^{-\frac{1}{2}} \tilde{A} \tilde{D}^{-\frac{1}{2}} H^{(l)} \theta^{(l)} \right), \tag{3}$$

où $\tilde{A} = A + I_N$ est la matrice d'adjacence avec l'ajout de connections pour chaque nœud à lui même; $I_N$ est la matrice identité. La matrice $\tilde{D}$ est la matrice des degrés diagonale de $\tilde{A}$. Enfin, $\sigma(\cdot)$ est la fonction d'activation, pour laquelle nous choisissons ReLU($\cdot$) = max(0, $\cdot$). Notre GCN $f$ prend en entrée toute instance $I$ encodée en vecteur $\mathbf{x}$. Trois couches de convolution de type graphe sont appliquées successivement, chacune couplée avec la fonction d'activation ReLU($\cdot$) = max(0, $\cdot$). Ensuite, une couche fully-connected est appliquée et donne en sortie un vecteur $z \in \mathbb{R}^{|\mathcal{V}|}$. Enfin, nous utilisons la fonction *softmax* (cf Equation 3.9 en anglais) pour convertir $z$ en une distribution de probabilités $\hat{\mathbf{y}} \in \mathbb{R}^{|\mathcal{V}|}$ pour tous les nœuds du graphe. Pour résumer, $f(\mathbf{x}) = \hat{\mathbf{y}}$. De plus, arg max($\hat{\mathbf{y}}$) correspond au choix du GCN concernant prochain nœud obligatoire à visiter depuis le nœud *start* en vue de résoudre optimalement $I$.

Le GCN est entrainé de manière supervisée sur des données d'entraînement avec label, *i.e.* un ensemble de paires input-output ($\mathbf{x}_i, \mathbf{y}_i$) sélectionnées depuis une base d'apprentissage. Dans notre cas, $\mathbf{x}_i$ est une instance $I$ et $\mathbf{y}_i$ est l'index du prochain nœud obligatoire pour $\mathbf{x}_i$ dans un chemin solution optimal de $I$. Nous entraînons le GCN sur des instances qui ont déjà été résolues optimalement par un solveur exact, qui sert d'*enseignant*. Le GCN apprend à approximer les solutions données par le solveur. Pour ce faire, nous optimisons la perte cross-entropy multi-classe définie à l'Equation 3.10 par descente stochastique de gradient.

Enfin, pour une instance donnée $I = (start, dest, M)$, le GCN prédit uniquement le 1$^{er}$ nœud obligatoire à visiter. Afin d'obtenir un ordre de visite complet des nœuds obligatoires $M$, nous effectuons une utilisation récursive du GCN en nous basant sur les choix correspondants $q_i$ du prochain nœud obligatoire sélectionné par le GCN à chaque étape. Plus précisément, après que le GCN calcule une prédiction $q_i$, nous créons une sous-instance pour laquelle le nœud de départ devient le nœud $q_i$ que le GCN vient de prédire, et pour laquelle la liste de nœuds obligatoire reste la même que précédemment à l'exception de $q_i$ qui lui est retirée, *i.e.* ($q_i, dest, \tilde{M}_i = \tilde{M}_{i-1} \setminus q_i$), où $\tilde{M}_0 = M$ et $q_0 = start$. Le GCN est appelé récursivement sur ces sous-instances jusqu'à obtenir une sous-instance avec seulement un seul nœud obligatoire restant, après quoi un ordre de visite des nœuds obligatoires est défini.

## Apprentissage auto-supervisé

Nous présentons une stratégie d'apprentissage auto-supervisé pour entraîner le GCN sur le graphe $\mathcal{G}$. Dans un premier temps, nous définissons un domaine de planification pour résoudre des instances, ensuite nous présentons une version modifiée de l'algorithme A* pour générer des données optimales afin d'entraîner le GCN.

Dans cette section, nous assimilons une instance à un état de planification $s =$

$(start, dest, M)$. Un état terminal est une instance résolue, *i.e.* une instance pour laquelle tous les nœuds obligatoires ont été visités et le nœud de destination a été atteint. Nous notons les instances résolues $F_i = (i, i, \varnothing), i \in \{1, 2, ..., |V|\}$. Il y a exactement autant d'états terminaux qu'il y a de nœuds dans $\mathcal{G}$. Nous définissons respectivement les *successeurs* et les *prédécesseurs* d'un état $s = (start, dest, M)$ en tant que $succ(s)$ et $pred(s)$ comme indiqué dans le Tableau 1.

**Table 1:** Règles de transition vers les successeurs et les prédécesseurs pour les états de planification.

| $s = (start, dest, M)$ | |
|---|---|
| **État successeur $s'$** | **État prédécesseur $s'$** |
| $s' = (start', dest', M')$ | $s' = (start', dest', M')$ |
| $(start, start') \in \mathcal{E}$ | $(start', start) \in \mathcal{E}$ |
| $dest' = dest$ | $dest' = dest$ |
| $M' = M \backslash \{start'\}$ | $M' = M$                     $\triangleright$ [1]* |
| - | ou |
| - | $M' = M \cup \{start\}$       $\triangleright$ [2]* |
| $s$ n'est pas un état terminal | $s'$ n'est pas un état terminal |
| Coût de transition: $(start, start')$ | Coût de transition: $(start', start)$ |

[1]* seulement si $start' \notin M$

[2]* seulement si $start \neq dest$

Le coût de transition d'un état à un état voisin est le coût de l'arrête dans $\mathcal{G}$ qui relie les nœuds de départ de chacun de ces deux états. Avec ces règles, le nœud de destination reste toujours le même. Par conséquent, nous utilisons la version 'backwards' (recherche depuis l'état final vers l'état initial) de A* depuis chaque instance résolue $F_i$ comme état initial (en utilisant *pred* comme règle de succession), sans état final particulier recherché. Pour chaque état $s$ visité par A*, un chemin $p$ est construit de $s$ à $F_i$ qui est considéré être le plus court par l'algorithme. Nous définissons $g(s)$ en tant que coût de $p$, $a(s)$ comme prochain état visité après $s$ dans $p$, et $d(s)$ comme le premier nœud obligatoire des nœuds obligatoires de $s$ qui est visité dans $p$.

De plus, nous apportons les changements suivants à A*. Premièrement, lorsqu'un chemin plus court que connu est trouvé vers un état $s'$ lors du développement de l'état $s$, *i.e.* $g(s') > g(s) + c(s, s')$, les valeurs $a(s')$ et $d(s')$ sont aussi mises à jour en plus de $g(s')$ pour incorporer le nouveau chemin. Deuxièmement, nous utilisons l'heuristique $h(s) = 0$ pour tout état $s$. Puisque A* fonctionne en sens inverse depuis un état terminal $F_i$, nous ne désirons pas que l'algorithme atteigne un certain état en particulier, mais nous voulons atteindre autant d'états que possible. Ce choix garantit que lorsqu'un état $s$ est extrait de la liste prioritaire OPEN des états restants à développer, un chemin optimal de $s$ à $F_i$ a déjà été trouvé. Choisir $d(s)$ comme le prochain nœud obligatoire à visiter permet donc une résolution optimale. Par conséquent, nous ajoutons la paire $\langle s, d(s) \rangle$ dans la base d'apprentissage. Le pseudocode de cet algorithme est donné (en anglais) dans Algorithm 2. Les données générées par l'algorithme sont ajoutées à la base d'apprentissage, qui est ensuite mélangée aléatoirement. Le GCN est entraîné sur ces données par apprentissage supervisé.

## Recherche Branch & Bound avec GCN

Afin de résoudre une instance, nous utilisons l'algorithme de profondeur d'abord B&B qui est réputé pour son efficacité computationelle (Narendra and Fukunaga, 1977). L'algorithme B&B est appelé sur l'arbre de recherche mandatory pour déterminer un ordre de visite optimal des nœuds obligatoires. De plus, B&B a la spécificité de ne pas développer des branches lorsque cela n'est pas nécessaire. Pour cela, l'algorithme utilise une lower bound (borne inf) et une upper bound (borne sup). Pour un nœud dans l'arbre, la lower bound est la somme du coût total depuis la racine jusqu'au nœud avec une heuristique admissible $\zeta$ qui évalue le coût minimal restant du nœud jusqu'à une feuille de l'arbre. Nous utilisons l'heuristique définie à l'Equation 3.11 (en anglais). L'upper bound elle est initialisée à $+\infty$ au début, puis à chaque fois qu'une solution de coût plus faible que la meilleure solution connue est trouvée, ce coût est affecté à l'upper bound. Une branche conduisant à un noeud pour lequel la lower bound est supérieure à l'upper bound est coupée et non explorée car elle ne peut pas contenir de solution de coût plus faible que la meilleure solution trouvée actuellement.

Enfin, nous définissons GCN + B&B, qui fonctionne de la façon suivante pour résoudre une instance $I = (start, dest, M)$. Dans un premier temps, l'instance est encodée en vecteur $\mathbf{x}$. Le GCN est ensuite appelé récursivement sur $\mathbf{x}$ et ses sous-instances pour définir un ordre de visite des nœuds obligatoires $M$. Le coût associé à cet ordre de visite est ensuite utilisé en tant qu'upper bound initial (au lieu de $+\infty$) pour l'algorithme B&B qui cherche une solution optimale dans l'arbre de recherche mandatory. Cette upper bound permet à B&B d'effectuer des coupes plus tôt dans l'arbre et ainsi d'accélérer la recherche. Une fois l'ordre de visite optimal trouvé par B&B, le chemin solution optimal correspondant est reconstruit en concaténant les plus courts chemins. La Figure 3 résume ce procédé.

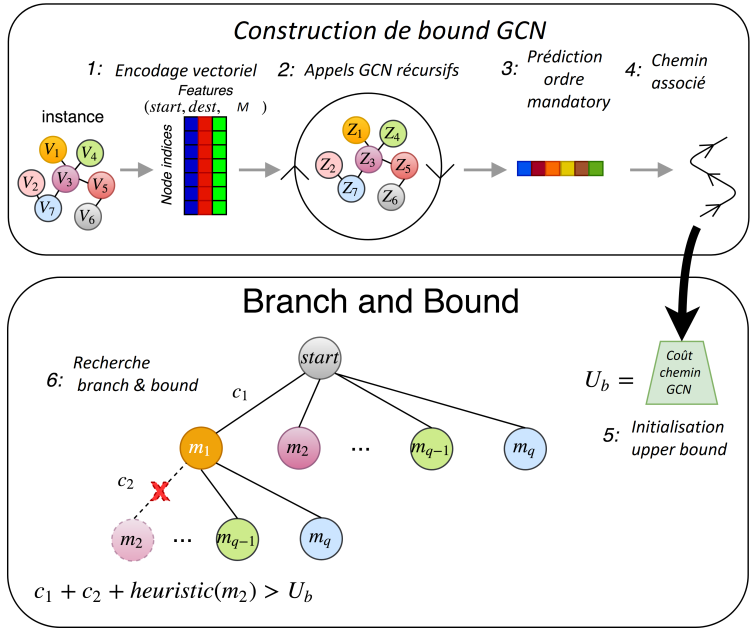

**Figure 3: Illustration du procédé GCN + B&B.**

## Expérimentations

Nous effectuons des expérimentations pour évaluer l'impact du GCN sur B&B en terme de performance. Nous considérons 4 graphes, $\mathcal{G}_1$, $\mathcal{G}_2$, $\mathcal{G}_3$ et $\mathcal{G}_4$, avec respectivement 15, 23, 22 et 23 nœuds. Ces graphes représentent des scénarios réalistes de VA dans lesquels de l'aide doit être acheminée dans certaines zones (Guettier, 2007). Nous générons 1512, 2928, 2712 et 2712 instances benchmark pour chaque graphe respectivement, avec le générateur d'instances défini dans l'Algorithme 1 (différent du générateur A* utilisé pour entraîner le GCN). Nous générons les instances par groupes avec une cardinalité de nœuds obligatoires croissante, allant de 5 à 12.

Nous comparons 4 différents solveurs sur les instances benchmark. Chaque solveur retourne une solution optimale. Le premier est la Programmation Dynamique (PD) qui fouille l'arbre de recherche mandatory. Le second est l'algorithme B&B qui fouille également l'arbre de recherche mandatory, avec et sans l'upper bound du GCN (GCN + B&B et B&B). Enfin, le dernier est A* utilisé dans le domaine de planification défini dans le Tableau 1 avec une heuristique admissible conçue spécialement pour ce type de problème basée sur les Minimum Spanning Trees (MST), que nous notons A* + MST. Les résultats pour le graphe $\mathcal{G}_2$ sont rapportés dans la Figure 4. Les expérimentations sur les autres graphes montrent des résultats similaires. Nous notons que pour les instances avec 7 nœuds obligatoires ou plus, l'algorithme A* + MST appliqué sur le domaine de planification est plus efficace que B&B qui est appliqué sur l'arbre de recherche mandatory. En effet, bien que chaque état de planification prend plus de temps à être calculé pour prendre en compte les spécificités du domaine de planification, au final bien moins d'états de planification sont visités que de nœuds dans l'arbre de recherche mandatory. Ceci est expliqué par le fait que le domaine de planification tire profit de la structure du graphe, ce qui donne A* + MST un net avantage sur PD et B&B. Cependant, lorsque l'upper bound du GCN est utilisée par B&B, B&B bat nettement A*+ MST sur toutes les instances, même les plus complexes, et résiste mieux à l'augmentation du nombre de nœuds obligatoires. Dans le Tableau 3.7 (en anglais), des détails additionnels expliquent cette bonne performance de GCN + B&B par le fait que l'upper bound donnée par le GCN permet d'éviter de lourds calculs nécessaires à l'obtention d'une upper bound ayant un coût proche du coût d'une solution optimale.

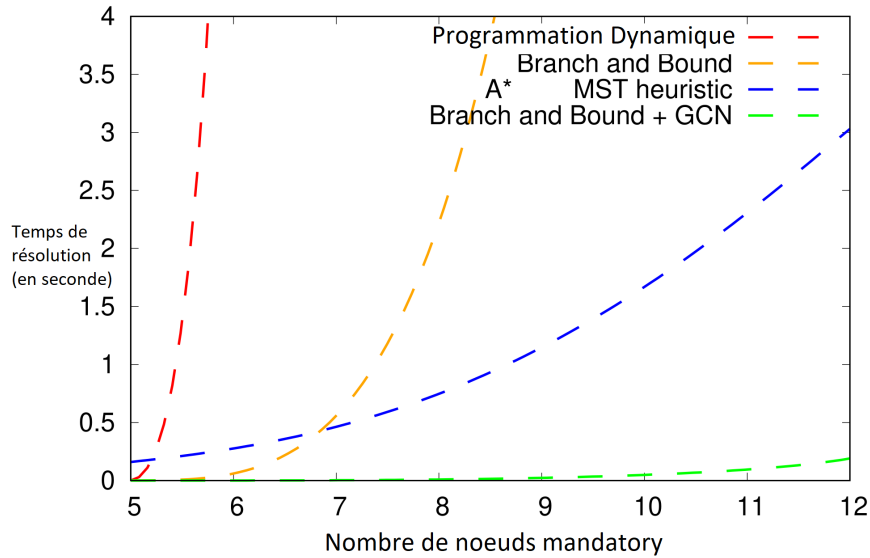

**Figure 4: Comparaison de la performance des différents solveurs sur les instances benchmark du graphe $\mathcal{G}_2$.** L'axe $X$ represente le nombre de nœuds obligatoires des instances, l'axe $Y$ le temps moyen de résolution. Nous limitons $Y$ à [0,4] pour une meilleure échelle de comparaison.

## Conclusion

Nous avons proposé une méthode pour résoudre le problème en question qui combine un GCN avec de la recherche B&B. Nous avons présenté une stratégie d'entraînement auto-supervisée basée sur A* qui permet d'entraîner le GCN et ainsi l'adapter à un graphe $\mathcal{G}$ dans la phase hors-ligne. Les expérimentations montrent que le GCN parvient à généraliser à des instances créées avec un générateur différent que celui utilisé pour son entraînement. Ensuite, pour la phase en-ligne, nous résolvons les instances efficacement avec GCN + B&B, un framework qui utilise une upper bound prédite par le GCN pour accélérer la performance de la recherche B&B. De plus, GCN + B&B résiste nettement mieux aux problèmes complexes que B&B seul, et démontre une performance nettement supérieure à un algorithme A* avec une heuristique développée manuellement et adaptée au domaine de planification en question.

# Planification de Stratégies de Manœuvres

Nous considérons dans cette section un scénario pour lequel le VA se trouve dans un convoi militaire et un VP, qui possède un équipage, est assigné au VA. Les deux véhicules doivent avancer en manœuvre 'perroquet', c'est à dire qu'ils sont espacés de profil, l'un est situé devant l'autre, et seul un véhicule peut avancer à la fois. Le véhicule situé à l'arrière avance, dépasse le véhicule à l'arrêt et s'arrête une distance précise devant celui-ci. Ensuite les rôles sont inversés.

Nous supposons que la manœuvre de chaque véhicule prend $\delta$ unités de temps. Nous utilisons des variables temporelles $a_1, a_2, ..., a_n$ pour faire référence aux manœuvres successives du VA, qui correspondent à des actions contrôlables pour le VA. De la même façon, nous utilisons des variables temporelles $u_1, u_2, ..., u_n$ pour les manœuvres successives du VP, qui correspondent à des actions incontrôlables pour le VA. L'affectation d'une valeur $t$ à une des ces variables signifie que la manœuvre correspondante est exécutée à l'instant $t$ et qu'elle se terminera à $t + \delta$. Bien entendu, le VA ne peut affecter des valeurs qu'aux variables contrôlables. De plus, l'exécution de chaque variable $a_i$ *déclenche* l'activation de $u_i$ dans l'intervalle de temps $[\delta, D_{max}]$ après $a_i$, *i.e.* après chaque manœuvre $a_i$ du VA le VP va effectuer sa manœuvre $u_i$ entre $a_i + \delta$ (temps minimum requis pour que le VA finisse sa manœuvre et que le VP soit autorisé à rouler) et $a_i + D_{max}$ (où $D_{max} - \delta$ correspond au temps maximum que le VP peut rester immobile après autorisation de déplacement).

Enfin, les contraintes liées au déplacement perroquet imposent que chaque $a_{i+1}$ soit exécuté dans l'intervalle de temps $[\delta, H_{max}]$ après $u_i$, *i.e.* après la manœuvre $u_i$ du VP, le VA peut décider de commencer sa manœuvre $a_{i+1}$ au plus tôt à $u_i + \delta$ (pour que le VP soit arrêté) et au plus tard à $u_i + H_{max}$ (où $H_{max} - \delta$ est le temps maximum que le VA peut rester à l'arrêt après autorisation de déplacement). La Figure 5 illustre le déplacement perroquet.

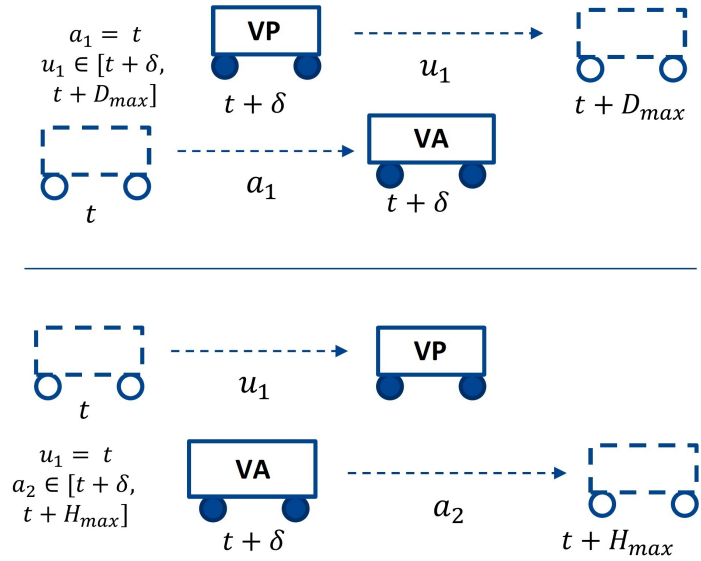

**Figure 5: Illustration du déplacement perroquet.**

Nous considérons un scénario dans la Figure 6 dans lequel le VA et le VP sont au sein d'un convoi militaire se dirigeant vers le nord le long d'un bosquet. Un trou dans le bosquet va créer une exposition du convoi à un objet non identifié détecté par radar de l'autre côté du bosquet. Le temps actuel de la figure étant $t$, l'exposition aura lieu entre $t + \Delta_1$ et $t + \Delta_2$. Le VA devra être entièrement à l'arrêt pendant l'exposition pour couvrir le VP et si besoin protéger l'équipage du VP. Pour ce faire, le VA doit calculer, en amont (*i.e.* au temps $t$ dans la figure), une stratégie réactive d'exécution de ses manœuvres $a_i$ qui s'adaptera aux manœuvres incertaines du VP au fur et à mesure qu'elles sont observées de telle sorte que le VA soit à l'arrêt entre $t + \Delta_1$ et $t + \Delta_2$ quel que soit le comportement du VP. Ce problème est un problème de planification de stratégie réactive pour un Disjunctive Temporal Network with Uncertainty (DTNU) tel que formalisé dans la partie inférieure de la Figure 6. Enfin, il n'existe pour ce scénario qu'une phase en-ligne. Le terrain où se trouvera le VA et le VP n'est pas connu d'avance, ni les dangers potentiels qu'ils pourront y rencontrer. Nous souhaitons donc une méthode de résolution efficace applicable pour tout DTNU.

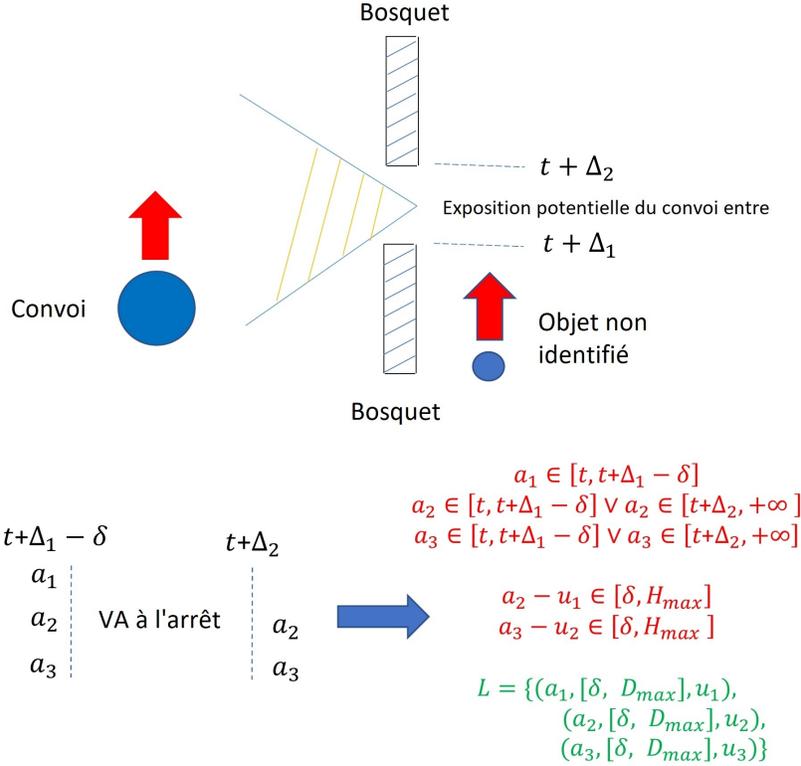

**Figure 6: Illustration du scénario considéré.** Le problème mathématique DTNU correspondant est formalisé en dessous. En rouge: les contraintes du DTNU. Nous imposons que $a_1$ se déroule uniquement avant l'exposition pour démarrer la manœuvre générale avant l'exposition. En vert: les *contingency links* du DTNU.

Un DTNU est composé d'un ensemble de variables contrôlables $a_i \in A$, de variables incontrôlables $u_i \in U$, de contraintes $c_i \in C$ chacune ayant pour forme $\vee_{k=1}^{q} v_{k,j} - v_{k,i} \in [x_k, y_k] \mid v_{k,j}, v_{k,i} \in A \cup U$, $(x_k, y_k) \in \mathbb{R}^2$ et enfin de *contingency links* $l_i \in L$, chacun de la forme $(a_i, \vee_{k=1}^{q'}[x_k, y_k], u_j)$. Un contingency link définit un lien entre une variable incontrôlable $u_j$ et une variable contrôlable $a_i$. Plus précisément, il définit divers intervalles de temps après l'exécution de $a_i$ dans lesquels $u_j$ pourrait s'exécuter (par lui même, car incontrôlable).

Seule une approche connue permet de calculer une stratégie réactive pour un DTNU, dans une forme de contrôlabilité dite *Dynamic Controllability* (DC). Il s'agit de l'approche proposée dans (Cimatti, Micheli, and Roveri, 2016; Cimatti et al., 2014) qui est basée sur les Timed-Game Automatas (TGA). Dans nos travaux, nous étudions la controllabilité dynamique des DTNUs en tant que problème de recherche, ce qui nous permet d'introduire une nouvelle forme de contrôlabilité dite Restricted Time-based Dynamic Controllability (R-TDC). Nous proposons ensuite un algorithme tree search nommé TS pour rechercher des stratégies R-TDC de DTNUs. De plus, nous utilisons une heuristique basée sur un type de graph neural network appelé Message Passing Neural Network (MPNN) pour guider TS et accélérer sa recherche de stratégie. Nous appelons le framework guidé MPNN-TS. Nous comparons la performance de TS et MPNN-TS contre PYDC-SMT, le solveur le plus performant de l'état de l'art DC issu de (Cimatti, Micheli, and Roveri, 2016), et réalisons des performances supérieures.

## Recherche de Stratégie en Contrôlabilité Temporelle R-TDC

Une stratégie DC fonctionne de la manière suivante. **(1)** Soit elle exécute des variables contrôlables à un instant $t$ si aucune variable incontrôlable ne s'exécute entre temps; **(2)** Soit elle attend l'exécution de quelconque variable incontrôlable. Une stratégie R-TDC quant à elle: **(1)** Soit exécute des variables contrôlables à l'instant actuel; **(2)** Soit attend une durée de temps $\Delta_t$ non interruptible. Chaque attente dans une stratégie R-TDC peut être de durée arbitraire. Les variables incontrôlables sont susceptibles de s'exécuter à l'intérieur de l'intervalle d'attente. De plus, R-TDC permet également l'exécution d'une variable contrôlable au même moment que s'exécute une variable incontrôlable dans un intervalle d'attente, nous appelons ce procédé *exécution réactive*. Dans R-TDC, seules des bornes sont connues concernant les temps d'exécution des variables incontrôlables, ce qui rend R-TDC moins flexible que DC qui peut attendre le moment exact de l'exécution d'une variable incontrôlable pour réagir. R-TDC est inclus dans DC et implique DC. La Figure 6.1 et sa description (en anglais) détaille des exemples de stratégies R-TDC pour deux différents DTNUs.

Nous présentons maintenant notre approche tree search TS. Celle-ci permet de rechercher des stratégies R-TDC pour un DTNU. TS discrétise les intervalles pendant lesquels des variables incontrôlables sont activées mais ne se sont pas encore exécutées (*i.e.* pour chacune, la variable contrôlable correspondante a été exécutée, donc le contingency link correspondant définit dans quels intervalles de temps elle est susceptible de s'exécuter). Ainsi, la discrétisation de ces intervalles est utilisée pour prendre en compte les différentes possibilités d'exécution des variables incontrôlables et rechercher une stratégie réactive de planification adaptée. Pour ce faire, l'algorithme TS construit un arbre de recherche pour lequel la racine est le DTNU pour lequel nous recherchons une stratégie R-TDC, les autres nœuds de l'arbre sont des sous-DTNUs de ce DTNU pour lesquels des décisions ont été prises dans l'arbre, et enfin certains nœuds sont des nœuds logiques (*OR, AND*) qui représentent respectivement une bifurcation pour prendre une décision et les diverses combinaisons possibles d'exécution des variables incontrôlables. A un nœud de l'arbre donné, nous pouvons prendre des décisions telles que l'exécution d'une variable contrôlable non encore exécutée, ou alors attendre une période de temps définie. Des nœuds fils sont alors créés (des sous-DTNUs) pour lesquels ces décisions sont propagés et intégrés dans les contraintes. Enfin, la contrôlabilité R-TDC d'une feuille de l'arbre, *i.e.* un sous-DTNU pour lequel toutes les variables contrôlables ont été exécutées et toutes les variables incontrôlables se sont exécutées dans des intervalles spécifiques, indique si ce sous-DTNU a bien été résolu ou non à la fin (*i.e.* si ses contraintes ont bien été satisfaites ou non). Nous qualifions également la contrôlabilité R-TDC d'un nœud DTNU dans l'arbre comme valeur 'truth'. Enfin, TS combine de façon logique la contrôlabilité R-TDC des nœuds DTNUs enfants pour déterminer la contrôlabilité R-TDC des nœuds DTNU parents.

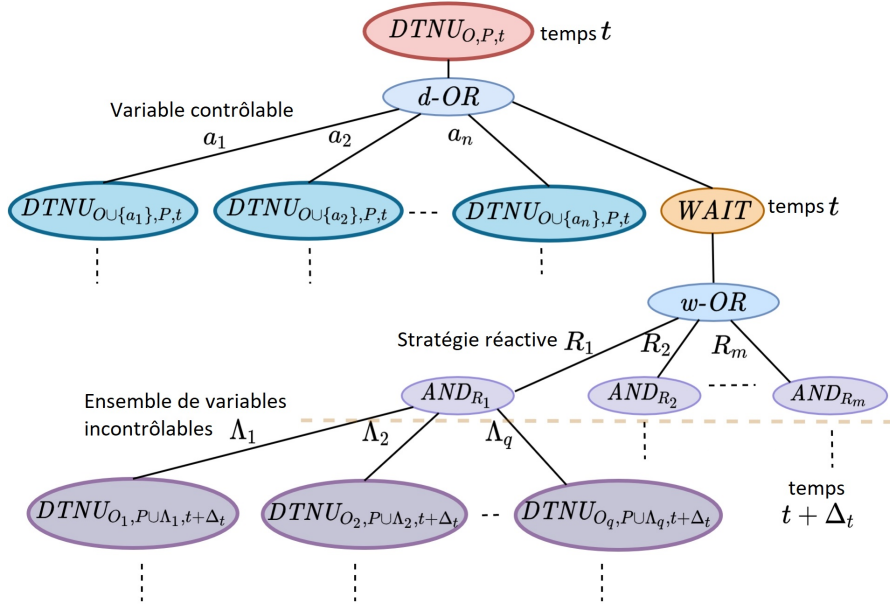

**Figure 7: Structure de l'arbre construit par TS.**

**Structure de l'Arbre de Recherche**

Soit $\Gamma = \{A, U, C, L\}$ un DTNU. Ici, $A$ est l'ensemble des variables contrôlables, $U$ l'ensemble des variables incontrôlables, $C$ la liste des contraintes et $L$ l'ensemble des contingency links. La racine de l'arbre de recherche est $\Gamma$. Il y a 4 types de nœuds différents dans l'arbre et chaque nœud possède une valeur truth, qui est initialisée à *inconnu* au départ, et peut soit devenir True, soit False. Les différents types de nœuds sont listés ci-dessous et illustrés dans la Figure 7.

**Nœuds *DTNU*.**  Tout nœud DTNU autre que le nœud DTNU original $\Gamma$ correspond à un sous-problème de $\Gamma$ à un instant $t$ donné dans le temps, pour lequel certaines variables contrôlables peuvent avoir été exécutées plus haut dans l'arbre, un laps de temps peut s'être écoulé, et certaines variables incontrôlables peuvent s'être exécutées. Un nœud DTNU est composé des mêmes variables $A$ et $U$, contraintes $C$ et contingency links $L$ que le DTNU $\Gamma$. Le nœud comporte également une mémoire $S$ qui retient à quel instant exact, ou dans quel intervalle de temps, les variables exécutées précédemment dans l'arbre l'ont été. Enfin, le nœud garde aussi trace des intervalles d'activation des variables incontrôlables activées $B$. La mémoire $S$ est utilisée pour créer une liste de contraintes $C'$ mise à jour suite à la propagation du temps d'exécution, ou de l'intervalle du temps d'exécution, des variables dans les contraintes $C$ tel que décrit dans § 6.2.3.5 (en anglais). Un nœud DTNU non-terminal, *i.e.* un nœud DTNU pour lequel toutes les variables n'ont pas été exécutées, a exactement un nœud fils: un nœud *d-OR* .

**Nœuds *OR*.**  Quand un choix peut être effectué à l'instant $t$, ce contrôle de transition est représenté par un nœud *OR*. Nous distinguons deux types de nœuds OR, les nœuds *d-OR* et *w-OR* . Pour les nœuds *d-OR* , le premier type de choix possible est quelle variable contrôlable $a_i$ exécuter. Ce type de choix conduit à un nœud DTNU. L'autre type de choix possible est d'attendre une période de temps, ce qui conduit à un nœud *WAIT* . Les nœuds *w-OR* peuvent être utilisés pour créer des *stratégies d'attente réactive, i.e.* qui stipulent que certaines variables contrôlables doivent être exécutées de façon réactive à des variables incontrôlables pendant une attente *WAIT* . Le parent d'un nœud *w-OR* est un nœud *WAIT* , ses fils sont des nœuds *AND* .

**Nœuds *WAIT* .**  Le nœud *WAIT* est utilisé après une décision d'attendre une certaine période de temps $\Delta_t$. Le parent d'un nœud *WAIT* est un nœud *d-OR* . Un nœud *WAIT* a exactement un fils: un nœud *w-OR* , qui a pour but d'explorer des stratégies d'attente réactives différentes. La gestion d'incertitude liée aux nœuds incontrôlables est gérée par les nœuds *AND* .

**Nœuds *AND*.** Les nœuds *AND* sont utilisés après qu'une décision d'attente est prise et qu'une stratégie d'attente réactive est décidée, représenté respectivement par un nœud *WAIT* et *w-OR*. Chaque fils du nœud *AND* est un nœud DTNU à l'instant $t + \Delta_t$, $t$ étant le temps avant l'attente et $\Delta_t$ la durée de l'attente. Chaque fils représente une combinaison possible d'exécution des variables incontrôlables qui sont susceptibles de s'exécuter. Ces combinaisons sont construites à partir de l'ensemble des variables incontrôlables activées pour lesquels le temps d'exécution potentiel coïncide avec l'attente.

La Figure 7 montre également comment un sous-problème de $\Gamma$, que nous notons $DTNU_{O,P,t}$, est développé. Ici, $O \subset A$ est l'ensemble des variables contrôlables qui ont déjà été exécutées, $P \subset U$ l'ensemble des variables incontrôlables qui sont déjà survenues, et $t$ le temps. Ce nœud racine possède exactement un fils: un nœud *d-OR*. Le nœud *d-OR* à son tour présente plusieurs nœuds DTNU fils $DTNU_{O \cup \{a_i\},P,t}$ ainsi qu'un nœud fils *WAIT*. Chaque nœud $DTNU_{O \cup \{a_i\},P,t}$ correspond à un sous-problème qui est obtenu à partir de l'exécution d'une variable contrôlable $a_i$ à l'instant $t$. Le nœud *WAIT* désigne le processus d'attendre une période de temps donnée $\Delta_t$ dans la figure, avant de prendre la prochaine décision. Le nœud *WAIT* conduit directement à un nœud *w-OR* qui établit la liste des différentes stratégies d'attente réactives $R_i$. S'il y a $l$ variables incontrôlables activées, il y a $2^l$ sous-ensembles de variables incontrôlables $\Lambda_i$ qui pourraient s'exécuter pendant l'attente. Chaque nœud $AND_{R_j}$ possède, pour chaque $\Lambda_i$, un nœud DTNU sous-problème où la combinaison $\Lambda_i$ est supposée survenue. En effet, chaque sous-problème $DTNU_{O_i,P \cup \Lambda_i,t+\Delta_t}$ du nœud $AND_{R_j}$ est un DTNU à l'instant $t + \Delta_t$ pour lequel toutes les variables incontrôlables dans $\Lambda_i$ sont supposées s'être exécutées durant l'attente, *i.e.* dans l'intervalle de temps $[t, t + \Delta_t]$. De plus, quelques variables contrôlables peuvent avoir été exécutées de façon réactive pendant l'attente et avoir été incluses dans l'ensemble des variables contrôlables exécutées $O_i$. Si ce n'est pas le cas, $O_i = O$.

Deux types de feuilles existent dans l'arbre. Le premier type est un nœud $DTNU_{A,U,t}$ pour lequel toutes les variables contrôlables $a_i \in A$ ont été exécutées et toutes les variables incontrôlables $u_i \in U$ se sont exécutées. Le second type est un nœud $DTNU_{A \setminus A',U,t}$ pour lequel toutes les variables incontrôlables $u_i \in U$ se sont exécutées, mais quelques variables contrôlables $a_i' \in A'$ n'ont pas été exécutées. La vérification des contraintes du premier type est triviale: tous les temps d'exécutions des variables sont propagés dans les contraintes avec la méthode décrite (en anglais) dans § 6.2.3.5. La valeur truth de la feuille devient *true* si toutes les contraintes sont satisfaites, *false* sinon. Pour le deuxième type, nous propageons les temps d'exécutions de toutes les variables incontrôlables ainsi que ceux de toutes les variables contrôlables exécutées, et obtenons une liste de contrainte $C'$ mise à jour. La feuille en question, $DTNU_{A\setminus A',U,t'}$ correspond donc à un DTNU caractérisé par $\{A', \varnothing, C', \varnothing\}$, c'est donc un DTN. Nous lui ajoutons les contraintes $a_i' \geq t, \forall a_i' \in A'$ et utilisons un solveur MILP (Mixed Integer Linear Programming, (Cplex, 2009)) pour résoudre ce DTN. Si une solution est trouvée, le temps d'exécution de chaque $a_i' \in A'$ est stocké et la valeur truth de la feuille est assignée *true*. Sinon, on lui assigne *false*. Après qu'une valeur truth est assignée à la feuille, une fonction de propagation de valeur truth est appelée pour effectuer une inférence logique des valeurs truth pour les nœuds parents autant que possible. Enfin, l'algorithme TS explore l'arbre en profondeur d'abord. A chaque nœud *d-OR* , *w-OR* et *AND* , les fils sont visités dans leur ordre de création. Une fois un fils sélectionné, le sous-arbre entier associé à ce fils sera visité par TS avant d'explorer un autre fils. Le pseudocode de TS est donné (en anglais) dans l'Algorithme 4.

### Caractéristiques de l'Arbre et son Exploration

**Action d'attente *WAIT* .** Lorsqu'une action d'attente est envisagée à partir d'un DTNU, nous prenons en compte toutes les possibilités de transition vers les DTNUs fils en examinant les variables incontrôlables activées. Ces dernières appartiennent à deux catégories. La 1ère est les variables incontrôlables activées $Z$ dont l'intervalle d'activation coïncide avec le temps d'attente et continue au delà de l'attente. La 2ème est les variables incontrôlables activées $H$ dont l'intervalle d'activation coïncide avec le temps d'attente mais se termine avant la fin de l'attente. Ces dernières sont donc certaines de s'exécuter avant la fin de l'attente. Plus de détails sont donnés dans § 6.2.3.1 (en anglais).

**Période *WAIT* et Stratégies d'Attente Réactives**  Nous définissons 3 règles qui définissent, pour chaque nœud *WAIT* , quel est le temps de l'attente correspondant. Chacune de ces règles examine les éléments du nœud DTNU parent: ses contraintes, ses variables incontrôlables activées et ses contingency links et donne un temps d'attente. Le minimum des 3 temps d'attente est retenu. Nous montrons dans nos expérimentations que ces règles établissent une discrétisation du temps qui permet en général une résolution efficace sans perte significative de capacité à trouver de stratégies. Plus de détails sont donnés dans § 6.2.3.2 (en anglais). Enfin, lors d'une action d'attente *WAIT* , un nœud *w-OR* liste les différentes stratégies d'attente réactives possibles. Ces différentes stratégies sont construites en examinant les contraintes du nœud DTNU parent et chacune permet l'exécution d'une variable contrôlable au même moment qu'une variable incontrôlable s'exécute pendant l'attente (cf § 6.2.3.3 en anglais).

**Propagation de Valeurs Truth.** Lorsqu'une valeur $\beta$ (True ou False) est assignée à une feuille de l'arbre, TS est momentanément arrêté et une fonction récursive de propagation est appelée pour effectuer une inférence logique de la valeur assignée $\beta$ vers le haut de l'arbre. Le procédé est le suivant. Un nœud parent $\omega$ est sélectionné récursivement et nous distinguons les cas suivants:

- Le nœud parent $\omega$ est un nœud DTNU ou un nœud *WAIT* : Nous assignons $\beta$ à $\omega$.

- Le nœud parent $\omega$ est un nœud *d-OR* ou *w-OR* : si $\beta = true$, alors nous assignons *true* à $\omega$. Si $\beta = false$ et tous les nœuds fils de $\omega$ ont une valeur *false* , nous assignons *false* à $\omega$. Sinon, la propagation s'arrête.

- Le nœud parent $\omega$ est un nœud *AND* : si $\beta = false$, alors nous assignons *false* à $\omega$. Si $\beta = true$ et tous les nœuds fils de $\omega$ ont la valeur *true* , nous assignons *true* à $\omega$. Sinon, la propagation s'arrête.

Après l'arrêt de la propagation, TS reprend là où il s'était arrêté. Si une valeur *true* atteint la racine de l'arbre de recherche, cela signifie qu'une stratégie R-TDC a été trouvée. Si une valeur *false* atteint la racine, cela signifie qu'aucune stratégie R-TDC n'a été trouvée. Le pseudocode de l'algorithme de propagation des valeurs truth est donné dans l'Algorithme 5 (en anglais).

**Propagation de Contraintes et Optimisations** Nous définissons des règles pour propager les temps d'exécution des variables dans les contraintes dans § 6.2.3.5 (en anglais). Si ces temps d'exécution sont exacts, la propagation est triviale. Si ces temps d'exécution sont seulement bornés (seul l'intervalle est connu, ce qui est le cas par exemple après une attente durant laquelle une variable incontrôlable s'est exécutée), nous définissons un concept appelé *tight bound* qui nous permet de propager ces bornes dans les contraintes d'une façon qui garantit la validité des stratégies R-TDC. Enfin, nous définissons également des optimisations pour TS dans § 6.2.3.6 (en anglais). Ces optimisations permettent entre autre d'éviter l'exploration des parties symétriques de l'arbre de recherche, ainsi que les nœuds de l'arbre pour lesquels la valeur truth est déjà définie ou peut l'être.

## Heuristique MPNN

Nous utilisons une heuristique basée sur l'architecture graph neural network MPNN de (Gilmer et al., 2017) pour guider TS. Cette architecture définit des règles de transmission de messages et permet de prendre en entrée un graphe dans lequel les nœuds et les arrêtes ont des attributs. Soit $\Gamma = \{A, U, C, L\}$ un DTNU. Nous expliquons dans un premier temps comment nous convertissons $\Gamma$ en un graphe $\mathcal{G} = (\mathcal{K}, \mathcal{E})$. D'abord, nous convertissons l'échelle de toutes les valeurs temporelles d'absolu en relatif en posant $t = 0$ pour le temps actuel de $\Gamma$. Nous recherchons à l'intérieur de tous les intervalles de temps convertis $[x_i, y_i]$ dans $C$ et $L$ la valeur la plus grande $d_{max}$, *i.e.* le point le plus loin dans le temps. Nous normalisons toutes les valeurs dans $C$ et $L$ en les divisant par $d_{max}$. Par conséquent, toutes les nouvelles valeurs sont comprises entre 0 and 1. Ensuite, nous convertissons chaque variable contrôlable $a \in A$ et incontrôlable $u \in U$ en nœuds d'un graphe avec des attributs *contrôlable* ou *incontrôlable*. Les contraintes temporelles dans $C$ et les contingency links $L$ sont représentées par des arrêtes entre les nœuds avec comme attributs 10 classes de distances différentes $(0 : [0, 0.1], 1 : [0.1, 0.2], ..., 9 : [0.9, 1])$. Nous utilisons également des attributs supplémentaires pour les arrêtes pour inclure le type d'arrête considéré (contrainte, disjonction, contingency link, signe de direction pour les bornes inf et sup des intervalles). De plus, des nœuds intermédiaires sont utilisés avec un attribut distinct pour représenter les différentes disjonction entre les contraintes et les contingency links. Enfin, nous ajoutons un *WAIT* node avec un attribut distinct qui représente implicitement l'acte d'attente *WAIT* . La Figure 6.5 (en anglais) donne un exemple précis de conversion d'un DTNU en graphe.

La conversion en graphe $\mathcal{G}$ de $\Gamma$ est caractérisée par 3 éléments: la matrice des attributs des nœuds $X_\kappa$, la matrice d'adjacence du graphe $X_\epsilon$ et la matrice des attributs des arrêtes $X_\rho$. Ces éléments définissent l'entrée du MPNN. Notons $f$ le MPNN, $\theta$ l'ensemble de ses paramètres. Le MPNN est composé de plusieurs couches de message passing de (Gilmer et al., 2017). Chacune de ces couches est alternée avec la fonction d'activation ReLU$(\cdot) = \max(0, \cdot)$, sauf la dernière couche. Pour la dernière, nous utilisons la fonction sigmoid $\sigma(\cdot) = \frac{1}{1+\exp(-\cdot)}$ pour obtenir une liste de probabilités $\pi$ sur tous les nœuds du graphe $\mathcal{G}$ : $f_\theta(X_\kappa, X_\epsilon, X_\rho) = \pi$.

La probabilité de chaque nœud $\kappa$ dans $\pi$ correspond à la probabilité de transiter dans un DTNU R-TDC (*i.e.* ayant une valeur *true* ) depuis le DTNU orignal $\Gamma$ en prenant la décision correspondant à $\kappa$. Si $\kappa$ représente une variable contrôlable $a$ dans $\Gamma$, sa probabilité correspondante dans $\pi$ est la probabilité que le sous-DTNU issu de l'exécution de $a$ soit R-TDC. Si $\kappa$ représente une décision d'attente *WAIT* , sa probabilité est celle que le nœud *WAIT* ait une valeur *true* , *i.e.* la probabilité que tous les DTNUs fils issus de l'attente soient R-TDC. Nous appelons ces deux types de nœuds les nœuds *actifs*. Sinon, si $\kappa$ est un autre type de nœud, sa probabilité n'est pas utile au problème et est ignorée. Notre MPNN est entraîné sur des DTNU générés et résolus dans une base d'apprentissage en minimisant la perte binaire cross-entropy multi-classe prenant en compte uniquement les nœuds actifs:

$$\frac{1}{m} \sum_{i=1}^{m} \sum_{j=1}^{q} -Y_{ij} \log(f_\theta(X_i)_j) - (1 - Y_{ij}) \log(1 - f_\theta(X_i)_j)$$

Ici $X_i = (X_{i_\kappa}, X_{i_\epsilon}, X_{i_\rho})$ est le DTNU numéro $i$ au sein d'une base de $m$ exemples d'apprentissage, $Y_{ij}$ est la contrôlabilité R-TDC (1 ou 0) du nœud actif $j$ pour le DTNU numéro $i$.

Un avantage significatif de cette architecture est que le même MPNN peut prendre en entrée des graphes de n'importe quelle taille (*i.e.* avec un nombre de nœuds variables, et une connectivité variable), à condition que le nombre d'attributs pour chaque nœud et arrête reste le même, ce qui est le cas avec notre conversion de DTNU en graphe. Par conséquent, notre MPNN peut prendre en entrée n'importe quel DTNU (converti en graphe), quel que soit son nombre de variables, contraintes, contingency links. Enfin, l'heuristique MPNN est utilisé par TS de la façon suivante. Lorsqu'un nœud *d-OR* est atteint, le nœud DTNU parent est converti en graphe et le MPNN $f$ est appelé sur les éléments du graphe correspondant $X_\kappa, X_\epsilon, X_\rho$. Les nœuds actifs dans les probabilités $\pi$ données par le MPNN sont ordonnés par probabilité décroissante, et TS visite les nœuds fils de l'arbre dans cet ordre suggéré, donnant priorité ainsi aux nœuds fils ayant une plus grande chance d'être R-TDC. Enfin, pour générer la base de données d'apprentissage pour entraîner le MPNN, nous utilisons dans un premier temps un générateur DTNU pour créer des DTNUs aléatoires, puis un algorithme auto-supervisé défini dans § 6.2.5 (en anglais), qui se base sur une version modifiée de TS reposant sur des simulations aléatoires de courtes durées. Les DTNUs utilisés pour l'entraînement possèdent entre 10 et 20 variables contrôlables, et au plus 3 variables incontrôlables.

## Expérimentations

Nous faisons des expérimentations sur le même ordinateur pour évaluer l'efficacité de TS, MPNN-TS (TS guidé par le MPNN) et PYDC-SMT ordered (solveur DC de l'état de l'art de (Cimatti, Micheli, and Roveri, 2016)). Dans cette section, MPNN-TS représente TS guidé jusqu'au 15ème nœud *d-OR* rencontré en profondeur, depuis la racine, tandis que MPNN-TS-X est guidé jusqu'au Xème nœud *d-OR* . Nous utilisons dans un premier temps le benchmark de (Cimatti, Micheli, and Roveri, 2016) duquel nous enlevons les DTNs et STNs. Le benchmark résultant est composé de 290 DTNUs et 1042 STNUs. Ici, limiter l'utilisation en profondeur maximale du MPNN à 15 offre une bonne balance entre gain en guidage et coût d'appel du MPNN. Les résultats sont donnés dans la Figure 8.

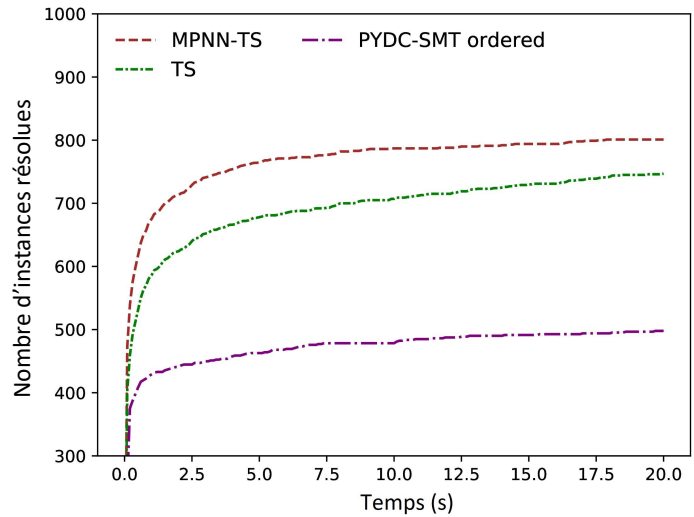

**Figure 8: Expérimentations sur le benchmark de (Cimatti, Micheli, and Roveri, 2016).** L'axe X représente le temps alloué en secondes, l'axe Y le nombre d'instances dans le benchmark que chaque solveur peut résoudre dans le temps alloué. Le timeout est réglé à 20 secondes par instance.

Nous observons que TS résout environ 50% d'instances de problème en plus que PYDC-SMT parmi le temps alloué (20 secondes). De plus, TS résout 56% de toutes les instances, les instances restantes résultent en timeout. Parmi les instances résolues, une stratégie est trouvée pour 89% d'entre elles, les 11% restantes sont prouvées non-R-TDC. PYDC-SMT lui résout 37% de toutes les instances. Une stratégie est trouvée pour 85% des instances que PYDC-SMT résout alors que les 15% restantes sont prouvées non-DC. Enfin, parmi toutes les instances résolues par PYDC-SMT, TS résout 97% de celles-ci avec exactement la même conclusion *i.e.* R-TDC pour DC et non-R-TDC pour non-DC. De plus, l'utilisation de l'heuristique MPNN permet de résoudre 6% d'instances en plus sur le temps alloué. Nous justifions ce faible gain par le fait que la plupart des problèmes résolus dans ce benchmark sont de petits problèmes avec peu de variables, et sont donc résolus très rapidement. Malgré cela, le fait que l'heuristique MPNN permet un gain en efficacité montre que le MPNN généralise bien à des DTNUs créés avec un générateur DTNU différent que celui utilisé pendant l'entraînement du MPNN, *i.e.* le biais introduit par notre générateur de DTNU est limité.

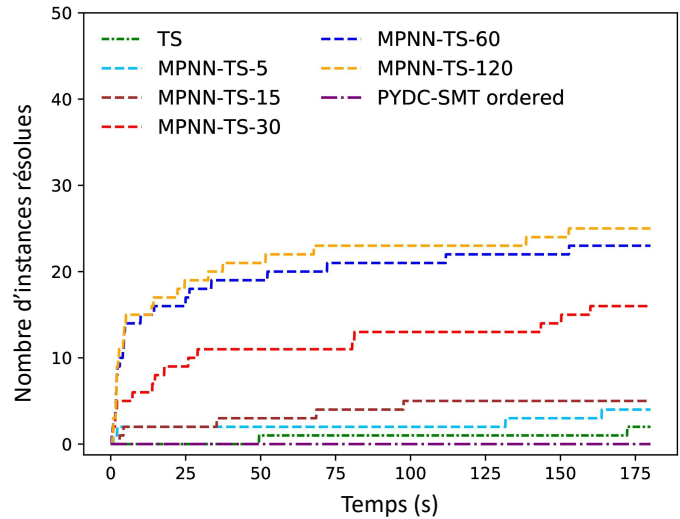

**Figure 9: Expérimentations sur le benchmark $B_3$.** Les axes sont les mêmes que dans la Figure 8. Le timeout est réglé à 180 secondes.

Pour évaluer notre heuristique MPNN de manière plus précise, nous générons des benchmarks avec des DTNUs plus complexes en utilisant notre générateur DTNU avec un nombre de variables plus élevé. Chaque benchmark contient 500 DTNUs ayant de 1 à 3 variables incontrôlables. De plus, chaque DTNU possède entre 10 à 20 variables contrôlables dans le 1er benchmark $B_1$, 20 à 25 dans le 2ème benchmark $B_2$ et 25 à 30 dans le dernier benchmark $B_3$. Nous rapportons les résultats des expérimentations sur $B_1$, $B_2$ et $B_3$ respectivement dans la Figure 6.7 (en anglais), Figure 6.8 (en anglais) et Figure 9 (ci-jointe).

D'abord, nous notons que pour ces 3 benchmarks aucun algorithme ne parvient à prouver une non-contrôlabilité R-TDC ou non-contrôlabilité DC avant de subir un timeout, due à la grande taille de ces problèmes. PYDC-SMT montre de mauvaises performances sur $B_1$ et ne résout aucune instance de $B_2$ et $B_3$. TS a une mauvaise performance sur $B_2$ et résout seulement 2 instances de $B_3$. Cependant, nous notons un large gain en performance avec l'utilisation de l'heuristique MPNN qui varie en fonction sa profondeur maximale d'utilisation. A la profondeur la plus adaptée, le gain est de +91% instances résolues pour $B_1$, +980% pour $B_2$ et +1150% pour $B_3$. Le plus de variables une instance possède, le plus le guidage donné par l'heuristique semble compenser son coût d'appel. En effet, nous notons que la profondeur optimale d'utilisation du MPNN augmente avec la taille du problème: 15 pour $B_1$, 60 pour

xxviii

$B_2$ et 120 pour $B_3$. Nous justifions ceci par le fait que plus de variables implique un arbre de recherche plus large, surtout pour les sections plus profondes de l'arbre où le gain de l'heuristique ne compensait pas nécessairement son coût pour les problèmes plus petits. De plus, le MPNN est entraîné sur des DTNUs ayant 10 à 20 variables contrôlables. Les gains observés sur les benchmark $B_2$ et $B_3$ suggèrent que le MPNN généralise à des problèmes plus grands que ceux qui on été utilisés pour son entraînement.

L'approche proposée présente une bonne balance entre complétude de recherche et efficacité: presque tous les exemples résolus par PYDC-SMT dans le benchmark de (Cimatti, Micheli, and Roveri, 2016) sont résolus avec la même conclusion, et beaucoup d'autres que PYDC-SMT ne peut résoudre le sont également. De plus, notre approche R-TDC fonctionne mieux sur des problèmes avec plus de variables, et la structure de l'arbre permet l'utilisation d'heuristiques basées sur de l'apprentissage. Bien que ces heuristiques ne sont pas la clé pour résoudre les problèmes de très grande échelle, nous avons montré qu'elles peuvent tout de même apporter un gain non négligeable en performance.

## Conclusion

Nous avons formulé mathématiquement notre problème en DTNU. Afin de calculer des stratégies réactives pour ces DTNUs, nous avons proposé un nouveau type de contrôlabilité pour les DTNUs nommé R-TDC. Nous avons également présenté un algorithme TS qui est capable de rechercher des stratégies R-TDC pour un DTNU. Les stratégies sont construites en discrétisant le temps et explorant les différentes décisions qui peuvent être choisies à des moments clés, et en intégrant comment les variables incontrôlables peuvent s'exécuter. Nous montrons dans nos expérimentations que TS résout les DTNUs en R-TDC plus efficacement que le solveur état de l'art DC (PYDC-SMT) les résout en DC, avec presque toujours la même conclusion, malgré le fait que R-TDC ne comporte pas la complétude de DC. Enfin, nous avons créé MPNN-TS, le framework qui combine l'heuristique MPNN avec TS. L'heuristique MPNN est entraînée par apprentissage auto-supervisé et généralise sur des DTNUs de taille plus grande que ceux utilisés pour son entraînement. MPNN-TS résiste bien mieux à la complexité des DTNUs et peut traiter des DTNUs avec plus de variables. Un avantage significatif de MPNN-TS est que le même MPNN entraîné peut être appliqué sur n'importe quel graphe (DTNU) de n'importe quelle taille. Ainsi, notre framework MPNN-TS est prêt à être utilisé dans la phase en-ligne dans toute situation. Des expérimentations supplémentaires indiquent qu'il est capable de calculer des stratégies intégrant 40 manœuvres perroquet (20 contrôlables, 20 incontrôlables) en moins de 2.5 secondes et 60 manœuvres perroquet en moins de 6 secondes, ce qui est amplement satisfaisant.

# Bibliographie

# Table of Contents

# List of Tables

# List of Figures

# Chapter 1

# Introduction and Context

Research and works carried out in this thesis were part of a CIFRE thesis contract involving Safran Electronics & Defense, Paris Dauphine University's LAMSADE laboratory and Saint-Cyr Coetquidan's CREC laboratory. They are part of a wider effort to incorporate state-of-the-art machine learning algorithms into autonomous vehicles. Learning algorithms are increasingly being explored for planning purposes, and this thesis follows in line with this trend.

Automated planning and scheduling is an area of interest in artificial intelligence with a wide panel of applications. It is becoming increasingly critical in a world where autonomy is starting to prevail. In this thesis, we focus on Autonomous Unmanned Ground Vehicles (AUGV) and Autonomous Ground Vehicles (AGV) in off-road situations. AUGV and AGV operations are constrained by terrain structure, observation abilities, embedded resources. Missions must be executed in minimal time, while meeting objectives. Missions of interest for this thesis include disaster relief, logistics or area surveillance, where maneuvers must consider terrain knowledge. Other types of missions of interest include cooperative maneuvers with partner vehicles in military convoys, where maneuvers must be synchronized in a way which guarantees convoy safety. In most applications, the ability to plan maneuvers in the environment has a direct impact on operational efficiency. AUGVs and AGVs integrate several perception capabilities (on-line mapping, geolocation, optronics, LIDAR) in order to update situation awareness on-line. This knowledge is used by various on-board planning layers to maintain mission goals, provide navigation waypoints and dynamically construct platform maneuvers. Resulting actions and navigation plans are used to control the robotic platform. Once plans are computed, trajectory is automatically managed and navigation waypoints are followed using control algorithms and time sequence. Figure 1.1 presents the eRider, an AUGV which can also serve as AGV with high mobility and the ability to perform maneuvers on difficult terrain for exploring disaster areas.

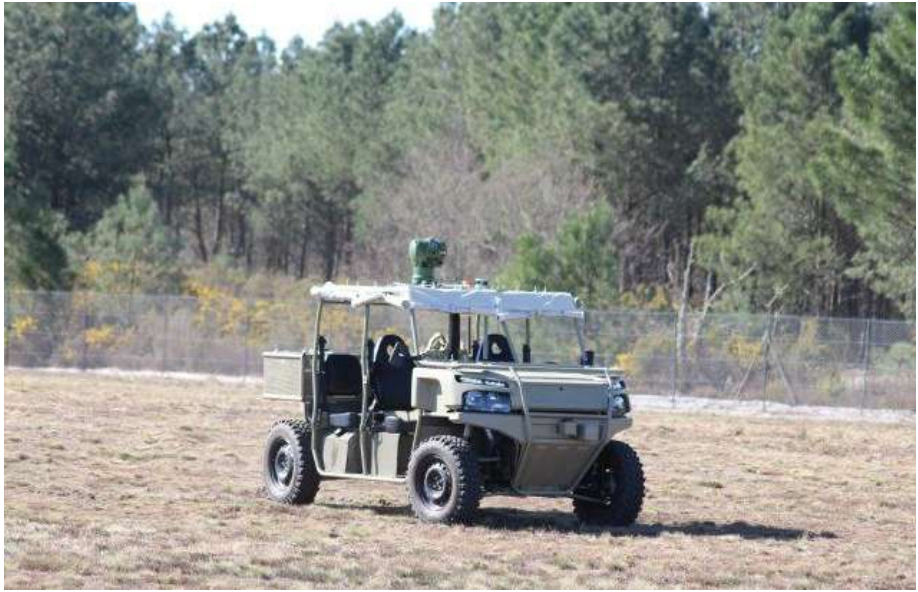

**Figure 1.1: The autonomous vehicle eRider, developed by SAFRAN.** It is an all-purpose optionally piloted vehicle that can patrol a given area, observe at long range or carry goods for various off-road applications in difficult environments.

Three main problems are studied in this thesis and presented in three scenarios that follow. In the first scenario, an AUGV is tasked with bringing supplies to remote areas hit by flooding. The AUGV is able to travel between locations by taking off-road paths, and the aim of the AUGV's planning system is to conceive travel plans which minimize the total distance the AUGV needs to travel to accomplish objectives. In the second scenario, an AGV is carrying passengers in an area which was subject to an earthquake. Passengers are tasked with performing damage assessment of hardest-hit areas. Again, the AGV is travelling on off-road paths, and additionally, some paths may require intervention of passengers for manual driving because of potentially difficult weather. The AGV's planning system needs to conceive plans which minimize the total distance required to meet mission objectives, as well as minimize the likelihood of requiring manual driving so as to allow passengers to focus on other tasks. Lastly, in the third scenario, an AUGV is travelling in a military convoy, in off-road situations. The AUGV has a partner vehicle, which is manned, and needs to perform synchronized maneuvers to ensure safety of the manned vehicle. However, taking into account maneuvers of the partner vehicle only as the AUGV observes them is insufficient and can lead to risky situations for the manned vehicle.

Instead, the AUGV's planning system needs to compute a safe maneuver strategy some steps prior to executing maneuvers which takes into account the uncertainty linked to the partner vehicle's maneuvers and adapts the strategy accordingly.

In this thesis, we view these problems from a planning perspective. We investigate whether or not machine learning algorithms, particularly neural networks, can benefit such problems. Indeed, neural networks have increasingly been used to learn heuristics and help solve similar types of problems. We make the hypothesis they can help solve the problems considered in this thesis as well, and we carry out experiments supporting this hypothesis. In particular, we use neural network architectures from graph representation learning, which can be naturally coupled with automated planning and scheduling problems that can be expressed as graphs.

## 1.1 Operational Problems

### 1.1.1 First Problem

In this problem, we consider a mission of first-aid supply delivery for an AUGV following flooding in a region. Major roads have become impassable due to the flooding, and the AUGV needs to take off-road paths to travel to stranded villages in need of supplies. The AUGV starts at a given position where it is provided with supplies. The AUGV needs to travel to each of the positions in need of supplies at least once. No order of visit is imposed for those positions. Then, the AUGV needs to make its way to a destination position, which is not necessarily the same as the start position, where its current mission will end and a new one may be given. The AUGV's energy source being limited, it needs to complete those objectives while minimizing the total distance it travels. Figure 1.2 depicts this scenario.

The terrain can be represented as a connected graph $\mathcal{G} = (\mathcal{V}, \mathcal{E})$. In this graph, each node $v \in \mathcal{V}$ represents a position; $\mathcal{V}$ represents all positions. Each edge $e = (v, v') \in \mathcal{E}$ is an off-road path which links two different positions $v, v' \in \mathcal{V}$. Edges have a weight (or cost), which is the distance of the associated path in our case (but it could be some other criterion, like edge traversal time for instance). The AUGV starts its mission at a given node, the start position, needs to visit each mandatory node (each position in need of supplies) at least once, in no specific order, and finish its mission at a destination node. To this end, the AUGV travels on edges, and the objective is to determine an itinerary which satisfies mission objectives and minimizes the sum of all edge weights included in the itinerary. There is no limit to how many times the AUGV can visit a particular node or cross a particular edge. The graph illustration of the scenario is provided in Figure 1.3. In our scenario, edges are undirected, as we do not consider dangerous off-road paths with high slopes where ascent is not possible but descent is. Nevertheless, resolution methods proposed in the thesis for this problem are also applicable to directed graphs. Only the type of heuristic used (the type of graph neural network architecture) needs to be replaced to take into account edge direction.

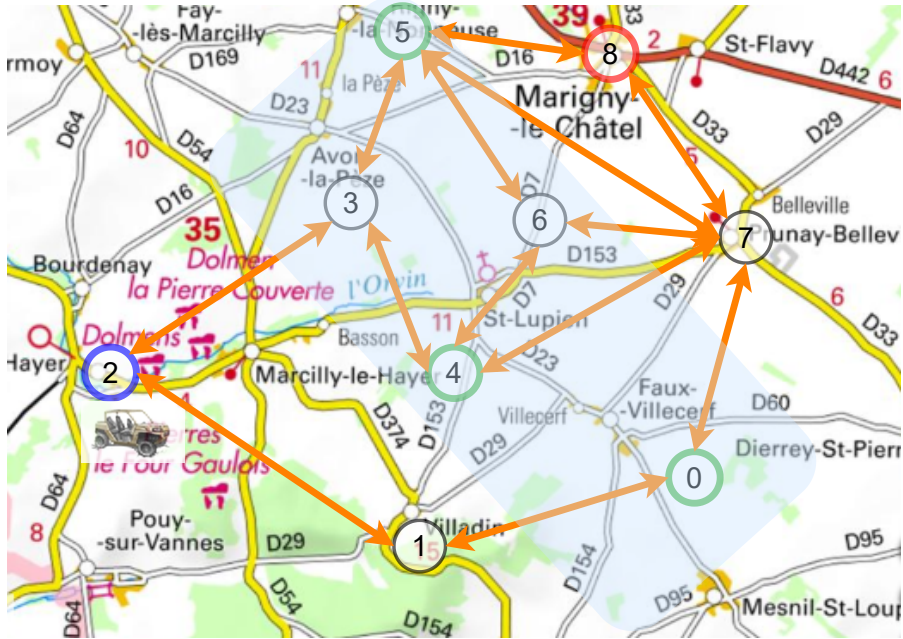

**Figure 1.2: Typical scenario for the first problem.** Area between the Seine and the Vanne rivers near Troyes, France. Circles represent positions in the terrain, orange arrows represent off-road paths linking positions. In this scenario, flooding has caused severe damage in the area, resulting in major roads being impassable. The blue area on the map represents the sub-area most affected by the flooding. The AUGV is located at the blue circle (position 2) and needs to bring supplies to positions 0, 4 and 5 (circled in green), before heading to position 8 (circled in red).

We consider two phases for this scenario: *offline* and *online*. In the offline phase, the AUGV is not on the terrain, but the structure of the terrain (major off-road paths connecting the positions) is known, *i.e.* the graph $\mathcal{G}$ is known. In this phase, the AUGV designer has as much time as desired to conceive a planning system for the AUGV which is suited to this terrain. However, AUGV missions are not known ahead of time, as it is impossible to know with certainty which areas will be in need of supplies in the event of future floods. In the online phase, flooding has occurred and the AUGV is already somewhere on the terrain. Missions can be given to the AUGV at any time. For each mission it is assigned, the AUGV needs to compute, as quickly as possible, an itinerary with the lowest distance that meets mission objectives.

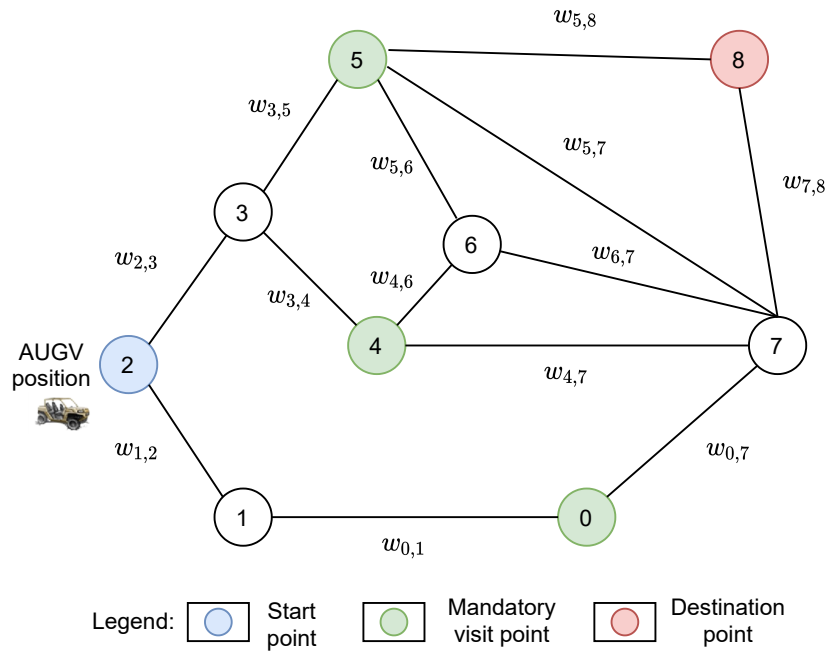

**Figure 1.3: Graph representation of the first problem.** The terrain is represented by a graph $\mathcal{G} = (\mathcal{V}, \mathcal{E})$: nodes represent positions and edges paths linking positions. Each edge $e_{i,j}$ has a weight $w_{i,j}$, corresponding to a cost which needs to be paid by the AUGV to cross the edge. The AUGV is located at the start node (number 2, in blue). Nodes in green (number 0, 4 and 5) represent areas where first-aid and supplies, carried onboard the AUGV, need to be brought to. Finally, the destination node (number 8, in red) represents the position where the AUGV should go to in order to end its mission.

### 1.1.2 Second Problem

We consider here a damage assessment scenario. An earthquake has struck a region, making major roads impassable. The autonomous vehicle is carrying passengers this time, hence we refer to it as AGV. The AGV starts at a given position in the affected terrain and has to take off-road paths to travel between positions. The AGV needs to visit each of the areas hardest hit by the earthquake at least once so that passengers can assess the extent of the damage in those places. No order of visit is imposed for those areas. Finally, the AGV needs to carry the passengers to a final position where they can deliver their assessment of the situation to a crisis unit. The AGV's source of energy is limited, thus it needs to travel the shortest distance possible while meeting these objectives. However, difficult weather may affect the capability of the AGV to drive completely on its own on certain paths, requiring manual driving from passengers. Therefore, in addition to minimizing the total distance it travels, the AGV is also required to avoid paths likely to require human driving intervention as much as possible so that passengers can focus on other tasks than driving.

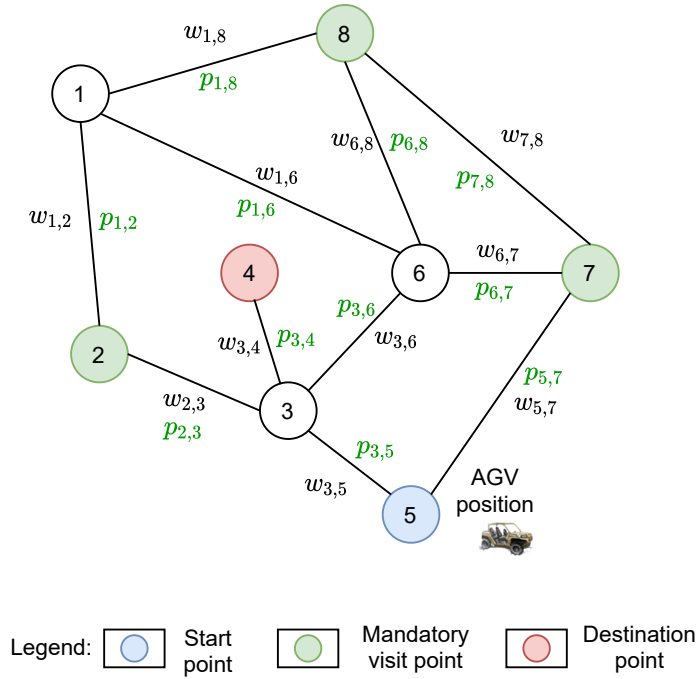

**Figure 1.4: Typical scenario for the second problem.** The terrain is represented by a graph $\mathcal{G} = (\mathcal{V}, \mathcal{E})$: nodes represent positions and edges paths linking positions. Each edge $e_{i,j}$ has a weight $w_{i,j}$, corresponding to a cost which needs to be paid by the AGV to cross the edge, as well as an autonomous feasibility $p_{i,j}$, corresponding to whether the AGV can drive autonomously or not on the edge. The AGV is located at the start node (number 5, in blue). Nodes in green (number 2, 7 and 8) represent areas where AGV passengers need to stop to perform damage assessment of areas hit hardest by the earthquake. Finally, the destination node (number 4, in red) represents the final position where the AGV should take the passengers to so they can report their assessment of the situation.

The terrain is represented by a connected graph $\mathcal{G} = (\mathcal{V}, \mathcal{E})$, where each node represents a position and each edge $e = (v, v') \in \mathcal{E}$ an off-road path which links two different positions $v, v' \in \mathcal{V}$. Here, each edge has two elements: a weight and an *autonomous feasibility*. The weight is the cost the AGV needs to pay to cross the edge, and corresponds to the distance in our case. The autonomous feasibility is a criterion equal to either 1 or 0. It states whether or not, under given weather conditions, the AGV will be able to cross the edge autonomously. The AGV starts at a given node, needs to visit each mandatory node (where AGV passengers have to assess damage) at least once, but in no specific order, and finally head to a destination node where the crisis unit is located. The itinerary computed by the AGV has to meet these objectives, minimize the total distance travelled (sum of the weights of all edges included in the itinerary) and maximize the total autonomous feasibility (sum of the autonomous feasibility of all edges included in the itinerary). There is no limit to how many times the AGV can visit a particular node or cross a particular edge. A key problem is that unlike the distance, the autonomous feasibility of an edge is not known, and depends on terrain structure and weather conditions. This scenario is depicted in Figure 1.4. Edges are undirected in the scenario, but the contribution proposed in the thesis for this problem is applicable to directed graphs.

Finally, there is an offline phase and an online phase. In the offline phase, the AGV is not located on the terrain, but fully knows its structure: not only is the graph $\mathcal{G}$ known, but all terrain characteristics (slopes, obstacles, ...) are known and can be emulated in a simulator. The AGV designer can take as long as desired to conceive a planning system for the AGV which is suited to this terrain. Moreover, the planning system will need to learn in this phase to provide accurate estimates of the autonomous feasibility criterion for each edge under given weather conditions. However, AGV missions are not known ahead of time, as it is impossible to predict which areas will be subject to most damage from a future earthquake. In the online phase, the AGV is on the terrain after an earthquake has struck. Missions can be given to the AGV at any time. For each mission, the AGV needs to compute, as quickly as possible, an itinerary with the lowest distance and highest autonomous feasibility that meets mission objectives.

### 1.1.3 Third Problem

We consider an AUGV travelling among a convoy of military vehicles. More specifically, the AUGV has a designated Partner Vehicle (PV), driven by humans. The AUGV and the PV need to perform synchronized maneuvers in order to meet safety constraints while moving forward. The type of maneuver considered is called "perroquet" (parrot in French). In this type of maneuver, the AUGV and the PV are progressing in the same direction, are laterally spaced and one is in front of the other. Only one vehicle can move at a time: when the vehicle at the front is stopped, the other at the rear can start moving, pass the stopped vehicle, continue until a set distance in front of the stopped vehicle and stop there. This cycle is repeated continuously with roles being reversed during each cycle.

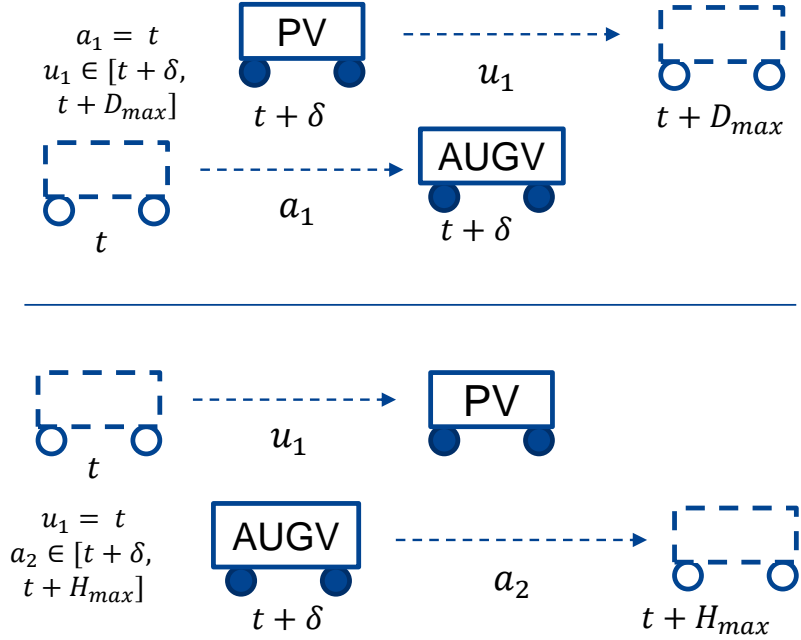

**Figure 1.5: Perroquet maneuver.** In the upper part of the figure, the AUGV has started its movement maneuver $a_1$ at time $t$, and will stop moving at $t + \delta$. The PV will execute its movement maneuver $u_1$ sometime between $t + \delta$ and $t + D_{max}$. In the lower part of the figure, the PV has executed its movement maneuver $u_1$ at time $t$ and will stop moving at $t + \delta$. The AUGV needs to execute its movement maneuver $a_2$ sometime between $t + \delta$ and $t + H_{max}$.

We assume the movement of each vehicle takes $\delta$ time units. We use time variables $a_1, a_2, ..., a_n$ to refer to each successive movement of the AUGV, all of which are 'controllable actions' to the AUGV. Likewise, we use time variables $u_1, u_2, ..., u_n$ to refer to each successive movement of the PV, all of which are 'uncontrollable actions' to the AUGV. The assignment of a value $t$ to a time variable means the movement maneuver corresponding to the time variable will be executed at time $t$. The AUGV can only assign values to controllable time variables. Furthermore, the execution of each time variable $a_i$ *triggers* the activation of $u_i$ in the interval of time $[\delta, D_{max}]$ after $a_i$, *i.e.* after each movement $a_i$ of the AUGV the PV will start its movement maneuver $u_i$ somewhere between $a_i + \delta$ (minimum time required for AUGV to finish its maneuver so the PV is allowed to move) and $a_i + D_{max}$ (where $D_{max} - \delta$ corresponds to the maximum time allowed for the PV to remain still after it is allowed to move). Lastly, constraints impose that each $a_{i+1}$ is executed in the interval of time $[\delta, H_{max}]$ after $u_i$,

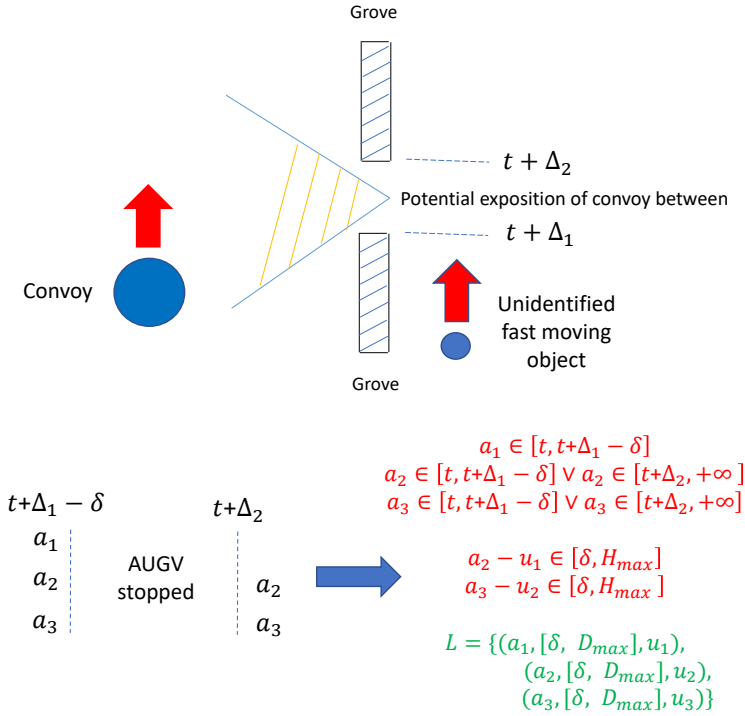

**Figure 1.6:** Typical scenario for the third problem. Current time is $t$. **The AUGV and the PV are among a convoy of vehicles moving north. An unidentified object is also moving north behind a grove, and a gap in the grove is going to expose the convoy between $t + \Delta_1$ and $t + \Delta_2$. The AUGV needs to compute a reactive maneuver strategy which guarantees it will be still during the exposure interval. The problem is expressed as a set of constraints (red) and** *contingency links* **(green).**

*i.e.* after the PV's maneuver $u_i$, the AUGV can decide to start its movement maneuver $a_{i+1}$ at the earliest at $u_i + \delta$ (so that the PV is stopped) and at the latest at $u_i + H_{max}$ (where $H_{max} - \delta$ is the maximum time the AUGV is allowed to remain still after being allowed to start moving). If desired, we can choose to allow the AUGV to remain still for longer periods of time than the PV by choosing $H_{max} > D_{max}$, thus giving the AUGV greater 'control' of the overall movement. Figure 1.5 depicts the perroquet maneuver.

We consider now a scenario in Figure 1.6 in which a convoy of military vehicles is progressing north in a plain alongside a grove. A gap in the grove is going to expose the convoy to an unidentified object detected on radar on the other side of the grove. Current time is $t$ in the figure. Given the velocity of the object, exposure of the convoy will occur between $t + \Delta_1$ and $t + \Delta_2$. The AUGV and the PV are among the convoy of vehicles, and the AUGV needs to be fully stopped during the exposure time so that if the object turns out to be a threat, the AUGV can provide cover and protection for the PV, and potentially act as bait if necessary to protect humans in the PV. To this end, the AUGV needs to plan ahead of time a strategy defining how it should execute its movement maneuvers in accordance with the PV's maneuvers. The PV's maneuvers being uncertain to the AUGV, this strategy needs to reactively adapt the AUGV's maneuvers as it observes how the PV is behaving, in a way which guarantees the AUGV will be stopped between $t + \Delta_1$ and $t + \Delta_2$ regardless of the PV's behavior.

In general, we propose to plan 20 AUGV and 20 PV maneuvers ahead of time with the solving method proposed in the thesis in order to provide strategy computation time compatible with scenario requirements (roughly 2.5 seconds with our proposed solver). For the sake of readability we describe how the AUGV plans strategies for the following 3 AUGV and PV maneuvers (i.e. we only consider $a_1$, $a_2$, $a_3$ and $u_1$, $u_2$, $u_3$). Constraints in Equation 1.1 need to be satisfied to ensure the AUGV is stopped during exposure time, and constraints in Equation 1.2 need to be satisfied to ensure the AUGV is performing perroquet maneuvers correctly. We impose that $a_1$ can be executed only before the exposure interval, unlike $a_2$ and $a_3$ which can be executed either before or after, in order not to allow the AUGV and PV to trivially wait until after the exposure interval to start their maneuvers. Lastly, Equation 1.3 models *contingency links* (see §2.6) which define the intervals of time when the PV will execute its maneuvers $u_i$ following the AUGV's maneuvers $a_i$.

$$a_1 \in [t, t + \Delta_1 - \delta]$$

$$a_2 \in [t, t + \Delta_1 - \delta] \vee a_2 \in [t + \Delta_2, +\infty] \quad (1.1)$$

$$a_3 \in [t, t + \Delta_1 - \delta] \vee a_3 \in [t + \Delta_2, +\infty]$$

$$a_2 - u_1 \in [\delta, H_{max}]$$

$$a_3 - u_2 \in [\delta, H_{max}] \quad (1.2)$$

$$(a_1, [\delta, D_{max}], u_1)$$

$$(a_2, [\delta, D_{max}], u_2) \quad (1.3)$$

$$(a_3, [\delta, D_{max}], u_3)$$

Those equations define a Disjunctive Temporal Network with Uncertainty (DTNU). DTNUs define scheduling problems in which time is uncertain due to the presence of uncontrollable events. To solve this scenario, the AUGV needs to compute a reactive strategy at time $t$ for this DTNU, which maps time and possible uncontrollable event behavior (of the PV) to controllable actions (of the AUGV) that should be taken. In other words, the strategy needs to adapt the AUGV's maneuvers $a_i$ to the PV's maneuvers $u_i$ as they are observed in a way which guarantees all constraints in Equation 1.1 and 1.2 are satisfied.

Lastly, for this problem, there is no offline preparation phase: terrains where such maneuvers are performed can be diverse, and scenario details cannot be known in advance. In the online phase, details are determined on the fly and the AUGV needs to compute a reactive strategy to a random DTNU. Therefore, the AUGV needs an efficient planning system capable of solving different kinds of DTNUs, whose size, constraints and contingency links can vary.

## 1.2 Contributions

This thesis studies benefits of state-of-the-art machine learning algorithms for planning and scheduling purposes, with applications to the problems stated above. We describe in § 2 existing technologies to solve these problems and their limitations, then we present the following contributions:

- Path-planning combining constraint programming and graph neural networks for optimal itinerary computation with constraints.

- Path-planning combining tree search and graph neural networks for optimal itinerary computation with constraints.

- Path-planning combining local search and graph neural networks for approximate itinerary computation with constraints.

- Multi-criteria path-planning with neural network-based criterion prediction and constraint programming for pareto-optimal itinerary computation with constraints.

- Scheduling under uncertainty with a new form of controllability for disjunctive temporal problems with time discretization, tree search and graph neural network heuristic.

The following papers were published during this thesis:

- Osanlou, Kevin, Christophe Guettier, Andrei Bursuc, Tristan Cazenave, and Eric Jacopin (2018). "Constrained Shortest Path Search with Graph Convolutional Neural Networks". In: *Workshop on Planning and Learning (PAL-18). International Joint Conference on Artificial Intelligence (IJCAI)*.

- Osanlou, Kevin, Andrei Bursuc, Christophe Guettier, Tristan Cazenave, and Eric Jacopin (2019). "Optimal Solving of Constrained Path-Planning Problems with Graph Convolutional Networks and Optimized Tree Search". In: *2019 IEEE/RSJ International Conference on Intelligent Robots and Systems (IROS)*. IEEE, pp. 3519–3525.

- Osanlou, Kevin, Christophe Guettier, Andrei Bursuc, Tristan Cazenave, and Eric Jacopin (2019). "Learning-based Preference Prediction for Constrained Multi-Criteria Path-Planning". In: *Scheduling and Planning Applications workshop (SPARK). International Conference on Automated Planning and Scheduling (ICAPS)*.

- Osanlou, Kevin, Jeremy Frank, J. Benton, Andrei Bursuc, Christophe Guettier, Eric Jacopin, and Tristan Cazenave (2020). "Time-based Dynamic Controllability of Disjunctive Temporal Networks with Uncertainty: A Tree Search Approach with Graph Neural Network Guidance". In: *Workshop on bridging the Gap Between AI Planning and Reinforcement Learning (PRL). International Conference on Automated Planning and Scheduling (ICAPS)*.

The following patent was published: Osanlou, Kevin (2019). GCN-LSCSP: Learning-initialized Local Search for Approximate Solving of Complex Constrained Path-Planning Problems.

## 1.3  Thesis Structure

The thesis is organized as follows:

- Chapter 2 describes state-of-the art work relating to the planning, scheduling and machine learning methods used in the thesis.

- Chapter 3 deals with optimal path-planning. In this chapter, we explain in detail the machine learning architectures we use for path-planning. Two methods are proposed for optimal path-planning, one based on constraint programming, the other on tree search.

- Chapter 4 focuses on approximate path-planning. Two approaches are presented, each aiming to train machine learning models via reinforcement learning. One is based on value iteration, the other one combines policy gradient and local search.

- Chapter 5 introduces a multi-criteria path-planning approach in which an unknown criterion is learned with machine learning, and pareto-optimality is achieved with constraint programming.

- Chapter 6 deals with scheduling problems under uncertainty. This chapter studies a new form of reactive scheduling controllability and presents an approach based on tree search and machine learning for solving.

- Chapter 7 concludes works carried out in the thesis and lists potential future works.

# Chapter 2

# State of the Art

This chapter aims to make the reader familiar with state-of-the-art works relating to planning and learning. First, we study state-of-the-art planning algorithms. We give a brief introduction of neural networks. Then we explore in more detail *graph neural networks*, a recent variant of neural networks suited for processing graph-structured inputs. We describe briefly the concept of reinforcement learning algorithms. Next, we study some successful approaches combining neural networks for path-planning. Lastly, we focus on temporal planning problems with uncertainty.

## 2.1  Planning

The aim of planning is to conceive plans in order to achieve a particular goal. Those plans represent a sequence of actions executed by an agent that enable the transition from a *start state* of an environment, where goal requirements are not satisfied, to an *end state* where they are. In some planning tasks, states are fully observable, in others, only partially. Actions taken by the agent can be deterministic (*i.e.* lead to a certain future state) or non-deterministic (*i.e.* lead to different future states based on some probabilities that are either known or not). State variables can be continuous or not, resulting in a possibly finite or infinite number of states. Actions can be taken in parallel or only one at a time, and have a duration or not. There can be several initial start states or only one. There can be several agents or only one. Planning environments can be diverse, varying from simple positioning in a graph to the complex dynamics of a *first person shooter* (FPS) video game.

In classical planning, models are restricted in the following aspects. The environment is fully observable, there is a single agent, states are finite, there is only one known initial start state, actions are instantaneous and deterministic: there is no uncontrollable event. Actions can only be taken one at a time. Therefore, a sequence of actions from a start state will accurately define the end state, which needs to satisfy goal requirements. Generally, classical planning can be represented mathematically by a set $(S, A, P)$ where:

- $S$ is the set of states

- $A$ is the set of actions

- $P$ is a state transition function

The state transition function $P : S \times A \longrightarrow 2^S$ defines a transition from a current state $s \in S$ to another state $s' \in S$ by considering an action $a \in A$.

To express and solve planning tasks in computer science, different languages have been proposed. Each language represents components of the planning environment differently. These include the Stanford Research Institute Problem Solver (STRIPS) (Fikes and Nilsson, 1971) from SRI International and the popular Planning Domain Definition Language (PDDL) (McDermott et al., 1998). NASA introduced its own planning language, the Action Notation Modeling Language (ANML) (Smith, Frank, and Cushing, 2008).

## Applications of Planning in Autonomous Systems

A system is considered autonomous if it is able to generate and execute plans to achieve its assigned goals without human intervention, and if it is able to deal with unexpected events. Planning has benefited autonomous systems greatly in the past 50 years. Early on, Shakey the robot (Nilsson, 1984), the first general-purpose autonomous mobile robot, was a project that saw the rise of a powerful planning algorithm known as A* (introduced in the next sections), still used nowadays. Space exploration has benefited greatly from planning techniques. Autonomy in satellites or other space vehicles reduces the need of human presence as well as communication to ground, which can be especially useful for long term missions. Applications include Deep Space 1 (Muscettola et al., 1998), or more recently the Curiosity rover (Rabideau and Benowitz, 2017) which is currently exploring Mars. Aerospace applications include Unmanned Aerial Vehicles (UAVs). These have a wide array of applications. Civil applications include road traffic monitoring, remote sensing, security, goods delivery (Motlagh, Taleb, and Arouk, 2016), networking and communications (Hayat,

Yanmaz, and Muzaffar, 2016). UAVs can also be used for operational situations such as search and rescue tasks (Silvagni et al., 2017; Doherty and Rudol, 2007). Search and rescue operations are typically very costly both in terms of costs and human resources, and can present human risks. UAVs reduce those costs and their ability to fly autonomously allows to remove human presence for dangerous tasks. Autonomous Unmanned Ground Vehicles (AUGV) and Autonomous Ground Vehicles (AGV) are also at the center of automation efforts where (trajectory) planning is playing a crucial role. Among AGVs, self-driving cars have been the main focus for civil applications given the potentially revolutionary impact they can have on society. The most advanced self-driving cars combine the latest sensors and computer vision tools for environment perception and use planning to make relevant decisions. We refer the reader to (Badue et al., 2020) for a complete survey on self-driving cars. AUGVs on the other hand are intended for other tasks such as typical disaster relief situations, in which they can be required to perform technical actions (*e.g.* observations, measurements, communications, etc...) while navigating mostly in off-road environments across defined trajectories (Guettier and Lucas, 2016). Automation allows AUGVs to perform dangerous tasks without human presence, and AGVs to move on their own while passengers can focus on other activities. Applications of this thesis focus mostly on AUGVs and AGVs.

## 2.2 Motion Planning and Path-Planning

Path-planning consists of finding a path leading to a desired point from a start point. Motion planning consists of determining motion and path decisions for an agent in order to allow it to achieve a specified motion-related task. Motion planning is more general than path-planning in the sense that, in addition to determining a path the agent needs to take to reach an end point from a start point, it also requires motion characteristics for the agent to reach the end point. Such characteristics can be, but are not limited to, a sequence of positions over time, acceleration values to provide in order to reach a potentially required speed or parameters such as directional angles. Figure 2.1 illustrates the example of a motion planning task in which a robot manipulator is tasked with grabbing an object located at a START position and moving it to the GOAL position (Kumar, 2020). The robot has 4 joints which can revolve. The last joint is used to grab and release objects. Let $\alpha_1, \alpha_2, \alpha_3, \alpha_4$ be the angles for each joint, starting from the base of the robot. A planning state is defined by a vector $s = (\alpha_1, \alpha_2, \alpha_3, \alpha_4)$ which entirely defines the position of the robot and the potential object it is carrying. The configuration space $S$ is made of all possible combinations of values each $\alpha_i$ can take. Some states are 'legal', *i.e.* the robot can actually be in those states, others 'illegal', *i.e.* the robot cannot be in those states. For instance, supposing the angle axis is horizontal and revolves counterclockwise, any state written as $(\frac{3}{2}\pi, \alpha_2, \alpha_3, \alpha_4)$ will not be valid regardless of the values of $\alpha_2, \alpha_3, \alpha_4$ since the first joint cannot bend the joined arm downward. By discretizing values of each angle, the robot can determine a series of consecutive angle changes for each joint, which will be considered as actions, that will allow it to grab and move the object from START to GOAL.

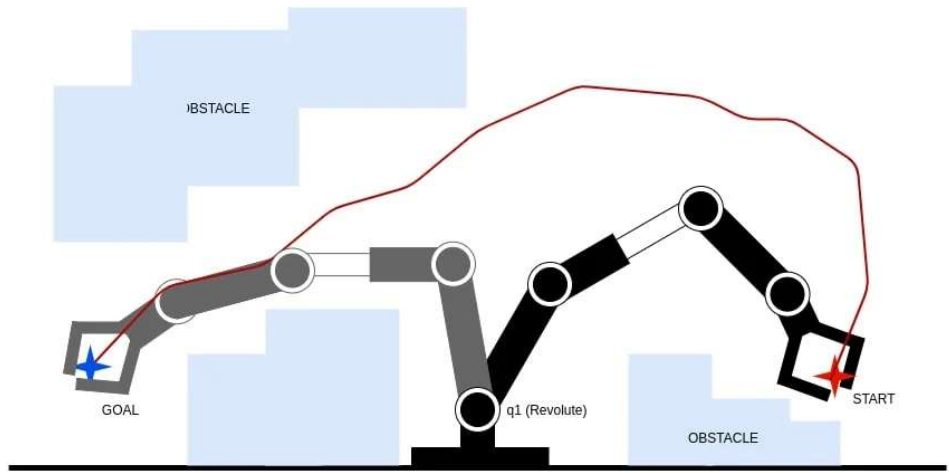

**Figure 2.1: A motion planning task for a robot manipulator.** The robot has to carry an object from the START location to the GOAL position. Source: (Kumar, 2020)

Figure 2.2 shows a path-planning problem in which an agent has to move from a start grid to an end grid. In this environment, black grids represent obstacles. At each step, the agent can move to the adjacent top, left, right or bottom grids. A possible path, in red, allows the agent to fulfill its goal, and minimizes the total distance it needs to travel. The environment in Figure 2.2 can be represented by a geographical graph $\mathcal{G} = (\mathcal{V}, \mathcal{E})$. In this graph, nodes in $\mathcal{V}$ are grids not blocked by obstacles, and edges in $\mathcal{E}$ link adjacent (non-diagonal) grids. Edges are assigned a default weight of 1 as we suppose adjacent grids to be equidistant from one another.

Next, we describe some popular deterministic heuristic-based algorithms for path-planning related problems. These algorithms are well-suited for planning domains with low dimensionality.

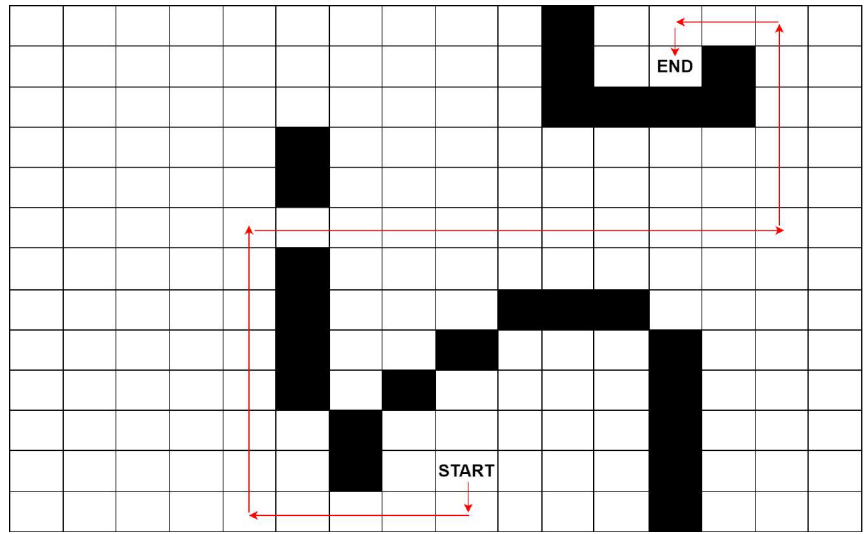

**Figure 2.2: A path-planning task for an agent.** Black grids represent obstacles. The agent is located at the START grid and needs to move to the END grid. The red arrows represent a possible path to satisfy that goal.

## A*

The A* algorithm (Hart, Nilsson, and Raphael, 1968) is a popular best-first search approach to compute an optimal path. Note that A* and related algorithms remain applicable more broadly in other planning domains than path-planning, which is what makes them so popular. A* can be considered as a specialized form of *Dynamic Programming* (DP) (Bellman, 1966). DP essentially breaks down a problem into sub-problems in a recursive fashion and seeks to find the optimal choice to make at each step. This can be expressed as a search tree, which a DP algorithm will explore entirely to return an optimal solution. On the other hand, A* differs in that it will guide search towards most promising states first in order to potentially save a significant amount of computation. In path-planning problems, states are graph nodes and transition cost from a state to another is the cost of the edge linking the corresponding nodes in the graph. A* is complete: it will always find a solution if one exists in a finite search space. Depending on requirements, a heuristic which guarantees to find an optimal solution can be used, or a heuristic which simply aims to find a good solution very efficiently, even if possibly sub-optimal. We describe this process next.

Let $S$ be the finite set of states A* explores, $s_{start}$ the start state where the agent starts and $s_{end}$ the state the agent wants to transition into to satisfy goal requirements. In order to guide search, A* proceeds in a best-first fashion by keeping track, for any state $s$ it explores, of an estimate cost $g(s)$ it took to reach that state from the start state $s_{start}$. Algorithm initialization is as follows: $\forall s \in S$, $g(s) \leftarrow \infty$ and $g(s_{start}) \leftarrow 0$. Additionally, A* uses a heuristic $h$ which estimates the remaining best cost from any state $s$ to the goal state $s_{goal}$. The heuristic can be *admissible* to ensure that A* will return an optimal path: once the goal is reached, the path found is guaranteed to be optimal. The fact that the heuristic is admissible means that for any state $s$, $h(s)$ is lower than or equal to the actual cost of the optimal path from $s$ to $s_{goal}$. A* maintains a priority queue, the *OPEN* list, in which it inserts states by their '$f$' value. For any state $s \in S$, $f(s) = g(s) + h(s)$. It then proceeds to extract the state $s_{min}$ with the lowest such value in the *OPEN* list. A* then develops all neighboring states $s' \in S' \subset S$ it can transition into from the state $s_{min}$. For each of those states, costs are updated if possible. More specifically, $\forall s' \in S'$, if $g(s_{min}) + TC(s_{min}, s') < g(s')$ then $g(s') \leftarrow g(s_{min}) + TC(s_{min}, s')$. Here, $TC(s, s')$ returns the transition cost from state $s$ to state $s'$. Additionally, if $g(s')$ is updated, the state $s'$ is added to the *OPEN* list with its new $f$ value (or its $f$ value is updated if already present in the *OPEN* list). The best predecessor state for $s'$ is also stored in memory if $g(s')$ is updated, *i.e.* $prev(s') \leftarrow s_{min}$, where function $prev$ stores a predecessor for each state. The A* algorithm will keep extracting states from the *OPEN* list until the goal state $s_{goal}$ is extracted, at which point a path has been found (and is optimal if an admissible heuristic is used) from $s_{start}$ to $s_{goal}$.

In this thesis, a planning domain is created for the path-planning problem considered and the A* algorithm is retained for two purposes (§3.3.3). First, it is modified in a way as to create optimal training data efficiently in order to train a learning-based model. Second, it is coupled with a handcrafted heuristic tailored to the path-planning problem in order to provide a strong baseline against which to evaluate our proposed work.

### Incremental Planning

In some scenarios, the agent might not have accurate information about graph structure. The agent may acquire more accurate information about graph structure only when it has started travelling on a computed plan. This is also the case for autonomous vehicles agents if the explored terrain, represented by a graph, is inaccurate at the time of path-planning. It may also be the case if terrain structure changes are likely to happen frequently. The agent will only be able to take into account corrections as it is exploring the terrain.

If the agent computes a path from $s_{start}$ to $s_{goal}$, proceeds on the path, and observes graph changes along the way (*e.g.* edge connection or weight modifications), the computed path may turn out to be in fact sub-optimal after taking into account the new graph structure. In order to compute the new optimal path from the agent's current position $s$ on the path (when the change is observed) to $s_{goal}$, two possibilities exist. The agent can re-plan from scratch in order to compute the shortest path from $s$ to $s_{goal}$. This approach can however cause expensive computations that may be avoidable (*e.g.* if the change in the graph does not change the optimality of the shortest path, or if it's a minor change that can be fixed with a small modification). The other possibility is to leverage information from the previously computed shortest path in order to repair it and make it optimal again. This is the approach taken in *incremental* planning algorithms. Two particularly popular such algorithms are the D* algorithm (Stentz et al., 1995) and an improved lighter version of D*, the D* Lite algorithm (Koenig and Likhachev, 2002a). D* Lite is quite efficient and remains a method of choice even now, with a wide array of applications relying on it (Al-Mutib et al., 2011; Liu et al., 2014; Sun and Zhu, 2016; Belanová et al., 2018).

Simply put, D* and D* Lite aim to re-expand and develop only parts of the search space relevant to registered graph changes and the potential new current state the agent is in. The following provides an overview of how D* Lite operates. First, it computes a path from $s_{start}$ to $s_{goal}$ using backwards A*. Backwards A* works in the same way as A* except the search is done backwards: from the goal state $s_{goal}$ to the start state $s_{start}$. Furthermore, a consistency criterion is used for each state $s$ explored. This criterion compares the cost of the optimal path found from state $s$ to $s_{goal}$ to the minimum of the costs to $s_{goal}$ obtained from each neighboring state plus the transition cost to said neighboring states. The state is said to be consistent if they are equal. Otherwise, it is said to be inconsistent (either overconsistent or underconsistent if respectively higher than or lower than). When a change is observed in the graph while the agent is proceeding on the computed shortest path, edges are updated, and the resulting inconsistent states are re-processed in a defined priority order. Once the process is over, the path has been repaired and is optimal again.

Sometimes changes observed which would result in no impact on the optimality of the path will still require computations by D* and D* Lite to guarantee optimality. Such is the case if some edge weights, all of which are outside the computed path, increase. The algorithm would still need to reprocess states becoming inconsistent due to their connection to edge changes before guaranteeing optimality, even though it is clear the path computed is still optimal. To address this issue, a modified version of D*, delayed D* (Ferguson and Stentz, 2005), has been proposed. To avoid useless computations in such situations, delayed D* initially ignores underconsistent states and only focuses on overconsistent states first. This enables it to potentially save a lot of computations in such cases, making it more suited than D* Lite in some planning domains (Ferguson and Stentz, 2005).

Another incremental approach worthy of note is Lifelong Planning A* (LPA*) (Koenig and Likhachev, 2002b). LPA* starts by running an A* instance to determine an optimal path from a start state $s_{start}$ to a goal state $s_{goal}$. Once edge changes are observed, it uses previous search information to re-compute an optimal path more efficiently in a similar way to D*. The main difference with D* is that LPA* does not allow $s_{start}$ and $s_{goal}$ to be modified. In other words, the approach can only be used before the agent starts moving on the path, in case some last-minute changes are learned (presumably remotely). It is thus unsuitable in situations where the agent observes changes as it is already moving on a computed path and needs to adjust the plan from a new position. More recently, (Przybylski and Putz, 2017) proposed the D* Extra Lite algorithm. Similarly to D* Lite, D* Extra Lite is based on A* and propagates changes to the previously processed search space in order to re-optimize a path. Unlike D* Lite, the reinitialization of the affected search space is achieved by cutting search tree branches. This allows the algorithm to often outperform D* Lite, with experiments suggesting it can be almost up to twice faster on typical path-planning problems.

Previously described approaches are applicable in graphs, and are therefore well-suited to, for example, grid environments where agents can move with 45 or 90 degree angles. Such a representation of the environment can cause the optimal path in the graph to actually be sub-optimal in reality. In *any angle* path-planning, the agent can take any angle to move around in its environment. Some incremental planning work have also emerged for such environments. (Ferguson and Stentz, 2007) introduce Field D*, an adaptation of the D* algorithm for any angle path-planning, which reportedly returns a solution path often close to the optimal solution. Other works include Theta* from (Nash et al., 2007). Based on A*, Theta* is shown to give even shorter paths than Field D*, though not necessarily optimal either. However, Theta* lacks Field D*'s fast replanning capabilities. Finally, (Harabor et al., 2016) introduced ANYA, which they show to be significantly faster than previous approaches. Moreover, ANYA also guarantees to find optimal any-angle paths.

In this thesis, the typical graphs where an agent needs to solve path-planning problems are known in advance. Planning being done on a high-level scale (*i.e.* an edge linking two nodes in the graph represents a path between two positions which are geographically far), we assume that knowledge of the graph is accurate. For this reason, any changes observed by the agent as it moves in the graph will only concern low-level movements, and will have no impact on the optimality of the path computed on the graph. Incremental approaches are therefore not considered in the thesis.

## Anytime Planning

In some situations a path needs to be computed quickly. Such could be the case for example for an agent detecting possible obstruction on a planned path while in movement. A solution would be required by the agent as fast as possible to avoid having to come to a complete stop and waste time while re-computing a path. Computing a new optimal path can quickly become very hard, even for incremental algorithms if the number of search states required to be re-processed is high. In such a situation, it can be acceptable to compute a solution which is not guaranteed to be optimal very quickly first, so that the agent can keep moving. In the remaining time available (*e.g.* time for the agent to reach decisive points), the previously computed (likely sub-optimal) path can be improved. *Anytime* algorithms, sometimes referred to as *Hierarchical* path-planners, are designed to address that problem. They build a likely sub-optimal path very quickly and improve the path in the remaining time available.

There have been various works on anytime algorithms. Early works include (Zilberstein and Russell, 1995; Dean and Boddy, 1988). Then, (Likhachev, Gordon, and Thrun, 2003) introduced the well-known Anytime Repairing A* (ARA*). This algorithm is made of successive weighted A* searches. In a weighted A* search, the heuristic function $h$ used is multiplied by a factor $\epsilon > 1$. In doing so, substantial speedup is often provided at the cost of solution optimality. ARA* executes successive weighted A* searches with a decreasing inflation factor $\epsilon$, each of which uses information from previous searches and provides a sub-optimality bound. During each weighted A* search, ARA* considers only states whose costs at the previous search may not be valid anymore due to the new, lower $\epsilon$ value. Another anytime algorithm is the Anytime Weighted A* (AWA*) (Hansen and Zhou, 2007), which is very similar to ARA*. Authors show that AWA* is seven times faster than ARA* on certain domains such as the sliding-tile planning problem of eight puzzles.

From another perspective, (Likhachev et al., 2005) introduced Anytime Dynamic A* (AD*). Unlike previous approaches, AD* does not differentiate incremental and anytime approaches. Instead, it provides a framework which combines the benefits of both to provide solutions efficiently to hard dynamic problems. Experiments are carried out in an environment where a robotic arm is manipulating an end-effector through a dynamic environment and show AD* generating significantly better trajectories than ARA* and D* Lite in the same time budget. (Botea, Müller, and Schaeffer, 2004) presented Hierarchical Path-Finding A* (HPA*). HPA* proceeds to divide the environment into square clusters with connections, making an abstract search graph which is searched to find a shortest path. Another approach, Partial-Refinement A* (PRA*) (Sturtevant and Buro, 2005), builds cliques of nodes to construct a multi-level search space. The original problem is reduced to finding a set of nodes on the optimal shortest path. However, both HPA* and PRA* address homogenous agents in homogenous-terrain environments. An extension of HPA*, Annotated Hierarchical A* (AHA*), has been proposed by (Harabor and Botea, 2008). It is still one of the most advanced anytime path-planning algorithms to date. AHA* is able to deal with heterogeneous multi-terrain environments by reducing them to simpler single-size, single-terrain search problems. Authors' experiments suggest that near-optimal solutions are returned by the algorithm for problems in a wide range of environments, with an exponentially lower search effort than A*.

We do not use anytime planning algorithms in this thesis. The main reason being that, for the scenarios of the path-planning problem considered, the optimal path is usually computed in insignificant time with the exact approaches proposed in this thesis.

## Probabilistic Methods for Path-Planning

In high-dimensional search spaces, probabilistic approaches can provide a solution quickly but not necessarily an optimal one. We describe two popular approaches, Probabilistic Roadmaps (PRM) (Kavraki et al., 1996) and Rapidly-exploring Random Trees (RRT) (LaValle, 1998). The intuition behind PRMs is to generate random 'points' in the search space, connect these points to nearby points, and repeat the procedure until a path can be computed from the start state $s_{start}$ to the goal state $s_{goal}$ by moving along these points. More specifically, PRM starts by generating random states. It checks whether the generated states are valid, *i.e.* if they do not possess contradictory features (*e.g.* for the robot manipulator in 2.1, one would need to check if the combination of angles does not leave the robot arm in an impossible position). Invalid states are removed, and remaining states are named "milestones". Each milestone is connected to its $k$-nearest neighbor states, $k$ being a parameter. The process is repeated until the roadmap (the milestones and their connections) becomes dense enough and a connection between $s_{start}$ and $s_{goal}$ is created. A shortest path on the roadmap is then computed between $s_{start}$ and $s_{goal}$. PRM is *probabilistically complete*, *i.e.* as the roadmap building process goes on in time, the probability that the algorithm will find an existing path from $s_{start}$ to $s_{goal}$ tends to 1. Figure 2.3 illustrates a PRM. Notable follow-up works include Hierarchical PRMs (Collins, Agarwal, and Harer, 2003), which are a variant of PRMs refined recursively, providing better performance at finding narrow passages than uniform sampling. Other works have attempted to improve the efficiency of PRMs by altering the state sample generation process. Recently, (Kannan et al., 2016) built a PRM variant with adaptive sampling. They assign probabilities to different samplers dynamically based on the environment and use the one with the highest probability. (Ichter et al., 2020) proposed to learn to identify 'critical' states with a neural network from local environment features, *i.e.* states that are key to building the wanted path (*e.g.* doorways in an office environment). They draw these critical samples more often and thus are able to build a hierarchical roadmap more efficiently, with reportedly up to three order of magnitude improvements in computation time.

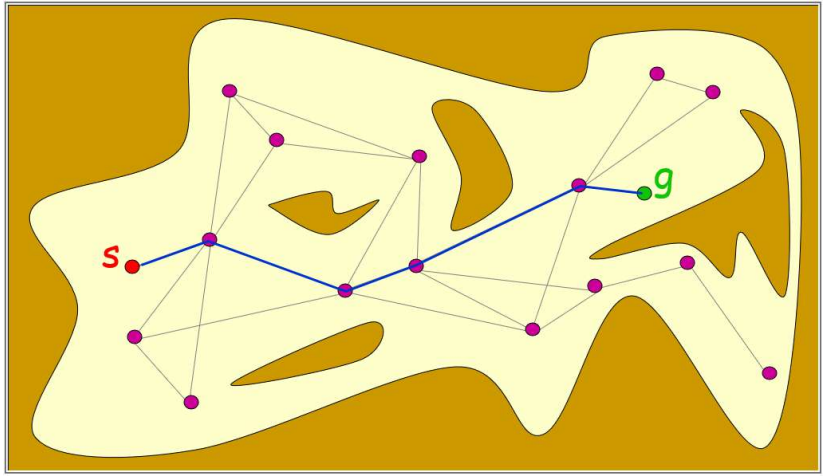

**Figure 2.3: A Probabilistic roadmap.** The white space represents feasible states, purple points milestones. Point 's' is the start state, point 'g' the goal state. The shortest path on the roadmap linking s to g is shown in blue. Credit for the picture goes to Jean-Claude Latombe. Source: (Latombe, 2020)

An RRT, on the other hand, starts growing a tree rapidly from the start state $s_{start}$, which is considered to be its root. To do so, RRT repeatedly uses a randomly sampled state $s_{rand}$ and attempts to connect $s_{rand}$ to any nearest state in the tree via feasible paths. When successful, RRT expands the size of the tree further with the addition of $s_{rand}$ and intermediary states found on the path. The sampling of random states is done in a way which expands the tree towards unsearched areas of the search space. Furthermore, for a randomly generated state $s_{rand}$ the length of its connection to the tree is limited by a growth factor. If the total length of the connection is above this distance, $s_{rand}$ is dropped and $s'_{rand}$, the state at the maximally allowed distance from the tree along the connection is selected instead. In this manner, the position of the randomly generated samples determines towards which areas the tree gets expanded, while the growth factor limits how far the tree is expanded in those directions. A drawback of RRTs is that they tend to often converge to non-optimal solutions. To address this issue, (Karaman and Frazzoli, 2010) introduced RRT*, which they showed to almost surely converge towards the optimal path without any significant overhead against RRT. We describe some notable follow-up works which are variants of RRT*. (Adiyatov and Varol, 2013) proposed a variant called RRT* Fixed Nodes (RRT*FN). Since there is no limit to the number of nodes RRT* can develop, the algorithm is not suited for embedded systems with limited memory. RRT*FN aims to solve the issue by using a node removal procedure which allows it to limit the number of nodes developed without hindering the convergence of the algorithm towards an optimal solution. (Gammell, Srinivasa, and Barfoot, 2014) proposed informed-RRT*, a variant which uses a heuristic to shrink the planning problem to subsets of the original domain. Informed-RRT* reportedly outperforms RRT* in rate of convergence, final solution cost, and ability to find difficult passages. More recently, (Lai, Ramos, and Francis, 2019) presented Rapidly-exploring Random disjointed-Trees* (RRdT). It is a RRT* variant which explores the search space with locally exploring disjointed trees and actively balances global exploration and local-connectivity exploitation. This is done by expressing the problem as a multi-armed bandit problem, and leads to improved performance.

For high-scale path-planning problems, probabilistic approaches were considered in this thesis but not retained due to the difficulty in coupling them with learning techniques. Instead, we preferred an approach which simplified the search space and made possible the use of a local search algorithm. The latter is easier to enhance with learning techniques and has the advantage of being simpler to implement.

## 2.3 Graph Representation Learning with Graph Neural Networks

### Neural Networks

We start by giving a brief description of neural networks and convolutional neural networks, architecture types from which graph neural networks originated.

Neural Networks (NNs) allow abstraction of data by using models with trainable parameters coupled with non-linear transformations of the input data. In spite of the complex structure of a NN, the main mechanism is straightforward. A *feedforward neural network*, or *Multi-Layer Perceptron (MLP)*, with $L$ layers describes a function $f_{\boldsymbol{\theta}}(\mathbf{x}) = f(\mathbf{x}; \boldsymbol{\theta}) : \mathbb{R}^{d_x} \to \mathbb{R}^{d_{\hat{y}}}$ that maps an input vector $\mathbf{x} \in \mathbb{R}^{d_x}$ to an output vector $\hat{\mathbf{y}} \in \mathbb{R}^{d_{\hat{y}}}$. Vector $\mathbf{x}$ is the input data that we need to analyze (*e.g.* an image, a signal, a graph, etc.), while $\hat{\mathbf{y}}$ is the expected decision from the NN (*e.g.* a class index, a heatmap, etc.). The function $f$ performs $L$ successive operations over the input $\mathbf{x}$:

$$h^{(l)} = f^{(l)}(h^{(l-1)}; \theta^{(l)}), \qquad l = 1, \dots, L \tag{2.1}$$

where $h^{(l)}$ is the hidden state of the network (*i.e.* features from intermediate layers, corresponding to intermediary values) and $f^{(l)}(h^{(l-1)}; \theta^{(l)}) : \mathbb{R}^{d_{l-1}} \mapsto \mathbb{R}^{d_l}$ is the mapping function performed at layer $l$; $h_0 = \mathbf{x}$. In other words:

$$f(\mathbf{x}) = f^{(L)}(f^{(L-1)}(\dots f^{(1)}(\mathbf{x}) \dots))$$

Each intermediate mapping depends on the output of the previous layer and on a set of trainable parameters $\theta^{(l)}$. We denote by $\boldsymbol{\theta} = \{\theta^{(1)}, \dots, \theta^{(L)}\}$ the entire set of parameters of the network. The intermediate functions $f^{(l)}(h^{(l-1)}; \theta^{(l)})$ have the form:

$$f^{(l)}(h^{(l-1)}; \theta^{(l)}) = \sigma\left(\theta^{(l)} h^{(l-1)} + b^{(l)}\right), \tag{2.2}$$

where $\theta^{(l)} \in \mathbb{R}^{d_l \times d_{l-1}}$ and $b^{(l)} \in \mathbb{R}^{d_l}$ are the trainable parameters and the bias, while $\sigma(\cdot)$ is an *activation* function, *i.e.* a function which is applied individually to each element of its input vector to introduce non-linearities. Intermediate layers are actually a combination of linear classifiers followed by a piecewise non-linearity. Layers with this form are termed *fully-connected layers*.

NNs are typically trained using labeled training data from a dataset, *i.e.* a set of input-output pairs $(\mathbf{x}_i, \mathbf{y}_i)$, $i = 1, \dots, N$, where $N$ is the size of the dataset. During training we aim to minimize the training loss:

$$\mathcal{L}(\boldsymbol{\theta}) = \frac{1}{N} \sum_{i=1}^{N} \ell(\hat{\mathbf{y}}_i, \mathbf{y}_i), \tag{2.3}$$

where $\hat{\mathbf{y}}_i = f(\mathbf{x}_i; \boldsymbol{\theta})$ is the estimation of $\mathbf{y}_i$ by the NN and $\ell : \mathbb{R}^{d_L} \times \mathbb{R}^{d_L} \mapsto \mathbb{R}$ is a loss function which measures the distance between the true label $\mathbf{y}_i$ and the estimated one $\hat{\mathbf{y}}_i$. Through *backpropagation*, the information from the loss is transmitted to all $\boldsymbol{\theta}$ and gradients of each $\theta_l$ are computed w.r.t. the loss. The optimal values of the parameters $\theta$ are then searched for via Stochastic Gradient Descent (SGD) which updates $\theta$ iteratively towards the minimization of $\mathcal{L}$. The input data is randomly grouped into mini-batches and parameters are updated after each pass. The entire dataset is passed through the network multiple times and the parameters are updated after each pass until reaching a satisfactory optimum. In this manner all the parameters of the NN are learned jointly and the pipeline allows the network to learn to extract features and to learn other more abstract features on top of the representations from lower layers.

In recent years, NNs, in particular Deep Neural Networks (DNNs), have achieved major breakthroughs in various areas. Some examples include image classification (Krizhevsky, Sutskever, and Hinton, 2012), (Simonyan and Zisserman, 2014), (He et al., 2016), object detection (Ren et al., 2015), (Redmon et al., 2016), (He et al., 2017), semantic segmentation (Long, Shelhamer, and Darrell, 2015), neural machine translation (Sutskever, Vinyals, and Le, 2014), computer games (Silver et al., 2016), (Silver et al., 2017) and many other fields. While the fundamental principles of training neural networks are known since many years, the recent improvements are due to a mix of availability of large datasets, advances in GPU-based computation and increased shared community effort. Similarly to NNs, DNNs enable a high number of levels of abstraction of data by using models with millions of trainable parameters coupled with non-linear transformations of the input data. It is known that a sufficiently large neural network can approximate any continuous function (Funahashi, 1989), although the cost of training such a network can be prohibitive.

Convolutional Neural Networks (CNNs) (Fukushima and Miyake, 1982; LeCun, Bengio, et al., 1995) are a generalization of multi-layer perceptrons for 2D data. In convolutional layers, groups of parameters (which can be seen as small fully-connected layers) are slided across an input vector similarly to filters in image processing. This reduces significantly the number of parameters of the network since they are now *shared* across locations, whereas in fully connected layers there is a parameter for element of the input. Since the convolutional units act locally, the input to the network can have a variable size. A convolutional layer is also a combination of linear classifiers (equation 2.2) and the output of such layer is 2D and is called *feature map*. CNNs are highly popular in most recent approaches for computer vision problems. Figure 2.4 shows a CNN architecture.

In this thesis, we deal with path-planning and scheduling problems in the forms of graph-structured data. To leverage MLPs for such problems, one would need to take into account the graph connections as input features, which would drastically increase the input size and decrease chances of learning useful information. As far as CNNs are concerned, representing the connections of a graph into an image adds

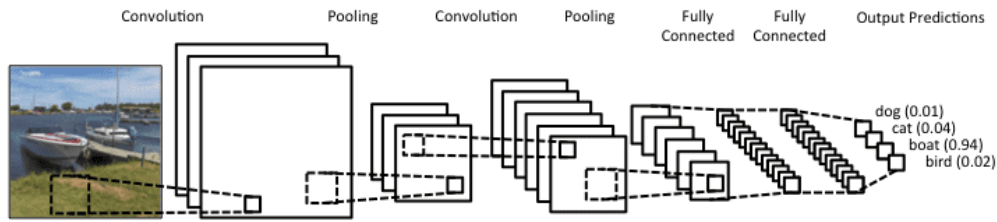

**Figure 2.4: A convolutional neural network.** The input of this CNN is an image, the output a prediction of a class among the following set of classes: {dog,cat,boat,bird}. Source: (*A CNN architecture*)

unnecessary information that will make the processing of interactions between nodes hard. Furthermore, a simple permutation of nodes (*i.e.* changing the position of one node with another), which has no effect on the considered graph problem, will cause the CNN to behave differently. Therefore, we explore graph neural networks in the next section. These networks have the advantage of being invariant to node permutations and incorporate node and edge features for better inference.

## Graph Neural Networks

In this section, we discuss various architectures of Graph Neural Networks (GNNs). GNNs are generalizations of CNNs to non-Euclidean data and aim to learn graph representations. Although no common groups have been precisely defined for GNNs, they tend to belong to four categories:

- **Converging Recurrent Graph Neural Networks (CRGNN)**. These architectures of neural networks mostly include the first works on extending NNs to graphs.

- **Graph Convolutional Networks (GCN)**. These networks are mostly inspired from the application of CNNs to graphs and are well-suited to supervised-learning for node classification.

- **Recurrent Graph Neural Networks (RGNN)**. RGNNs are designed to process input graphs for which a temporal sequence ordering exists. They should not be confused with CRGNNs, which are 'recurrent' in the sense they apply a process on the input graph repeatedly until convergence.

- **Graph Autoencoder Networks (GAN)**. GANs use an *encoder* to encode input graphs into a latent vector space and a *decoder* to decode them back. We skip

GANs in this section since they are mostly used for unsupervised learning tasks, which is out of the scope of this thesis.

These networks have different processing architectures but have the following common point. They take as input a graph in which nodes, and possibly edges, have features. They use intermediary layers, each of which produces new features for nodes (and possibly edges, depending on the architecture type). Let $\mathcal{G} = (\mathcal{V}, \mathcal{E})$ be an input graph of these network architectures with its set of nodes $\mathcal{V} = (v_1, v_2, ..., v_n)$ and its set of edges $\mathcal{E}$. We denote as $A$ the adjacency matrix of $\mathcal{G}$, $X = (x_{v_1}, x_{v_2}, ..., x_{v_n})$ the matrix of node feature vectors of the input graph $\mathcal{G}$ and $H^{(l)} = (h_{v_1}^{(l)}, h_{v_2}^{(l)}, ..., h_{v_n}^{(l)})$ the matrix of node feature vectors after the input graph $\mathcal{G}$ has been processed by $l$ layers. Each $x_{v_i}$ is the feature vector of node $v_i$ of the input graph $\mathcal{G}$ and each $h_{v_i}^{(l)}$ the feature vector of node $v_i$ after the $l^{th}$ layer. Finally, as the layer architecture types we describe next apply a similar process to each graph node, the same layer can be used on input graphs with any number of nodes, although the number of features per node needs to be fixed. In other words, a same GNN with layers made of these architectures can process input graphs with any number of node.

**Converging Recurrent Graph Neural Networks**

The idea behind CRGNNs was initially introduced in (Sperduti and Starita, 1997), with a contribution termed *generalized recursive neuron*, extending the idea of applying neural networks to inputs with structures. Those structures were essentially limited to acyclic graphs because of computational constraints at the time. In follow-up works, (Scarselli et al., 2008) extend this with an architecture capable of processing acyclic, cyclic, directed and undirected graphs. To that end, neighborhood information among graph nodes is exchanged repeatedly until convergence. The following formula describes how information is updated from layer $l$ to layer $l + 1$ for node $v$:

$$h_v^{(l)} = \sum_{w \in N(v)} f(x_v, x_{e_{vw}}, x_w, h_w^{(l-1)}) \tag{2.4}$$

where:

- $h_v^{(l)}$ and $h_v^{(l-1)}$ respectively designate the vector feature of node $v$ after layer $l$ and layer $l - 1$; $h_v^{(0)} = x_v$.

- $N(v)$ designates the nodes connected to node $v$ with an edge.

- $x_{e_{vw}}$ is the feature vector of the edge connecting $v$ and $w$.

- $f$ is a parametric function, called *local transition function* by Scarselli *et al.*

Intuitively, the information update of node $v$ from a layer $l-1$ to a layer $l$ proceeds in the following manner for each node. For each neighboring node $w$, a parametric function $f$ takes as input the following elements: the input feature vectors of node $v$, edge $(v, w)$, node $w$, as well as the feature vector of node $w$ after layer $l-1$. The sum of the output of $f$ for each neighbor of $v$ makes the new feature vector of node $v$ after layer $l$. Moreover, to ensure convergence after applying layers repeatedly, function $f$ needs to be a *contraction map* which reduces the distance between inputs and satisfies this property:

$$\forall z \in \mathbb{R}^m \ \exists \mu \in ]0, 1[ \ s.t \ \forall (x, y) \in \mathbb{R}^{m' \times m''} : \ \|f(x, z) - f(y, z)\| \leq \mu \|x - y\|$$

where $\| \cdot \|$ denotes a vectorial norm.

A convergence criterion also needs to be defined. Layers are applied recursively on each node in parallel until this criterion is satisfied. The converged node feature vectors $h_{v_i}^*$ of each node $v_i$ can then be forwarded to an output layer to perform either node classification tasks, edge classification tasks (by using for example a MLP which takes as input converged features $h_{v_i}^*$ and $h_{v_j}^*$ and outputs a value for edge $(v_i, v_j)$) or graph-level predictions (*e.g.* predict a class among a portfolio of classes for the input graph, which is typically done by using one or multiple *pooling* operations such as *max* or *min* to reduce the size of the converged graph into a fixed size, enabling the use, for example, of a fully-connected output layer).

A notable issue with this CRGNN architecture is the number of layers which need to be applied to meet the convergence criterion and the possibly ensuing complexity. More recently, a framework was proposed in (Li et al., 2015) to address this issue based on gated recurrent units (Cho et al., 2014). This allows Li *et al.* to only require a fixed number of layers to process an input graph, thereby lifting the constraints associated with the convergence criterion. Nevertheless, the approach in (Li et al., 2015) requires Back-Propagation Through Time (BPTT) to compute gradients when using the model in a loss function, which can cause severe overhead.

CRGNNs are mostly pioneer works which inspired the next architectures we describe, and even the newest CRGNN approach presents computational issues due to BPTT. Therefore we do not retain this architecture type for the works carried out in this thesis.

**Graph Convolutional Networks**

Unlike CRGNNs where a fixed recurrent model is applied repeatedly, GCNs use a fixed number of graph convolutional layers, each of which is different and has its own set of trainable parameters. GCNs are inspired from CNNs. They generalize their operations from grid-structured data (images) to graph-structured data. There are two main categories of GCNs: *spectral-based* and *spatial-based*. Spectral-based approaches use signal processing to define the neighborhood of a node and the ensuing feature update process, while spatial-based approaches rely directly on spatially close neighbors in the graph.

**Spectral-based GCNs**   Spectral-based architectures use the spectral representation of graphs and are thus limited to undirected graphs. They were introduced in (Bruna et al., 2014). The following layer propagation rule is used to compute $H^{(l)}$, the matrix of all node feature vectors at layer $l$, from $H^{(l-1)}$:

$$H^{(l)} = \sigma(U g_\theta(\Lambda) U^T H^{(l-1)}) \qquad (2.5)$$

Here, $U$ denotes the eigenvectors of the normalized graph Laplacian matrix $L = I_N - D^{-\frac{1}{2}} A D^{-\frac{1}{2}}$ ($A$ being the adjacency matrix, $D$ the node degree matrix) and $\Lambda$ its eigenvalues. Function $g_\theta(\Lambda) = diag_\theta(\Lambda)$ is a filter applied on the eigenvalues with a set of parameters $\theta$. Lastly, $\sigma$ is an activation function. A problem with this approach is that it results in non-spatially localized filters, making it unable to extract local features independently of graph size. In a follow-up work, (Defferrard, Bresson, and Vandergheynst, 2016) introduce *ChebNet*. The filters proposed in ChebNet are localized in space. Their idea is to replace $g_\theta(\Lambda)$ with a truncation of Chebyshev polynomials $T_k$ of the eigenvalues $\Lambda$: $g_\theta(\Lambda) = \sum_{k=0}^{K} \theta_k T_k(\tilde{\Lambda})$ where $\tilde{\Lambda} = \frac{2\Lambda}{\lambda_{max}} - I_n$, $\theta_k \in \mathbb{R}^K$ and $\lambda_{max}$ denotes the highest eigenvalue. Chebyshev polynomials are recursively defined by: $T_k(x) = 2x T_{k-1}(x) - T_{k-2}(x)$; $T_0(x) = 1$; $T_1(x) = x$. After further simplifications, the layer propagation rule is simplified to:

$$H^{(l)} = \sigma(\sum_{k=0}^{K} \theta_k T_k(\tilde{L}) H^{(l-1)}) \qquad (2.6)$$

where $\tilde{L} = \frac{2L}{\lambda_{max}} - I_n$. More recently, in (Kipf and Welling, 2017), authors introduce *GCN* by applying a first-order approximation of ChebNet ($K = 1$, and $\lambda_{max} = 2$). This enables them to avoid overfitting local neighborhood structures on graphs with unbalanced node degree distributions. Equation 2.6 becomes:

$$H^{(l)} = \sigma(\theta_0 H^{(l-1)} - \theta_1 D^{-\frac{1}{2}} A D^{-\frac{1}{2}} H^{(l-1)}) \tag{2.7}$$

An additional assumption is made in GCN that $\theta = \theta_0 = -\theta_1$ to further reduce overfitting, and the equation becomes:

$$H^{(l)} = \sigma(\theta(I_n + D^{-\frac{1}{2}} A D^{-\frac{1}{2}}) H^{(l-1)}) \tag{2.8}$$

Finally, a *re-normalization trick* is used to avoid numerical instabilities such as exploding or vanishing gradients: $I_n + D^{-\frac{1}{2}} A D^{-\frac{1}{2}} \rightarrow \tilde{D}^{-\frac{1}{2}} \tilde{A} \tilde{D}^{-\frac{1}{2}}$, with $\tilde{A} = A + I_n$ and $\tilde{D}$ being the degree matrix of $\tilde{A}$. Kipf and Welling generalize this definition to an input $H^{(l-1)} \in \mathbb{R}^{N \times C}$ where $C$ is the number of features per node at layer $l - 1$, $N$ the number of nodes in the input graph. Moreover, They use a weight matrix $W \in \mathbb{R}^{C \times F}$, where $F$ is the desired number of features per node after the layer has been applied. The equation becomes:

$$H^{(l)} = \sigma(\tilde{D}^{-\frac{1}{2}} \tilde{A} \tilde{D}^{-\frac{1}{2}} H^{(l-1)} W) \tag{2.9}$$

On a side note, during the information update for a node $v$, GCN takes a weighted sum of vector features from neighbors, where the weight for a neighbor $w$ is given by: $\frac{1}{\sqrt{deg(v) \times deg(w)}}$, where $deg(v)$ refers to the degree of node $v$. Several linear combinations are then applied, to create as many output features as needed for $v$ in the next layer.

Lastly, methods presented thus far rely on the adjacency matrix to define relations between nodes, possibly missing on implicit information between nodes. Authors in (Li and Wang, 2018) propose Adaptive Graph Convolutional Network (AGCN) to address this issue. AGCN basically learns a *residual* graph adjacency matrix by learning a distance function which takes as input the features of two different nodes in the graph, enabling it to better capture implicit dependencies.

**Spatial-based GCNs**  Spatial-based approaches rely on spatially close neighbors to define the feature update step for a node. In this sense, spatial-based GCNs are somewhat similar to CRGNNs in that they propagate node information through edges, although they do not retain the idea of convergence and they stack multiple different layers with different trainable weights. A significant advantage of spatial-based GCNs over spectral-based GCNs is that they can be used on directed graphs.

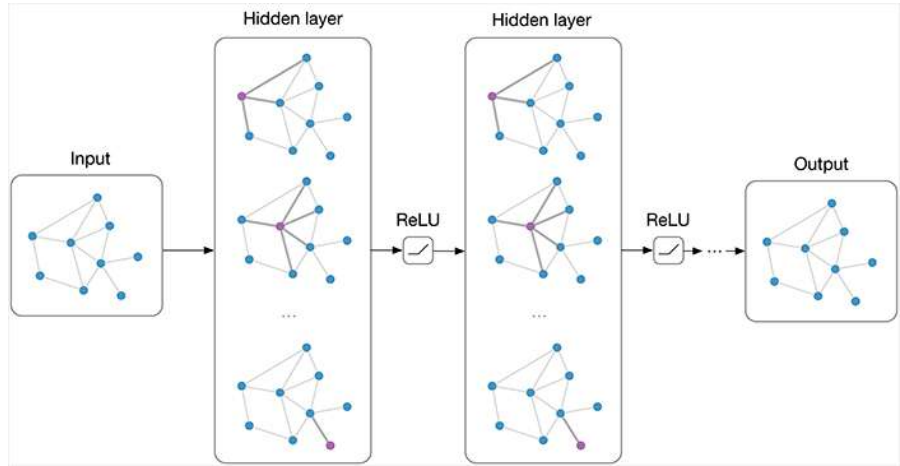

**Figure 2.5: A graph convolutional network.** In this illustration, an input graph with node features (and possibly edge features) is processed through multiple graph convolutional layers and $ReLU(\cdot) = max(0, \cdot)$ nonlinearities. An output graph is returned with new, updated node features. Credit goes to Thomas Kipf for the illustration, source: (*A GCN architecture*)

Among early spatial-based architectures, (Micheli, 2009) introduces neural network for graphs (NN4G). In the NN4G architecture, graph convolutions are performed at each layer (each of which has its own trainable weights). Each convolution basically consists in the sum, for each node, of the feature vectors of neighboring nodes. In this sense, it is somewhat similar to the GCN architecture of (Kipf and Welling, 2017) which performs a weighted sum based on the spectral graph instead. Additionally, NN4G applies residual skip connections between each layer to 'memorize' information. Each new layer is basically linked not only to the previous one, but also to all preceding layers and the input. The following equation defines NN4G's propagation rule, where $\Theta^{(l)}$ and $W^{(k)}$ are weight matrices:

$$H^{(l)} = \sigma(X\Theta^{(l)} + \sum_{k=1}^{l-1} AH^{(k)}W^{(k)})$$

(2.10)

The Diffusion Convolutional Neural Networks (DCNNs) proposed in (Atwood and Towsley, 2016) brings the concept of diffusion to graphs convolutions. A transition probability is defined when information from a node is passed to a neighboring node, causing the passing of information to converge after applying the process repeatedly. Transition matrices are used to define the neighborhood for a node. The propagation rule for DCNN is:

$$H^{(l)} = \sigma(W^{(l)} \odot P^l H^{(l-1)})$$ (2.11)

where $W^{(l)}$ is a weight matrix, $\odot$ denotes the element-wise product, $P^l$ (not to be confused with $P^{(l)}$) is $P$ to the power of $l$, with $P = D^{-1}A$ the probability transition matrix.

Message Passing Neural Networks (MPNN), on the other hand, are a general framework presented in (Gilmer et al., 2017) which aim to regroup different categories of previous works into one single architecture. In MPNNs, during the convolution phase of an input graph, messages are passed between nodes along edges by following an aggregation phase, called *message passing* phase, after which node features get updated in a *message update* phase. Each node $v$ has its feature vector $h_v^{(l-1)}$ updated to $h_v^{(l)}$ based on a message $m_v^{(l)}$:

$$m_v^{(l)} = \sum_{w \in N(v)} M_{l-1}(h_v^{(l-1)}, h_w^{(l-1)}, x_{e_{vw}})$$ (2.12)

$$h_v^{(l)} = U_{l-1}(h_v^{(l-1)}, m_v^{(l)})$$ (2.13)

where $N(v)$ designates the neighborhood of node $v$; $M_{l-1}$ and $U_{l-1}$ are learned differentiable functions; $x_{e_{vw}}$ is the vector feature of the edge connecting $v$ and $w$.

A *readout phase* is also introduced (after the last message passing layer $l_{max}$ has been applied), in which a readout function $R$ can optionally compute a feature vector $\hat{y}$ for the whole graph (assuming we want to do some other type of classification than node classification, such as graph-level class prediction):

$$\hat{y} = R(\{h_v^{(l_{max})} | v \in \mathcal{V}\})$$ (2.14)

$R$ needs to be invariant to the permutation of node states in order for the MPNN to retain invariance to graph isomorphism. Gilmer *et al.* proceed to express previous existing GNN architectures in the literature by specifying the corresponding message passing function $M_{l-1}$, message update function $U_{l-1}$ and readout function $R$. In their own work, they use an architecture in which $M(h_v, h_w, x_{e_{vw}}) = MLP(x_{e_{vw}})h_w$. Here, MLP is a multi-layer perceptron which takes as input the feature vector of edge $(v, w)$ and outputs a $out_c \times in_c$ sized-matrix, $out_c$ being the number of desired feature per node after applying the message passing layer and $in_c$ the number of feature per node of the input graph provided to the layer. Vector $h_w$ being of size $in_c \times 1$, the matrix multiplication results in a $out_c \times 1$ sized-matrix, *i.e.* a vector which has the desired number of new features after the message passing layer is applied. The sum of these vectors for the entire neighborhood defines $m_v$. Gilmer *et al.* apply this architecture for node classification tasks on a molecular property prediction benchmark and achieve state-of-the-art results.

Other recent relevant works include GraphSAGE (Hamilton, Ying, and Leskovec, 2017) and Graph Attention Networks (GATs) (Veličković et al., 2018). GraphSAGE has been conceived to handle graphs where the number of neighbors for nodes can vary greatly from one node to another. Since always taking into account the entire neighborhood can prove inefficient and costly, graphSAGE uses sampling to define neighborhoods and thus keep a fixed number of neighbors for each node. The propagation rule in a graphSAGE convolution is defined by:

$$h_v^{(l)} = \sigma[W^{(l)} AGG_l(\{h_v^{(l-1)}\} \cup \{h_u^{(l-1)}, \forall u \in N_r(v)\})] \tag{2.15}$$

where: $N_r(v)$ designates a fixed-size uniform draw from the set $\{u \in \mathcal{V} : (u,v) \in \mathcal{E}\}$ and $AGG_l$ is an aggregation function invariant to the permutations of node orderings (*e.g.* mean function). GATs, on the other hand, use an *attention* mechanism which defines weights for each connected pair of nodes. Weights are learned by the attention mechanism so as to reflect the importance of each neighbor of a node $v$. The layer propagation rule is defined by:

$$h_v^{(l)} = \sigma(W^{(l)} \sum_{u \in \{v\} \cup N(v)} \alpha_{uv}^{(l)} h_u^{(l-1)}) \tag{2.16}$$

where $N(v)$ refers to the neighborhood of $v$, and $\alpha_{uv}^{(l)}$ is the attention weight. This attention weight is defined by the following query-key mechanism:

$$\alpha_{uv}^{(l)} = \frac{exp(LeakyReLU(a^T[W^{(l)}h_v^{(l-1)}||W^{(l)}h_u^{(l-1)}]))}{\sum_{q \in N(v)} exp(LeakyReLU(a^T[W^{(l)}h_v^{(l-1)}||W^{(l)}h_q^{(l-1)}]))} \tag{2.17}$$

where: $LeakyReLU(\cdot) = \cdot$ if $\cdot > 0$, $\alpha \times \cdot$ otherwise ($\alpha$ is a small number); $a$ is a vector of learnable weights, $||$ denotes the concatenation operation and $N(v)$ refers to the neighborhood of node $v$. Additionally, GAT can use multi-head attention mechanisms (*i.e.* have multiple attention heads $\alpha_{uv}^{(l)}$, $\alpha_{uv}^{(l)'}$, $\alpha_{uv}^{(l)''}$, etc...). This enables the model to learn different attention schemes in parallel per layer, and shows considerable improvement over GraphSAGE on node classification benchmarks.

In this thesis, we use two architectures of GNNs. The first one is the spectral-based GCN defined in Equation 2.9. We refer to it as 'GCN' in the following sections, and use it for path-planning problems. We perform learning and predictions on a same graph for such problems (but with different input node features), therefore the lightweight GCN architecture is efficient and well-suited, and the lack of edge feature consideration is not a problem since edge weights are learned implicitly. The second architecture type we use is the MPNN architecture defined in Equation 2.12 and Equation 2.13, and we refer to it as 'MPNN' in the following sections. MPNNs are more complex than GCNs in that feature aggregation is spatial and based on MLPs instead of non-parametric weights defined by the graph Laplacian. This gives them the advantage of taking into account edge features in addition to node features. We use MPNNs to process different graphs with node and edge features, and apply them to scheduling problems, each of which is expressed as graphs.

**Recurrent Graph Neural Networks**

In many applications, graphs can not only present spatial structure, but also hold temporal dependencies. An example is road network traffic, for which the same graph at different time steps is going to represent the current flow of traffic in the network. RGNNs are inspired by Recurrent Neural Networks (RNNs) and aim to process a sequence of temporal graphs, in order for example to make predictions about future states (*e.g.* how traffic is going to be like in future time steps). For most RGNNs, an RNN-like mechanism is used to memorize and leverage temporal information. Nevertheless, some RGNNs use CNNs to capture temporal information instead. We first describe some RNN-based methods and then some CNN-based approaches.

The idea behind RNN-based RGNNs stems from the recurrent units used in RNNs. When a RNN is used on an input at time step $t$, each hidden layer $h^{(l)^t}$ is computed by combining both the input to the layer $h^{(l-1)^t}$, as well as a 'memory' equal to the output of the same layer at time step $t - 1 : h^{(l)^{t-1}}$. In RNN-based RGNNs, the equation for the layer propagation rule is of the form:

$$H^{(l)^t} = \sigma(graphConv(H^{(l-1)^t}, W^{(l)}, \mathcal{G}) + graphConv(H^{(l)^{t-1}}, \hat{W}^{(l)}, \mathcal{G}) + B^{(l)}) \quad (2.18)$$

Here, *graphConv* denotes a graph convolution operation (either spatial-based or spectral-based), $\hat{W}^{(l)}$ is a weight matrix different from $W^{(l)}$, and $B^{(l)}$ is a bias. The second *graphConv* operation aims to introduce to the computation the memory which was retained in the previous step.

Main such works include Structural-RNN (S-RNN) (Jain et al., 2016). S-RNN uses different RNNs to handle both node and edge information, namely *nodeRNN* and *edgeRNN*. Diffusion Convolutional Recurrent Neural Network (DCRNN) (Li et al., 2017) is an encoder-decoder framework which applies gated recurrent units on the DCNN architecture. In (Seo et al., 2018), a Long Short-Term Memory (LSTM) network is combined with the ChebNet graph convolution operator. LSTMs are a popular type of RNN architecture because they are able to maintain a longer memory than RNNs.

CNN-based RGNNs, on the other hand, abandon the idea of keeping a memory and instead use a CNN jointly with a graph convolution operator to capture temporal and spatial information at the same time. Their advantage over RNN-based RGNNs is that they do not require backpropagation through time for gradient computation. The idea is that for each node $v$ in the input graph, a 1D-CNN is applied and temporal information from previous states of the node is aggregated. Next, a graph convolutional layer is applied on the aggregated temporal information to aggregate spatial information. This process is repeated for each layer.

(Yu, Yin, and Zhu, 2017) propose Spatio-Temporal Graph Convolutional Networks (STGCN), which uses a 1D-CNN alongside the ChebNet convolutional layer. Graph WaveNet (Wu et al., 2019) introduces a framework with a self-adaptive adjacency matrix. This allows Graph WaveNet to learn latent structures, which can help discover implicit temporal dependencies between nodes in the graph. Lastly, (Guo et al., 2019) introduce an Attention based Spatial-Temporal Graph Convolutional Network (ASTGCN) to solve traffic flow forecasting problems. ASTGCN builds on STGCN by introducing attention mechanisms both for spatial and temporal aggregation. This allows ASTGCN to outperform state-of-the-art baselines on real-world datasets from the Caltrans performance measurement system.

We do not retain the RGNN architecture for this thesis. The reason being that for the path-problems and scheduling problems we consider, we are able to decompose these problems into sub-problems recursively. Moreover, to solve these sub-problems, the optimal choices to make are not influenced by decisions made in parent problems. This makes it uninteresting to take into account the temporal aspect.

## 2.4  Reinforcement Learning

Reinforcement Learning (RL) consists in designing an agent capable of learning through trial and error by interacting with an environment. This section only aims to briefly describe Markov Decision Processes (MDP) (Bellman, 1957) and RL concepts. We refer the reader to (Sutton and Barto, 2018) for a complete introduction to RL. We temporarily use the following notations here, not to be confused with notations from the previous section:

- Set $S$: a set of states.

- Set $A$: a set of actions.

- Set $P$: a set of transition probabilities. The probability $P(s'|s,a) = P_a(s,s')$ refers to the probability of transitioning from state $s \in S$ to state $s' \in S$ after taking action $a \in A$.

- Function $R$: a reward function. The transition from state $s \in S$ to state $s' \in S$ after taking action $a \in A$ results in an immediate reward $R(s'|s,a) = R_a(s,s')$.

In MDPs, the environment is fully observable and actions are instantaneous and non-deterministic. Nevertheless $\forall (s,a) \in (S,A)$, $\exists! P_a(s,s')$. In other words, after taking action $a \in A$ in the state $s \in S$, a given set of probabilities exist for each state $s' \in S$ which defines the likelihood of transitioning into those states. Moreover, an *immediate reward* function $R_a(s,s')$ defines a given reward obtained from transitioning to a state $s' \in S$ after taking the action $a \in A$ in state $s \in S$. The aim for the agent is to devise an optimal *policy* $\pi^*$ which specifies which action $\pi^*(s) \in A$ to take in any state $s$ in order to maximize the total cumulative reward.

RL can generally be formulated as a 4-tuple $(S,A,P,R)$, representing an agent interacting with the environment in a MDP. The agent interacts with the environment by following a policy $\pi$, and the goal is to find an optimal policy $\pi^*$ by trial and error. Two main RL approaches exist: policy gradient optimization and value function optimization. In policy gradient approaches, a parameterizable function $f_\theta$ ($\theta$ being parameters) is used to approximate $\pi$ directly. Through interaction with the environment, the agent is able to learn, given its current policy $f_\theta$, which actions are more suited for given states in $S$. Thus, the agent can modify its policy $f_\theta$ to prioritize these actions. A popular choice for the function $f_\theta$ is neural networks, whose number of layers can be chosen according to the assumed complexity of the function approximated. On the other hand, value function optimization learns two different value functions $Q$ and $V$, which define the policy $\pi$ to follow. The state value function $V$ is defined by: $V^\pi(s) = \mathbb{E}_\pi(\sum_{i=0}^\infty \gamma^i r_{i+1}|s_t = s)$, where $s_t$ refers to the state of the agent at the current time step, $s_{t+1}, s_{t+2}, ...$ at future time steps; $r_{i+1}$ refers to the immediate reward received by the agent at time step $t+i+1$; variable $\gamma \in ]0,1]$ is a discount factor and $\mathbb{E}$ denotes the expectation. Intuitively, $V^\pi(s)$ corresponds to the expected sum of rewards when starting in state $(s)$ and following policy $\pi$. The action value function is defined by $Q^\pi(s, a) = \mathbb{E}_\pi(\sum_{i=0}^\infty \gamma^i r_{i+1} | s_t = s, a_t = a)$ where $a_t$ refers to the action taken by the agent at the current starting time step. Intuitively, it corresponds to the expected sum of rewards when starting in state $(s)$, taking action $a$ and following policy $\pi$ afterwards. In environments with large and continuous state spaces, these functions are usually approximated using neural networks.

RL methods also belong to two categories: model-based and model-free. Model-based assumes knowledge of the transition probabilities in the MDP environment, while model-free does not. A popular approach for model-based is value iteration, which consists in updating Q-values by taking into account transition probabilities and known knowledge about transition states. Q-learning (Watkins and Dayan, 1992) is a popular approach for model-free approaches. It follows the idea of 'pulling' a given Q-value toward the result obtained from a simulation with the environment every time with a learning rate, so as to approximate transition probabilities indirectly. Q-learning follows this update scheme:

$$Q(s_t, a_t) \leftarrow Q(s_t, a_t) + \alpha [r_t + \gamma \max_{a_i} Q(s_{t+1}, a_i) - Q(s_t, a_t)]$$

where $\alpha$ is the learning rate; $\delta_t = r_t + \gamma \max_{a_i} Q(s_{t+1}, a_i) - Q(s_t, a_t)$ is called the temporal difference.

The use of Deep Q Neural Networks (DQN) (Mnih et al., 2013; Mnih et al., 2015) has allowed RL tasks to achieve human level gameplay on games from the Atari 2600 plaform. More recently, AlphaGo (Silver et al., 2016), an AI program conceived to play the game of Go combining deep CNNs and Monte Carlo Tree Search (MCTS), managed to defeat the world champion of Go. AlphaGo uses supervised learning to learn from expert gameplay, and then refines the learned policy with RL (policy gradient) by playing against itself. AlphaGoZero (Silver et al., 2017), a newer version, is only trained with RL and achieves superior gaming performance than AlphaGo.

Model-based RL algorithms are usually more efficient than model-free ones since they can leverage planning using known environment dynamics. A solution for model-free approaches would be to learn the dynamics from interactions with the environment. Although learning dynamics which are accurate enough for planning has remained a challenge in model-free approaches, the PlaNet approach from (Hafner et al., 2019) achieves a breakthrough on this subject for image-based domains. PlaNet learns the dynamics model by relying on a sequence of latent states generated by an encoder-decoder architecture, rather than images directly. PlaNet chooses actions purely by planning in this latent space, this allows it to require far lower interaction with the environment to optimize its policy than previous recent approaches in model-free RL.

Lastly, in some situations, the environment may not be fully observable by the agent. This is the case in Partially Observable Markov Decision Processes (POMDP). Agents get sensory information and derive a probability distribution of states they may likely be in, and need to adapt their policy accordingly. Some popular works dealing with planning in POMDP include (Kurniawati, Hsu, and Lee, 2008) who introduce a point-based POMDP algorithm for motion-planning, (Silver and Veness, 2010) who propose an MCTS algorithm for planning in large POMDPs, (Somani et al., 2013) who present a random scenario sampling scheme to alleviate computational limitations and (Zhu et al., 2017) who propose a Deep Recurrent Q-Network to adapt RL tasks in POMDPs.

In this thesis, we explore model-based value iteration and policy gradient algorithms to train a graph neural network to learn problem topology. We do not retain the Q-learning approach for our path-planning applications since decision effects are assumed to be deterministic and related transitions are fully controlled.

## 2.5 Path-Planning and Neural Networks

A*-based algorithms described previously are fast on small planning domains, but take exponentially longer as domain size and complexity grows. Probabilistic approaches such as PRMs and RRTs on the other hand construct a new graph with random sampling to bypass this complexity, but to guarantee consistent solution quality the sampling would need to be exponential again (LaValle, Branicky, and Lindemann, 2004). Therefore, the idea of using neural networks for path-planning has long been a problem of interest, although recent advances in machine learning has made it a viable option only recently. We explore a few such works.

(Glasius, Komoda, and Gielen, 1995) is an early work which specifies obstacles into a topologically ordered neural map, and uses a neural network to trace a shortest path. The minimum of a Lyapunov function is used for convergence for neural activity. (Chen, Karl, and Van Der Smagt, 2016), a more recent work, relies on *Deep Variational Bayes Filtering* (DVBF) (Karl et al., 2016) to embed dynamic movement primitives of a high dimensional humanoid movements in the latent space of a low dimensional variational autoencoder framework. RL has also been used for such purposes. (Levine and Koltun, 2013) present a guided policy search algorithm that uses trajectory optimization to direct policy learning and avoid poor local optima, where policies are approximated by neural networks. This method is successfully applied to learn policies for planar swimming, hopping, walking and simulated 3D humanoid running. (Tamar et al., 2016) introduce a neural network to approximate the value iteration algorithm in order to predict outcomes that involve planning-based reasoning. Their use of CNNs limits their approach to path-planning on 2D grids and not motion planning in general.

Some *imitation learning*-based approaches have also been proposed. Imitation learning consists in having an expert provide demonstrations, in this case of desired trajectories. A neural network can then be used to approximate the behavior of the expert, and hopefully generalize outside of the scope of provided demonstrations. Imitation learning has been successful in several areas involving complex dynamical systems (Abbeel, Coates, and Ng, 2010; Calinon et al., 2010). OracleNet, an extension of imitation learning for path-planning has been proposed recently in (Bency, Qureshi, and Yip, 2019). OracleNet relies on an LSTM to build end-to-end trajectories in an iterative manner. The LSTM needs to be trained on optimal trajectories that span the entire configuration space of the considered environment before being used. Those optimal trajectories can be computed by algorithms such as A*. Although the proposed approach can be problematic if the framework needs to be quickly used in a newly known environment and no training time is available, OracleNet achieves performance which makes up for it. Paths are generated extremely fast, scaling almost linearly with dimensions reportedly. On a benchmark comprised of a point-mass robot with multiple degrees of freedom, OracleNet is compared to A* and RRT*. It achieves solution quality reportedly rivaling A* and far above RRT*, while its execution time remains far below the other two.

The idea behind path-planning approaches presented in this thesis are comparable to OracleNet, although in this thesis an end-to-end learned model architecture is not privileged. Rather, we aim to provide a path-planner with useful information from a neural network so as to increase its computational efficiency. This way of processing presents the advantage over end-to-end architectures of offering the guarantees proposed by the path-planner (*e.g.* optimality guarantees for optimal path-planners).

## 2.6 Temporal Planning With Uncertainty

Scheduling in the presence of uncertainty is an area of interest in artificial intelligence. In this section, we present necessary notions and work leading up to the Disjunctive

Temporal Network with Uncertainty (DTNU).

Temporal Networks (Dechter, Meiri, and Pearl, 1991) are a common formalism to represent temporal constraints over a set of timepoints (*e.g.* start/end of activities in a scheduling problem). A Simple Temporal Network (STN) $\Gamma$ is defined by a pair:

$$\Gamma = (A, C)$$

Where:

- $A = (a_1, a_2, ..., a_n) \in \mathbb{R}^n$ is a set of $n$ real controllable timepoint variables.

- $C$ is a set of *free* constraints, each of which of the form: $a_j - a_i \in [x_k, y_k]$, where $a_i, a_j \in A$; $x_k \in \{-\infty\} \cup \mathbb{R}$; $y_k \in \mathbb{R} \cup \{+\infty\}$.

A solution to STN $\Gamma$ is a complete set of assignments in $\mathbb{R}$ for each $a_i \in A$ which satisfies all constraints in $C$.

The Simple Temporal Networks with Uncertainty (STNUs) (Tsamardinos, 2002; Vidal, 1999) explicitly incorporate qualitative uncertainty into temporal networks. In STNUs, some events are *uncontrollable*. The only controllable aspect is when they start: how long they take to complete, however, is not known. Although the duration for completion is uncertain, it is often known to be within some bounds. These uncontrollable events are represented by a *contingency link*, *i.e.* a triplet $(a, [x, y], u)$, where $a$ is a controllable timepoint (representing the start of the uncontrollable event), $[x, y]$ is the bounded duration of the uncontrollable event and $u$ is an uncontrollable timepoint which signifies the end of the uncontrollable event. Uncontrollable timepoint $u$ will occur on its own, at earliest $x$ units of time after execution of $a$, $y$ at latest.

Formally, an STNU $\Gamma$ is defined as :

$$\Gamma = (A, U, C, L)$$

Where:

- $A = (a_1, a_2, ..., a_n) \in \mathbb{R}^n$ is a set of real controllable timepoint variables, which can be scheduled at any moment in time.

- $U = (u_1, u_2, ..., u_q) \in \mathbb{R}^q$ is a set of uncontrollable timepoint variables.

- Each uncontrollable timepoint $u_j \in U$ is linked to exactly one controllable timepoint $a_i \in A$ by a contingency link $l \in L : l = (a_i, [x, y], u_j)$

- *C* is a set of free constraints of the same form as with STNs, except constraints can also involve uncontrollable timepoints in addition to controllable timepoints.

We refer to timepoints in general (controllable or uncontrollable) as $V = A \cup U$. Different types of *controllability* exist (Vidal, 1999):

- *Strong Controllablity* (SC): An STNU $\Gamma = (A, U, C, L)$ is strongly controllable if there exists at least one universal schedule of controllable timepoints $\{a_1 = w_1, a_2 = w_2, ..., a_n = w_n\}$ which satisfies the constraints in *C* regardless of the values taken by uncontrollable timepoints *U*.

- *Weak Controllablity* (WC): An STNU $\Gamma = (A, U, C, L)$ is weakly controllable if, for every value outcome of uncontrollable timepoints *U*, there is at least one schedule of controllable timepoints $\{a_1 = w_1, a_2 = w_2, ..., a_n = w_n\}$ which satisfies the constraints in *C*.

- *Dynamic Controllablity* (DC): An STNU $\Gamma = (A, U, C, L)$ is dynamically controllable if there is a reactive strategy which guarantees constraints in *C* will be satisfied if the scheduling strategy is followed by a controller agent, while observing possible occurrences of uncontrollable timepoints and using this knowledge to adapt decisions. It is said the problem is DC if and only if it admits a valid dynamic strategy expressed as a map from partial schedules to Real-Time Execution Decisions (RTEDs) (Cimatti, Micheli, and Roveri, 2016). A partial schedule represents the current scheduling state, *i.e.* the set of timepoints that have been scheduled so far and their timing. RTEDs are popular semantics used to express a DC strategy (Hunsberger, 2009). RTEDs regroup two possible actions: **(1)** The wait action, *i.e.* wait for an uncontrollable timepoint to occur. **(2)** The $(t, \mathcal{X})$ action, *i.e.* if nothing happens before time $t \in \mathbb{R}$, schedule the controllable timepoints in $\mathcal{X}$ at *t*. A strategy is valid if, for every possible occurrence of the uncontrollable timepoints, controllable timepoints get scheduled in a way that all free constraints are satisfied.

In this thesis, we are interested in DC. Considerable work has resulted in algorithms to determine whether or not an STNU is DC or not (Morris and Muscettola, 2005; Morris, 2014), leading to $\mathcal{O}(N^3)$ worst-case DC-checking algorithms, where *N* is the number of timepoints of the STNU. These DC-checking algorithms also synthesize valid DC strategies executable in $\mathcal{O}(N^3)$ (Hunsberger, 2016; Morris, 2014).

Disjunctive Temporal Networks with Uncertainty (DTNUs) generalize STNUs by allowing the presence of disjunctions in the constraints *C* or contingency links *L*.

Formally, each constraint in $C$ is of the form : $\vee_{k=1}^{q} v_{k,j} - v_{k,i} \in [x_k, y_k]$. Furthermore, each contingency link $l \in L$ is of the form : $(a_i, \vee_{k=1}^{q'} [x_k, y_k], u_j)$ where $x_k \leq y_k \leq x_{k+1} \leq y_{k+1}$ $\forall k = 1, 2, ..., q' - 1$. All controllability types for STNUs remain available for DTNUs. The introduction of disjunctions inside $C$ and $L$ renders STNU's $\mathcal{O}(N^3)$ DC-checking algorithms unavailable for DTNUs. In fact, the complexity of DC checking for DTNUs is *PSPACE*-complete (Bhargava and Williams, 2019), making this a highly challenging problem. The difficulty in proving or disproving DC arises from the need to check all possible combinations of disjuncts in order to handle all possible occurrence outcomes of the uncontrollable timepoints. The only known approach for DC-checking and DC strategy generation for DTNUs is based on expressing DTNUs as timed-game automata (TGAs) (Cimatti et al., 2014). TGAs can then be solved by the UPPAAL-TIGA software (Behrmann et al., 2007). In (Cimatti, Micheli, and Roveri, 2016), authors express DTNUs as TGAs in the same way, but use a pruning procedure based on satisfiability modulo theory and achieve superior results than with UPPAAL-TIGA.

In this thesis, we choose to introduce a new variant of DC based on time discretization instead, *Restricted Time-based Dynamic Controllability* (R-TDC), which is less flexible than DC. R-TDC allows higher strategy search efficiency while almost always finding a strategy when a DC one exists on considered benchmarks owing to proper time discretization rules. R-TDC also enables the use of learning-based heuristics which are at the center of interest in the thesis.

# Chapter 3

# Global Search with Optimal Methods

This chapter formalizes the first problem presented in the context section mathematically. Two exact resolution approaches are introduced to solve the problem optimally. The first approach combines a Constraint Programming (CP) solver with a Graph Convolutional Network (GCN), the second one a Branch & Bound (B&B) tree search algorithm with a GCN.

## 3.1 Preliminaries

### Problem Formalization

We consider the problem of itineraries with mandatory pass-by nodes in a terrain, as presented in the first problem in the context chapter. Let $\mathcal{G} = (\mathcal{V}, \mathcal{E})$ be the weighted connected graph corresponding to the terrain. We assume $\mathcal{G}$ to be undirected, as Autonomous Unmanned Ground Vehicle (AUGV) scenarios considered in experiments correspond to undirected graphs. Nevertheless, resolution methods proposed are also applicable to directed graphs (only the heuristic used, *i.e.* the spectral-based GCN architecture, needs to be replaced with a spatial-based architecture that takes edge direction into account, *e.g.* the message passing neural network architecture described in Equation 2.12).

In each mission in the online phase, the AUGV begins at a start node, has to pass by each mandatory node at least once (but in no specific order) and finish at a destination node. We define a mission as a *problem instance*. An instance $I$ is made of three elements:

$$I = (start, dest, M)$$

- $start \in \mathcal{V}$ is the start node in graph $\mathcal{G}$.

- $dest \in \mathcal{V}$ is the destination node in graph $\mathcal{G}$.

- $M \subset \mathcal{V}$ is a set of distinct mandatory nodes that need to be visited at least once, with no specific order of visit.

In order to solve instance $I$ optimally, one has to find a shortest path from node *start* to node *dest* that passes by each node in $M$ at least once. The order of visit of nodes in $M$ is not imposed. Moreover, there is no limit to how many times a node or an edge can be included in a solution path. Since the graph is connected, there is a solution path to every existing instances. A solution path $\pi$ is an ordered list of nodes $\pi = \{start, v_1, ..., v_q, dest\}$, where $v_i$ are graph nodes, such that $M \subset \pi$. The associated total distance cost for $\pi$ is the sum of the distance cost of all edges included: $d_{start,v_1} + d_{v_1,v_2} + ... + d_{v_q,dest}$ where $d_{v,v'}$ is the weight of the edge linking $v$ and $v'$. Let $A$ be the adjacency matrix of graph $\mathcal{G}$, used in (§3.2.2) by our GCN, defined as follows:

$$A_{vv'} = \begin{cases} 0, & \text{if there is no edge from node } v \text{ to node } v' \\ 1, & \text{otherwise} \end{cases}$$

We simply refer to a problem instance as an instance next.

## Number of Instances and Complexity

Suppose graph $\mathcal{G} = (\mathcal{V}, \mathcal{E})$ has $|V| = n$ nodes. Let $P_n$ be the set of all existing instances for graph $\mathcal{G}$. The number of instances $C_n = |P_n|$ that exist is given by the following formula:

$$C_n = 2! \binom{n}{2} \sum_{p=0}^{n-2} \binom{n-2}{p} \tag{3.1}$$

where:

- $2!\binom{n}{2} = n(n-1)$ is the number of existing combinations of source-destination pairs, taking ordering into account.

- $\sum_{p=0}^{n-2} \binom{n-2}{p}$ is, for every possible source-destination pair, the number of possible sets of mandatory nodes to visit. It can be simplified to $2^{n-2}$

Therefore, equation (3.1) can be rewritten as:

$$C_n = 2^{n-2}n(n-1) \tag{3.2}$$

The number of instances for a graph with $n = 20$ is equal to $99\,614\,720$. Calculating the optimal solutions of all instances in the offline phase so that they can then be used online becomes too difficult even if a long time is allocated to the offline preparation phase. Instead, we propose to learn the topology of graph problem instances in the offline phase with machine learning techniques, more specifically a GCN. We then use this learned knowledge in the online phase in order to accelerate solving of problems on the go, as AUGV missions are defined. Furthermore, in terms of complexity, we show in § 3.3.4 that the problem is NP-hard relative to the number of mandatory nodes, and a variant of the Travelling Salesman Problem (TSP) with different start and destination nodes. The most efficient optimal TSP solver to date is ("Concorde TSP Solver"). Our proposed contributions based on learning are meant to be applied in future work beyond constraints involving mandatory nodes only, and therefore to more complex problems than the TSP. Therefore, we do not make comparisons with state-of-the-art non learning-based approaches such as Concorde, as we do not intend for it to perform better. Experiments in § 3.3.5 outline how slow some solvers can be, and how solving time rises exponentially with an increasing number of mandatory nodes.

## Related Works

In the past few years there has been a growing interest for transferring the intuitions and practices from neural networks on structured inputs towards graphs (Gori, Monfardini, and Scarselli, 2005; Henaff, Bruna, and LeCun, 2015; Defferrard, Bresson, and Vandergheynst, 2016; Kipf and Welling, 2017). These works led to the development of Graph Neural Networks (GNN) with a wide array of architectures. Applications of these types of networks have since been starting to emerge. Recent works suggest GNNs are capable of making key decisions to solve planning problems and combinatorial problems. While this holds especially true for solving problems approximately, there are few approaches leveraging GNNs to solve such problems optimally. Optimal solving requires visiting most of the search space to ensure proof of optimality, except parts where it can be proven that no better solution exists. To take advantage of GNNs in this particular context, one needs to couple them with admissible heuristics in a way which ensures the search space that can be discarded is big enough to reduce computational costs while maintaining proof of optimality.

At the time this work was being done, we found no relevant related works. At the time this thesis is being written however, we found the approach in (Gasse et al., 2019) to be related to the work in this chapter. Gasse *et al.* use a GNN to assist a B&B solver for mixed-integer linear programming problems. More specifically, the GNN is used to learn B&B variable selection policies instead of node selection policies, which lead to a high-quality but expensive branching rule. They leverage the natural bipartite graph representation of mixed-integer linear programming problems to encode the state of the B&B process as a graph. This graph is then fed as input to their GNN which outputs a probability distribution over candidate branching variables. The resulting framework reportedly improves upon state-of-the-art machine learning methods for variable branching, which suffer from the need of extensive feature engineering (Khalil et al., 2016; Hansknecht, Joormann, and Stiller, 2018; Alvarez, Louveaux, and Wehenkel, 2017).

In this work, we explore two different approaches. A first approach aims at leveraging information learned by a GNN and use it to change the exploration order of decision variables of a CP solver. The second approach uses a GNN to create an initial upper bound for a B&B tree search algorithm.

## Definitions and Notations

### Constraint Optimization Problem

We refer to a constraint satisfaction problem as a triplet $(Y, F, C)$.

- $Y = (\varphi_1, \varphi_2, ..., \varphi_n)$ is the set of variables.

- $F = (F_1, F_2, ..., F_n)$ is the set of domains of the variables, *i.e.* $\varphi_i \in F_i \ \forall i \in [|1, n|]$.

- $C = (C_1, C_2, ..., C_m)$ is the list of constraints. A constraint $C_i = (\Phi_i, R_i)$ is made of a set $\Phi_i = (\varphi_{i1}, \varphi_{i2}, ..., \varphi_{ik}) \subset Y$ of variables and a relation $R_i \subset F_{i1} \times F_{i2} ... \times F_{ik}$ which defines the values allowed simultaneously for the variables in $\Phi_i$.

A constraint optimization problem is a constraint satisfaction problem with an associated objective function to optimize. The objective function is of the form: $\sum_{i=1}^{n} w_i \varphi_i$, where $w_i \in \mathbb{R} \ \forall i \in [|1, n|]$. In our case, we will be minimizing the function.

**Dijkstra's Algorithm**

We give the pseudocode for Dijkstra's algorithm in Algorithm 7 in the appendix, used in the following sections to compute a shortest path between a source node $v_i$ and a destination node $v_j$ in a graph $\mathcal{G} = (\mathcal{V}, \mathcal{E})$.

**Notations**

Notations can be different between chapters. We provide some notations at the beginning of each chapter when necessary to avoid confusion. The following notations are used in this chapter:

- $I = (start, dest, M)$ : A problem instance.

- $\mathbf{x}$ : A problem instance, encoded as a vector. Vector $\mathbf{x}$ contains the features of every node in the graph of the problem instance. We denote as $\mathbf{x}_1, \mathbf{x}_2, ..., \mathbf{x}_m$ vectors of different problem instances of a dataset.

- $\hat{\mathbf{y}}$: The output of a GCN for an input $\mathbf{x}$.

- $\mathbf{y}$ : A target graph node associated to $\mathbf{x}$.

- $\theta$: The parameters of machine learning models we use.

- $d_{v,v'}$: The weight (distance cost) of the edge $e_{v,v'}$.

- $\pi$: A solution path.

## 3.2 Optimal Solving with Constraint Programming and a Graph Convolutional Network

In this section, we propose a hybrid model that combines a CP-based solver and a GCN to improve search performance in the online phase. The idea consists in learning graph problem topology in the offline phase with a GCN in order to accelerate the CP planner in the online phase.

### 3.2.1 Constraint Programming Approach

#### 3.2.1.1 Constraint Programming for Navigation and Maneuver Planning

In our approach, planning is achieved using CP techniques, under a model-based development approach. CP provides a powerful baseline to model and solve constraint optimization problems (COP) and/or constraint satisfaction problems (CSP). It has been introduced in the late 70s (Laurière, 1978) and has been developed until now (Van Hentenryck, Saraswat, and Deville, 1998; Ajili and Wallace, 2004; Carlsson, 2015), with several real-world autonomous system applications, in space (Bornschlegl, Guettier, and Poncet, 2000; Simonin et al., 2015), aeronautics (Guettier et al., 2002; Guettier and Lucas, 2016) and defense (Guettier et al., 2015). In CP, it is possible to design global search algorithms that guarantee completeness and optimality. A CSP, formulated within a CP environment, is composed of a set of variables, their domains and algebraic constraints which are based on problem discretization. A COP is a CSP with an associated objective function to optimize. With CP, a declarative formulation of the constraints to satisfy is provided which is decoupled from the search algorithms, so that both can be worked out independently. The COP formulation and search algorithms exposed in this work are implemented with the CLP(FD) domain of SICStus Prolog library (Carlsson, 2015). It uses the state-of-the-art in discrete constrained optimization techniques Arc Consistency-5 (AC-5)((Deville and Van Hentenryck, 1991), (Van Hentenryck, Deville, and Teng, 1992)) for constraint propagation, managed by CLP(FD) predicates. With AC-5, variable domains get reduced until a fixed point is reached by constraint propagation. The search technique is also hybridized by statically defining search exploration structure using probing and information learned from multiple problem instances by a GCN.

#### 3.2.1.2 Planning Model with Flow Constraints

The set of flow variables $\varphi_e \in \{0, 1\}$ models a possible path from $start \in \mathcal{V}$ to $dest \in \mathcal{V}$, where an edge $e$ belongs to the navigation plan if and only if a decision variable $\varphi_e = 1$, 0 otherwise. The resulting navigation plan can be represented as $\Phi = \{e| \ e \in \mathcal{E}, \ \varphi_e = 1\}$. From an initial position to a requested final one, path consistency is enforced by flow conservation equations, where $\omega^+(v) \subset \mathcal{E}$ and $\omega^-(v) \subset \mathcal{E}$ represent respectively outgoing and incoming edges from vertex $v$. Since flow variables are $\{0, 1\}$, equation (3.3) ensures path connectivity and uniqueness while equation (3.4) imposes limit conditions for starting and ending the path:

$$\sum_{e \, \in \, \omega^+(v)} \varphi_e = \sum_{e \, \in \, \omega^-(v)} \varphi_e \leq N \tag{3.3}$$

$$\sum_{e \, \in \, \omega^+(start)} \varphi_e = 1, \quad \sum_{e \, \in \, \omega^-(end)} \varphi_e = 1, \tag{3.4}$$

These constraints provide a linear chain alternating pass-by nodes and navigation along the graph edges. Constant $N$ indicates the maximum number of times a node can be included in the solution ($+\infty$ in our case). With this formulation, the plan may contain cycles over several nodes. Mandatory nodes are imposed using constraint (3.5). The total path length is given by the metric (3.6) and is the criterion to minimize:

$$\forall i \in M \sum_{e \, \in \, \omega^+(i)} \varphi_e \geq 1 \tag{3.5}$$

$$T = \sum_{v \, \in \, \mathcal{V}} \sum_{v'v \, \in \, \omega^-(v)} \varphi_{v'v} d_{v'v} \tag{3.6}$$

### 3.2.1.3  Global Search Algorithm

The global search technique under consideration guarantees completeness, solution optimality and proof of optimality. It relies on three main algorithmic components:

- Variable filtering with correct values, using specific labeling predicates to instantiate problem domain variables. AC being incomplete, value filtering guarantees search completeness.

- Tree search with standard backtracking when variable instantiation fails.

- Branch and Bound (B&B) for cost optimization, using minimize predicate.

Designing a good search technique consists in finding the right variables ordering and value filtering, accelerated by domain or generic heuristics. Use of the learning heuristic introduced in the next section provides an initial variable selection ordering, computed before running the global branch and bound search. Note that in general probing techniques (El Sakkout and Wallace, 2000), the order can be redefined within the search structure (Ruml, 2001). Similarly, in our approach, the variable selection order provided by the probe can still be iteratively updated by the labeling strategy that makes use of other variable selection heuristics. Mainly, first fail variable selection is used in addition to the initial probing order. These algorithmic designs have already been reported with different probing heuristics (Guettier and Lucas, 2016), such as A* or meta-heuristics such as Ant Colony Optimization (Lucas et al., 2010), (Guettier, Lucas, and Siarry, 2009). In our design, the search is still complete, guarantying proof of optimality, but demonstrates efficient pruning. Instead of these heuristics techniques, we choose to train a GCN to predict the probing mechanism that provides a tentative variable order to the global search.

### 3.2.2 Graph Convolutional Network Architecture

In this section, we present how a GCN is trained on a particular graph in the offline phase. The aim is to let the GCN learn to approximate the behavior of a model-based CP planner on the graph. To this end, we first use our CP solver to compute solutions for several random instances and use them as training data. Then we train the GCN by using the previously generated solutions as supervision. We seek to leverage the powerful mechanism of GCNs to guide our path-planner in the online phase. We justify the use of a GCN architecture here over a Multi-Layer Perceptron (MLP) made of fully-connected layers because of the drastic reduction in the input space made possible by taking into account graph adjacency information directly inside graph convolutions rather than in inputs. Indeed, an MLP would need to incorporate adjacency information in inputs, which would result in increasing the size of the input space by $|\mathcal{V}|^2$ to encode edge connectivity and therefore a considerably harder model to train.

We consider here the approach of (Kipf and Welling, 2017). The GCN has the following layer propagation rule to compute the matrix of all node feature vectors $H^{(l+1)}$ at layer $l + 1$ from layer $l$:

$$H^{(l+1)} = \sigma \left( \tilde{D}^{-\frac{1}{2}} \tilde{A} \tilde{D}^{-\frac{1}{2}} H^{(l)} \theta^{(l)} \right),\tag{3.7}$$

where $\tilde{A} = A + I_N$ is the adjacency matrix of the graph with added self-connections such that when multiplying with $\tilde{A}$ we aggregate features vectors from both a node and its neighbors; $I_N$ the identity matrix. Matrix $\tilde{D}$ is the diagonal node degree matrix of $\tilde{A}$. Lastly, $\sigma(\cdot)$ is the activation function, which we set to ReLU$(\cdot) = \max(0, \cdot)$, and $\theta^{(l)}$ are the weights of the GCN. The diagonal degree $\tilde{D}$ is employed for normalization of $\tilde{A}$ in order to avoid change of scales in the feature vectors when multiplying with $\tilde{A}$.

(Kipf and Welling, 2017) argue that using a symmetric normalization, *i.e.* $\tilde{D}^{-\frac{1}{2}}\tilde{A}\tilde{D}^{-\frac{1}{2}}$, ensures better dynamics compared to simple averaging of neighboring nodes in one-sided normalization $\tilde{D}^{-1}A$. The input of our GCN is a vector $\mathbf{x}$, where $h^{(0)} = \mathbf{x}$, containing information about the problem instance. We describe next the chosen encoding for vector $\mathbf{x}$.

### 3.2.2.1 Vector Encoding of Instances

For every instance $I = (start, dest, M)$ of a given graph $\mathcal{G} = (\mathcal{V}, \mathcal{E})$, we associate a vector $\mathbf{x}$ made up of triplets features per node in $\mathcal{G}$, making up for a total of $3|\mathcal{V}|$ features. The three features for a node $v_j \in \mathcal{V}$ are:

- A *start node* feature:

$$st_j = \begin{cases} 1, & \text{if node } v_j \text{ is the start node in instance } I \\ 0, & \text{otherwise} \end{cases}$$

- An *end node* feature:

$$ed_j = \begin{cases} 1, & \text{if node } v_j \text{ is the destination node in instance } I \\ 0, & \text{otherwise} \end{cases}$$

- A *mandatory node* feature:

$$my_j = \begin{cases} 1, & \text{if node } v_j \text{ is a mandatory node in instance } I \\ 0, & \text{otherwise} \end{cases}$$

Vector $\mathbf{x}$ is the concatenation of these features:

$$\mathbf{x} = (st_1, ed_1, my_1, st_2, ed_2, my_2, ..., st_{|\mathcal{V}|}, ed_{|\mathcal{V}|}, my_{|\mathcal{V}|}) \tag{3.8}$$

Figure 3.1 illustrates an example of an instance $I$ in a graph and the associated vector $\mathbf{x}$.

### 3.2.2.2 Architecture

We define a GCN $f$ that uses a sequence of graph convolutions followed by a fully connected layer. The idea is to have the GCN take as input any instance $I$ on the graph $\mathcal{G}$, and output a probability $\hat{\mathbf{y}}$ over which node should be visited first from the start node in an optimal path that solves $I$. The weights of the GCN $\boldsymbol{\theta}$ are tuned in the training phase for this purpose. We train the GCN on instances that have already been processed by the CP solver. Here, the solver serves as a *teacher* to the GCN and the GCN learns to approximate the solutions given by the CP solver.

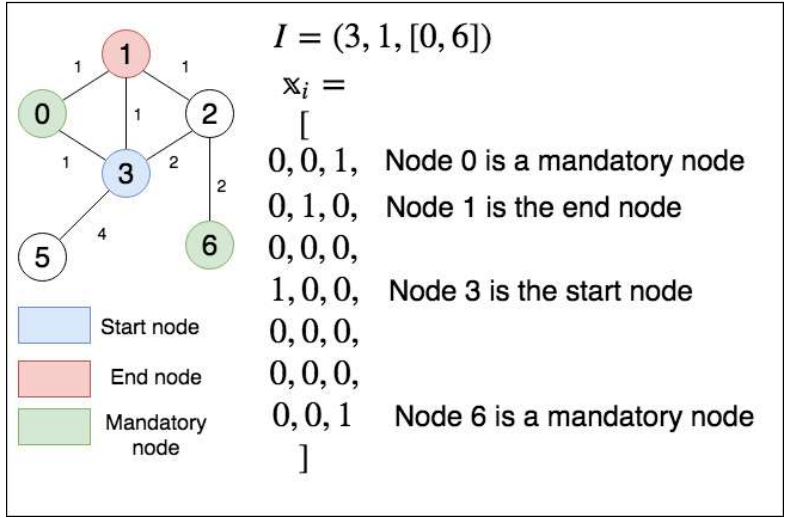

**Figure 3.1: Instance vector encoding**. This graph example has seven nodes. The instance $I$ defined in the figure requires finding an optimal path from node 3 to node 1 which passes by node 0 and 6 at least once. Vector $\mathbf{x}_i$ is the encoding of instance $I = (3, 1, [0, 6])$. The number written next to an edge is its distance cost.

The output $\hat{\mathbf{y}}$ of the GCN is a vector of size $|\mathcal{V}|$. We format the input vector $\mathbf{x}$ as following: we reshape $\mathbf{x}$ into a matrix of size $(|\mathcal{V}|, 3)$, which contains the *start*, *destination* and *mandatory* features of every node by rows. This matrix becomes the input for $f$, which aims to extract and aggregate local information over layers starting from the input. The output of the convolutions, a matrix, is flattened into a vector by concatenating its rows. A fully connected layer then links the flattened vector to a vector $z$ of real-numbers of size $|\mathcal{V}|$. Finally we use a softmax function to obtain a probability distribution $\hat{\mathbf{y}}$. Formally, the softmax function is given by:

$$\text{softmax}(\mathbf{z})_i = \frac{e^{z_i}}{\sum_{k=1}^{|\mathcal{V}|} e^{z_k}} \tag{3.9}$$

This formula ensures that:
$\forall i \in \{1, 2, ..., |\mathcal{V}|\}$, $\text{softmax}(\mathbf{z})_i > 0$ and $\sum_{i=1}^{|V|} \text{softmax}(\mathbf{z})_i = 1$.

Once the training of the GCN is concluded, we aim to have it return the next node to visit to optimally solve any instance $I$ in the same manner as the CP solver would find.

### 3.2.3 Data Generation

This subsection provides a global schema for data generation.

#### 3.2.3.1 Instance Generator

In order to generate instances $I_i = (start_i, dest_i, M_i)$, a generator function is built and given in Algorithm 1. It returns a set of instances $R$, all of which are solvable since graph $\mathcal{G}$ is assumed to be connected. These instances are sampled out of the set of all possible instance configurations $P_n = \{I_1, I_2, ..., I_{2^{n-2}n(n-1)}\}$. For our experiments, we built 2 different connected undirected graphs $\mathcal{G}_1$ and $\mathcal{G}_2$ containing respectively 15 and 22 nodes. Problem instances to be generated for those graphs must remain realistic in terms of mission data. To be close to some realistic instances, the instance generator works the following way to create instances. It first generates a shortest path length among all pairs of start to destination nodes $(start_i, dest_i)$ within the graph. A decimation ratio is then applied (typically 90%), keeping the 10% set of pairs that have the longest paths. For each resulting pair $(start_i, dest_i)$, multiple random instances are created, with an increasing cardinality for the set of mandatory nodes (typically from 1 to 10 mandatory nodes). For each instance, mandatory nodes are selected recursively. In this process, another decimation ratio (50%) is applied in order to extract a group of nodes for which the shortest path from the previously selected mandatory node to these nodes is longest, and a new mandatory node is selected from the extracted group following a discrete uniform distribution. As such, generated instances are likely to have long solution paths.

---

**Algorithm 1** Instance Generator

---

1: **function** GenerateInstance(Graph $\mathcal{G} = (\mathcal{V}, \mathcal{E})$, Param $\gamma$, Param $\delta$)     ▷ [1]*
2:     $C(.,.) \leftarrow 0$ // Initialization of shortest path cost matrix
3:     **for all** $v \in \mathcal{V}$ **do**
4:         **for all** $v' \in \mathcal{V} - \{v\}$ **do**
5:             $C(v, v') \leftarrow ShortestPathCost(v, v')$     ▷ [2]*
6:     $P \leftarrow DecimationRatio(C)$     ▷ [3]*
7:     Benchmark $B \leftarrow \varnothing$
8:     **for all** $(v, v') \in P$ **do**
9:         $i \leftarrow 0$
10:         **while** $i < \gamma$ **do**
11:             $mand \leftarrow \mathcal{V} - \{v, v'\}$
12:             $M \leftarrow \varnothing$
13:             $start \leftarrow v$
14:             $j \leftarrow 0$
15:             **while** $j < \delta$ **do**
16:                 $v'' \leftarrow SelectMandatory(mand, start, C)$     ▷ [4]*
17:                 $mand \leftarrow mand - \{v''\}$
18:                 $M \leftarrow M \cup \{v''\}$
19:                 $start \leftarrow v''$
20:                 $j \leftarrow j + 1$
21:             $B \leftarrow B \cup \{(v, v', M)\}$
22:             $i \leftarrow i + 1$
23:     Return $B$

---

[1]* $\gamma$: Number of instances generated per source-destination pair; $\delta$: Number of mandatory nodes of generated instances

[2]* $ShortestPathCost(v, v')$: Returns the cost of the shortest path linking nodes $v$ and $v'$.

[3]* $DecimationRatio(C)$: Returns the first 10% pairs of nodes which have the longest shortest paths in the cost matrix $C$.

[4]* $SelectMandatory(mand, start, C)$: Using shortest costs matrix $C$, extracts the first 50% nodes from $mand$ for which the shortest path cost from $start$ to these nodes is longest. Selects a node from these extracted nodes based on a discrete uniform distribution and returns it.

---

To create optimal data for graph $\mathcal{G}$, each instance $I_i$ generated in $R$ is fed to our CP solver, providing an optimal path $\pi_i^*$. We set a 'time out' at 3 seconds for each instance. If no optimal solution can be proven, the instance is discarded. The number of instances generated and resolved are reported in Table 3.1.

Only 651 (over 1256 generated) and 554 (over 2503) are solved optimally for experiment's graphs $\mathcal{G}_1$ and $\mathcal{G}_2$ respectively. Indeed, easier instances are optimally solved more often than difficult ones. Given the number of nodes in the graphs,

**Table 3.1:** Number of instances resolved under 3 seconds by the CP solver out of the generation process.

| Mandatory nodes #: | 0 | 1 | 2 | 4 | 6 | 8 | 10 |
|---|---|---|---|---|---|---|---|
| Instances for graph ($\mathcal{G}_1$) | | | | | | | |
| generated (1256): | 42 | 212 | 246 | 252 | 252 | 252 | |
| optimally solved (651): | 42 | 212 | 185 | 121 | 61 | 30 | |
| Instances for graph $\mathcal{G}_2$ | | | | | | | |
| generated (2503): | 69 | 368 | 410 | 414 | 414 | 414 | 414 |
| optimally solved (554): | 69 | 265 | 168 | 42 | 9 | 1 | 0 |

Equation 3.1 states we are working with less than 0.001% of the total instances in both graphs. As a consequence, most of the GCN training is performed on a small number of instances. For each instance $I_i = (start_i, dest_i, M_i)$, let $|M_i|$ be the number of mandatory nodes from $M_i$. The corresponding optimal path $\pi_i^*$ found by our CP solver is a path that passes through all the nodes in $M_i = \{m_{i1}, m_{i2}, ..., m_{i|M_i|}\}$, *i.e.* :

$$\pi_i^* = \{start_i, v_{i1}, v_{i2}, ..., v_{iq}, dest_i\}, m_{ij} \in \{v_{i1}, v_{i2}, ..., v_{iq}\} \ \forall j \in 1, 2, ..., |M_i|$$

Since our GCN model predicts the next node to visit for a given instance $I$, our data is reprocessed before the training phase using the following lemma:

**Lemma 1.** *Let $I = (start, dest, M)$ be a problem instance, and $\pi^* = \{start, v_1, v_2, v_3, ..., v_q, dest\}$ an optimal path that solves $I$. It comes that:*

- $\{v_1, v_2, v_3, ..., v_q, dest\}$ *is an optimal solution for the instance* $(v_1, dest, M\backslash\{v_1\})$

- $\{v_2, v_3, ..., v_q, dest\}$ *is an optimal solution for the instance* $(v_2, dest, M\backslash\{v_1, v_2\})$

- *...*

- $\{v_i, v_{i+1}, ..., v_q, dest\}$ *is an optimal solution for the instance*
  $(v_i, dest, M\backslash\{v_1, v_2, ..., v_i\}) \mid \forall i \in 1, 2, ..., q$

- *...*

- $\{v_q, dest\}$ *is an optimal solution for the instance* $(v_q, dest, M\backslash\{v_1, v_2, .., v_q\})$

*Proof.* Suppose $\{v_1, v_2, v_3, ..., v_q, dest\}$ is not optimal for $(v_1, dest, M\backslash\{v_1\})$. There is therefore a path $\pi'$ that starts from $v_1$, ends in *dest* and visits every node in $M\backslash\{v_1\}$ with a lower cost than the path $\{v_1, v_2, v_3, ..., v_q, dest\}$. Therefore $\pi^* = $ $\{start, v_1, v_2, v_3, ..., v_q, dest\}$ is not optimal for the original problem $I$ since it does not take that shorter path, which is contradictory. It results that $\{v_1, v_2, v_3, ..., v_q, dest\}$ is optimal for $(v_1, dest, M \backslash \{v_1\})$. The same reasoning is applied recursively. $\qquad \square$

### 3.2.3.2 Data Processing

Let $(I_i, \pi_i^*)$ be an instance-solution pair that was previously generated, such that $I_i = (start_i, dest_i, M_i)$ and $\pi_i^* = \{start_i, v_{i1}, v_{i2}, v_{i3}, ..., v_{iq_i}, dest_i\}$. This pair, which we call root pair, is split into several pairs $(I_{i,j}, \pi_{i,j}^*), \forall j \in 1, 2, ..., q_i$ as described in Lemma 1. This guarantees that for each instance $I_{i,j}$, $\pi_{i,j}^*$ is an optimal solution path. For each newly obtained pair $(I_{i,j}, \pi_{i,j}^*)$, we store $(I_{i,j}, t_{i,j})$ in a training set $\mathcal{X}$, where $t_{i,j} \in \mathcal{V}$ is the first node visited in path $\pi_{i,j}^*$ after the start node. The same process is applied for every root pair $(I_i, \pi_i^*)$ which was stored.

### 3.2.3.3 Supervised Learning and Implementation Details

Following the creation of training set $\mathcal{X}$, we train the GCN so that it can learn to approximate the behavior of our CP solver, and correctly predict the first node that should be visited from the start node in an optimal path for a given instance $I$. We take 80% of the data in training set $\mathcal{X}$ for training the GCN. The validation set is given the remaining 20% of the data in $\mathcal{X}$. Let $f$ be the function for the GCN defined previously. Function $f$ takes as input a vector instance $\mathbf{x}_i$ and outputs a distribution vector $\hat{\mathbf{y}}_i$ which is the probability distribution over all nodes in the graph of the next node to visit: $f(\mathbf{x}_i; \boldsymbol{\theta}) = \hat{\mathbf{y}}_i$ where $\boldsymbol{\theta}$ are the weights of the GCN. We use the multi-class cross-entropy loss function $\mathcal{L}$: the average of the logarithmic loss of the probabilities predicted by the GCN for each problem instance $I_i$ with the actual label target node $y_i$. Function $\mathcal{L}$ is defined as follows:

$$\mathcal{L}(\theta) = \frac{1}{m} \sum_{i=1}^{m} \sum_{j=1}^{n} -y_{ij} log(f(\mathbf{x}_i; \boldsymbol{\theta})_j) \tag{3.10}$$

- $m$ is the number of training examples in the training set,

- $n$ is the number of nodes in the graph,

- $\mathbf{x}_i$ is the vector of the problem instance $I_i$,

- $y_{ij}$ is a variable that indicates whether for the problem instance $I_i$, node number $j$ was the next optimal node to visit: $y_{ij} = 1$ if so, else $y_{ij} = 0$,

- $f(\mathbf{x}_i; \boldsymbol{\theta})_j$ is the probability the GCN outputs to visit node $j$ for the problem instance $I_i$.

The GCN is trained using a variant of the stochastic gradient descent (SGD) called *Adam* (Kingma and Ba, 2015). *Adam* uses adaptive learning rates for each variable of the GCN to decrease the number of steps required to minimize the loss function. To prevent overfitting, we apply the process of *early stopping*: the training is stopped once the average loss on the validation set stops decreasing, and restarts increasing. Training is done on an nvidia DGX-station (Intel Xeon E5-2698 v4 2.2 GHz, Tesla V100 GPU) on each training set obtained from graphs $\mathcal{G}_1$ and $\mathcal{G}_2$. We include training curves in Figures 3.2 and 3.3 for graph $\mathcal{G}_1$. The GCNs trained for each graph generalize relatively well on unknown instances: the GCN for graph $\mathcal{G}_1$ achieves 96% accuracy on its validation set, while the GCN for graph $\mathcal{G}_2$ achieves 92% accuracy on its validation set.

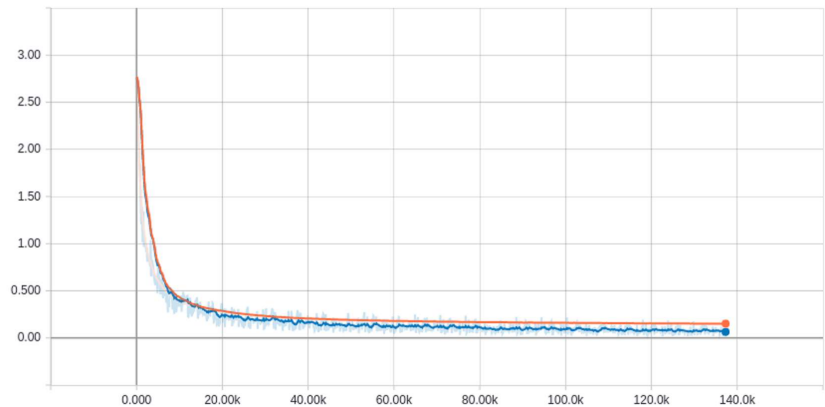

**Figure 3.2: Loss function curve during training for graph $\mathcal{G}_1$.** The X-axis is the iteration step. The Y-axis is the average logarithmic log loss. The blue curve is the loss on a batch of training examples from the training set, the orange curve is the loss for the validation set. Curves are smoothed.

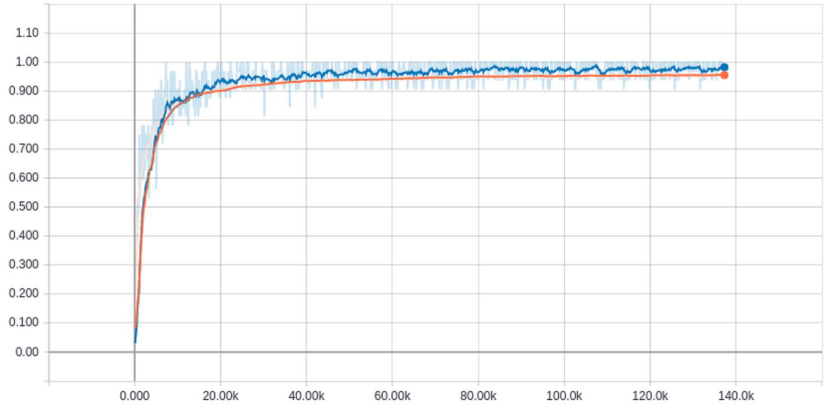

**Figure 3.3: The accuracy curve during training for graph $\mathcal{G}_1$.** The X-axis is the iteration step. The Y-axis is the prediction accuracy. The blue curve is the accuracy on a batch of training examples from the training set, the orange curve is the accuracy for the validation set. The model achieves an accuracy of 96% on the validation set. Curves are smoothed.

### 3.2.3.4 Framework and Hyperparameters

The GCN was implemented in Python using Tensorflow. To build the tensorflow graph, the adjacency matrix is first loaded in the python environment, and is then used for the graph convolutions. In the training phase, we used a dropout layer which connects the last hidden layer to the output layer to further reduce overfitting. The probability of a neuron dropping out was selected to be 0.1. We used batch normalization (Ioffe and Szegedy, 2015) for every layer in the graph convolutions with a decay parameter $\varepsilon = 0.9$. Lastly, we picked a learning rate $\eta = 10^{-4}$ and trained the GCN with mini-batches of data of size 32.

## 3.2.4 Experiments and Results

In this subsection, we evaluate to which extent the GCN can help the CP solver on new instances. When solving an instance $I$, the GCN is used at search start, and given as input the corresponding vector $\mathbf{x}$. The output $\hat{\mathbf{y}}$ is used to create a preference ordering over all existing nodes in the graph, and corresponds to the preference of the next node to visit from the start node of the instance $I$. This order suggested by the GCN is provided to the CP solver at the root of the search tree. Flow variables are then ordered accordingly to be explored in the suggested order: edges linking the start node to neighboring nodes are included and explored in the same order. Figure 3.4 depicts this process. The GCN is not used again to create preference orderings of resulting subtrees. Indeed, combining CP and the GCN at every choice point is complex and makes solving too slow and impractical.

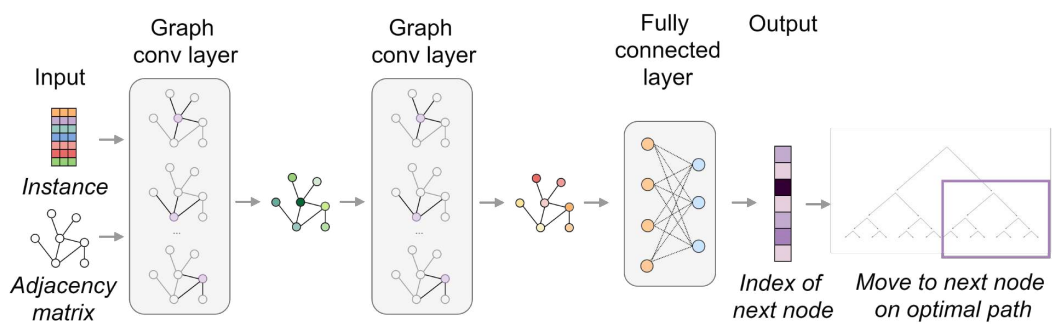

**Figure 3.4: Processing pipeline for path-planning using a GCN.** The GCN takes as input the adjacency matrix and the instance. Graph convolutional layers process each node in the graph and its neighbors. In the hidden layers, new features are generated for each node in the graph. In the last layer, the features are passed through a fully-connected layer and a softmax. The softmax layer indicates the predicted next node in an optimal path.

We generate instances for two benchmarks: $B_1$, associated with graph $\mathcal{G}_1$; and $B_2$, associated with graph $\mathcal{G}_2$. The first benchmark comprises 1008 instances, the second one 2208 instances. Those instances are generated with the instance generator defined in Algorithm 1, and therefore contain anywhere from 0 to 10 mandatory nodes. We solve the instances using our original CP solver without GCN support to obtain reference performance, denoted as CP. Then, we evaluate solving performance on the same instances using a modified version of the CP solver based on GCN probing, that was trained over data provided in § 3.2.3, which we denote as GCN + CP. In both cases, we only keep results where the proof of optimality could be achieved. Results are reported in Table 3.2, showing stable improvements on all datasets of instances. Experiments are done on a laptop with an Intel i5 processor and 8GB of RAM.

**Table 3.2:** Number of instances resolved with proof of optimality. Comparison between the reference version CP and the GCN probing one GCN + CP (under 3 second time out).

| Mandatory nodes #: | 3 | 5 | 7 | 9 |
|---|---|---|---|---|
| Benchmark $B_1$ | | | | |
| CP: | 167 | 99 | 43 | 15 |
| GCN + CP: | 220 | 185 | 131 | 89 |
| Benchmark $B_2$ | | | | |
| CP: | 104 | 29 | 4 | 1 |
| GCN + CP: | 210 | 65 | 18 | 4 |

**Table 3.3:** Global search features for proof of optimality, comparison between the reference version CP and the GCN probing one GCN + CP (under 3 seconds 'time out').

| Benchmarks | $B_1$ | $B_2$ |
|---|---|---|
| Number of instances resolved | | |
| CP | 324 | 138 |
| GCN + CP | 625 | 297 |
| Average solving time (ms) for optimality | | |
| CP | 504 | 1044 |
| GCN + CP | 394 | 1203 |
| Average number of backtracks for optimality | | |
| CP | 55787 | 26065 |
| GCN + CP | 35176 | 26504 |

Table 3.3 summarizes the number of instances solved by both solvers under 3 seconds with more detailed search features. Average number of backtracks and solving time on $B_1$ are significantly lower, denoting an efficient pruning of the search tree. The situation is different for $B_2$ with slightly more backtracks and higher solving time. Given that the graph for $B_2$ contains more nodes than the graph for $B_1$, this result is explained by more complex instances solved optimally with GCN + CP probing, with an accordingly high number of backtracks and long solving time. Reference CP probing was unable to solve those instances, thus not taking into account the number of backtracks and solving time for these complex instances. Lastly, higher performance of GCN + CP could be obtained by training the GCN on more complex instances. This would however require solving more instances with the reference CP solver before the training of the GCN, which is already time-consuming.

## Discussion

In order to solve the problem, we introduced a framework capable of significantly accelerating the performance of a CP-based solving method. The framework solves multiple random instances in the offline phase to train a GCN, which is then used to accelerate the CP solver on new instances in the online phase. The approach is efficient, even with small training datasets, as required by application preparation requirements. The performance obtained is realistic on two representative AUGV planning benchmarks. The considered approach nevertheless has two drawbacks. The first issue lies with the limited use of learned knowledge: the output of the GCN gives information only on the first node to visit from the start node for a given problem instance and is exploited as such in the CP model. No information is exploited on future directions with this setup. The second limitation stems from the use of the same instance generator to create the dataset on which the GCN is trained and the benchmark dataset used for experiments. Some bias could have been introduced as a result, and the generalization ability of the GCN may have been overestimated. We deal with these issues in the next approach.

## 3.3 Optimal Solving with Optimized Tree Search and a Graph Convolutional Network

We propose here a solving planner that combines a GCN with an optimal B&B tree search algorithm tasked with finding an optimal order of visit of mandatory nodes. The GCN is trained in the offline phase. We conduct experiments which show that GCN support can then enable substantial speedup of the B&B algorithm and smoother scaling to harder problems for the online phase.

### 3.3.1 Mandatory Node Ordering Approach

Solving $I = (start, dest, M)$ optimally means finding a path $\pi^*$, *i.e.* a sequence of nodes, which begins from $start$, ends in $dest$, minimizes the total weight of all edges included in $\pi^*$ and satisfies the mandatory constraints: the solution path is required to include all mandatory nodes at least once, *i.e.* $M \subset \pi^*$. The order of visit is not imposed for $M$. If it was, solving the problem could actually be done in polynomial time by consecutively computing pairs of shortest paths from the start node to the first mandatory node, then to the second mandatory node, and so on until the destination node. Figuring out an optimal order of visit of the mandatory nodes is what makes the problem difficult. In this approach, we tackle the problem from that angle.

First, we define a GCN architecture different from the previous approach which can define an entire solution path. Next, we propose a self-supervised training strategy for the GCN. We then provide a framework which combines the GCN with a depth-first B&B algorithm. We show that GCN support allows the B&B algorithm to significantly outperform several baselines, including an algorithm based on A* with handcrafted heuristic tailored to the planning domain of considered problems. Figure 3.5 summarizes the approach presented in this section.

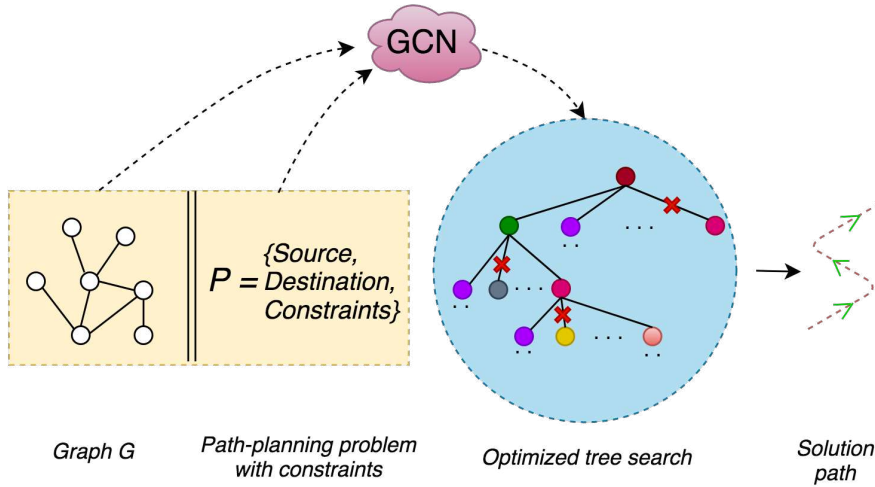

**Figure 3.5: Proposed framework for GCN-assisted branch & bound solving.** The GCN takes as input the graph and the problem and provides relevant information to speed up a tree search, after which an optimal solution path can be built.

## 3.3.2 Complete Path-building with a Graph Convolutional Network

In this section, we introduce a new architecture for the GCN which can be used recursively to build a solution path for a problem instance. We use the GCN convolution layer from (Kipf and Welling, 2017) again in this approach. More details can be found in § 3.2.2. We use the same encoding as in § 3.2.2.1 to encode a problem instance as a vector $\mathbf{x}$ (each node has 3 binary features: *start*, *end* and *mandatory*).

### 3.3.2.1 Graph Neural Network Architecture and Training

We define a GCN $f$ that consists of a sequence of multiple graph convolutions (each convolution is interleaved with the piece-wise $\text{ReLU}(\cdot) = \max(0, \cdot)$ activation function), followed by a fully connected layer. This GCN takes as input any instance $I$, and outputs a probability distribution $\hat{\mathbf{y}} \in \mathbb{R}^{|\mathcal{V}|}$ over all nodes in graph $\mathcal{G}$. After the final graph convolutional layer, the feature matrix is flattened into a vector by concatenating its rows and linked to a fully connected layer that maps it to a vector $z \in \mathbb{R}^{|\mathcal{V}|}$. We use the *softmax* function (Equation 3.9) to convert $z$ into probabilities $\hat{\mathbf{y}}$. Contrary to the approach taken in § 3.2.2.2, $\arg\max(\hat{\mathbf{y}})$ corresponds to the predicted next mandatory node to visit from the *start* node in an optimal path that solves $I$, not the first node to visit from the *start* node.

The GCN is trained in a supervised manner on labeled training data, *i.e.* a set of input-output pairs $(\mathbf{x}_i, \mathbf{y}_i)$ sampled from a training set. Here, $\mathbf{x}_i$ is an instance $I$ and $\mathbf{y}_i$ is the index of the next mandatory node for $\mathbf{x}_i$ in an optimal path solving $I$. We train the GCN on instances that have already been optimally solved by an optimal planner. The GCN learns to approximate the solutions computed by the planner. To this end, we train the GCN using the multi-class cross-entropy loss defined in Equation 3.10 and stochastic gradient descent.

### 3.3.2.2 Mandatory Node Ordering

For a given input instance, the GCN computes a mandatory node prediction at a time. In order to make it compatible with instances with varying amounts of mandatory nodes we make a few adjustments. Given an instance with multiple mandatory nodes $(start, dest, M)$, we perform multiple recursive GCN calls to get the next mandatory node predictions $q_i$. More specifically, after a prediction $q_i$ is computed, we generate a sub-instance where the start node becomes the current predicted mandatory node $q_i$ and where the list of mandatory nodes contains the remaining nodes except $q_i$, *i.e.* $(q_i, dest, \tilde{M}_i = \tilde{M}_{i-1} \setminus q_i)$, where $\tilde{M}_0 = M$ and $q_0 = start$. An interesting side-effect of this strategy is that it ensures an implicit balancing of the training samples by difficulty, as we generate sub-instances ranging from challenging (large $|M|$) to trivial (small $|M|$).

## 3.3.3 Self-Supervised Learning

Here we present a self-supervised learning strategy aimed at training the GCN on graph $\mathcal{G}$. First, we define a planning domain to solve instances. Second, we introduce a modified version of the A* algorithm for generating optimal data on which the GCN is trained. In this section, we refer to an instance as a planning state $s = (start, dest, M)$. An end state is a termination instance, *i.e.* an instance for which the destination node has been reached and all mandatory nodes have been visited. We denote the termination instances as $F_i = (i, i, \varnothing), i \in \{1, 2, ..., |V|\}$. There are exactly as many termination instances as there are nodes in $\mathcal{G}$. We respectively define the *successors* and *predecessors* of a state $s = (start, dest, M)$ as $succ(s)$ and $pred(s)$ in Table 3.4.

Transition cost from a state to a neighboring state is the cost of the edge in the graph linking the start nodes of both states. With these rules, the destination node remains always the same. Therefore, we run the backwards version of A* from every termination instance $F_i$ as initial state (using *pred* as rule of succession). For each

**Table 3.4:** Transition rules to successors and predecessors for planning states.

| $s = (start, dest, M)$ | |
| --- | --- |
| Successor state $s'$ | Predecessor state $s'$ |
| $s' = (start', dest', M')$ | $s' = (start', dest', M')$ |
| $(start, start') \in \mathcal{E}$ | $(start', start) \in \mathcal{E}$ |
| $dest' = dest$ | $dest' = dest$ |
| $M' = M \backslash \{start'\}$ | $M' = M$ ▷ [1]* |
| - | or |
| - | $M' = M \cup \{start\}$ ▷ [2]* |
| $s$ is not an end state | $s'$ is not an end state |
| Transition cost: $(start, start')$ | Transition cost: $(start', start)$ |

[1]* only if $start' \notin M$
[2]* only if $start \neq dest$

state $s$ visited by A*, a path $p$ is built from $s$ to $F_i$ which the algorithm considers the shortest. We define $g(s)$ as the cost of $p$, $a(s)$ as the next state visited after $s$ in $p$, and $d(s)$ as the first mandatory node of $s$ that is visited in $p$.

Furthermore we perform the following changes to A*. First, when a shorter path is found to a state $s'$ while developing a state $s$, *i.e.* $g(s') > g(s) + c(s, s')$, the values of $a(s')$ and $d(s')$ are also updated along with $g(s')$ to take into account the shorter path. Secondly, we set the heuristic function to $h(s) = 0$ for any state $s$. Since A* is run backwards from a termination state $F_i$, we are not aiming for the algorithm to reach a defined state in particular, but seek to reach as many states as possible. This ensures that when a state $s$ is taken from the OPEN priority list of states left to develop, an optimal path from $s$ to $F_i$ has already been found. Choosing $d(s)$ as the next mandatory node to visit thus enables optimal solving. Consequently we can add the pair $\langle s, d(s) \rangle$ to the training set. We provide the pseudo-code in Algorithm 2.

---

**Algorithm 2** Backwards A* for reverse instance solving

---

1: **function** ComputePaths()
2:    **while** *elapsed time < timeout* **do**
3:       remove state $s$ from the front of OPEN;
4:       **if** $length(mand(s)) > 0$ **then**                                    $\triangleright$ [3]*
5:          insert the pair $< s, d(s) >$ to the training set
6:       **for all** $s' \in pred(s)$ **do**                                       $\triangleright$ [3]*
7:          **if** $g(s') > g(s) + c(s, s')$ **then**
8:             g(s') = g(s) + c(s, s')
9:             a(s') = s
10:             **if** $mand(s) = mand(s')$ **then**
11:                $d(s') = d(s)$                              $\triangleright$ [4]*
12:             **else**
13:                $d(s') = startNode(s)$                 $\triangleright$ [3,5]*
14:             insert $s'$ into OPEN with value $g(s')$       $\triangleright$ [6]*
15: **function** main()
16:    **for all** $i \in \{1, 2, ..., |V|\}$ **do**
17:       **for all** $s \in S$ **do**
18:          $g(s) = \infty$
19:       $g(E_i) = 0, a(E_i) = \varnothing, d(E_i) = \varnothing$
20:       $OPEN = \varnothing$
21:       insert $F_i$ into OPEN with value $g(E_i)$          $\triangleright$ [6]*
22:       ComputePaths();

---

[3]* $length(x)$: returns length of list $x$; $mand(x)$: returns mandatory nodes of instance $x$; $startNode(x)$: returns the start node of instance $x$; $pred(x)$: returns all possible predecessors of instance $x$

[4]* $s'$ is created from $s$ without adding its start node to the list of mandatory nodes

[5]* $s'$ is created from $s$ by adding its start node to the list of mandatory nodes

[6]* Heuristic $h = 0$

---

The data generated by the algorithm is added to the training set. We only retain instances which are at least 4 planning states apart. With our experimental setup, the algorithm reaches instances with up to 6 mandatory nodes in 10 hours. The GCN is then trained on this set with supervised learning. Results show that training on the synthetic data generated with A*, which is uniformly spread in the planning domain, enables the GCN to generalize well on instances with more mandatory nodes outside of A*'s scope. We argue this is because the distribution obtained with A* is related to the path length of resolved instances, and graph patterns already explored in short path solutions are incrementally included into longest ones.

### 3.3.4 Depth-First Branch and Bound Tree Search

An instance can be solved using the forward path-planning domain defined in §3.3.3 and a planning algorithm such as A*. It is also possible to solve it within a combinatorial search tree, and we explore this method of resolution next. To this end, we compute the shortest source-destination paths for every pair of nodes $\langle v_i, v_j \rangle$ in $\mathcal{G}$ using Dijkstra's algorithm, as well as the associated path cost. Solving an instance $I = (start, dest, M)$ then becomes equivalent to finding the optimal order in which the mandatory nodes in $M$ are visited for the first time. The solution path associated with an order $o = (m_1, m_2, m_3, ..., m_{q-1}, m_q)$ of the mandatory nodes can be built by concatenating the shortest path from $start$ to $m_1$, from $m_1$ to $m_2$, from $m_2$ to $m_3$, ... , from $m_{q-1}$ to $m_q$, and from $m_q$ to $dest$. Its total cost is the sum of the cost of each shortest path used to build it. A particular tree can be searched for identifying an optimal order of the mandatory nodes. For an instance $I$, we define the root of this tree as the $start$ node, every leaf node as the $dest$ node, and every intermediate tree node as a mandatory node in $M$, such that a path from the root of the tree to a leaf defines an order in which to visit the mandatory nodes in $M$. The cost of of transitioning from a node $v$ to a child node $v'$ in the tree is the cost of the shortest path in $\mathcal{G}$ between the pair $\langle v, v' \rangle$. In the following, we refer to this tree as the mandatory search tree. The combinatorial optimization problem associated with this tree displays a similar structure with the Travelling Salesman Problem; but it differs in the choice of the start and destination nodes which are fixed. It is an NP-hard problem.

The B&B algorithm is a popular tree search algorithm that is well known for its computational efficiency (Narendra and Fukunaga, 1977). In this work, we consider a depth-first B&B search algorithm. When developing a node inside the tree, the algorithm checks if each branch is expected to host a better solution than the best solution found so far. Should that not be the case for a given branch, the branch is cut, and the algorithm will not develop nodes further down the branch. This is done by using a lower bound and an upper bound. The lower bound is the sum of the total cost from the root node to the current node and a heuristic function $\zeta$ that approximates the remaining cost from the current node to the best achievable solution in branches below. This heuristic function should return a value as close as possible to this remaining cost (to cut as frequently as possible), while staying smaller (for the algorithm to remain optimal). The upper bound is initialized to $+\infty$ at first and then, each time a best known solution is found during the search, it is assigned its corresponding cost.

Next, we define the admissible heuristic function $\zeta$ that we use. Let $v \in M$ be a mandatory node in the mandatory search tree, and $R$ the set of remaining mandatory nodes left to the leaf node $dest$, *i.e.* the nodes in $M$ that haven't been included between the root node $start$ and $v$. Let $D = \{v\} \cup R \cup \{dest\}$. We define two functions, $min_1$ and $min_2$, that respectively return for a node $x$ in $D$ the lowest shortest path cost in $\mathcal{G}$ from $x$ to any node in $D \backslash \{x\}$, and the second lowest such cost. We build the heuristic function $\zeta$ by considering the remaining nodes left. For each node left, we consider the weight of all edges connecting it to other nodes in $D$ and add the first and second smallest such weights, with the exception of $v$ and $dest$ for which only the first smallest weight is added, and divide the total by 2:

$$\zeta(v) = \frac{1}{2}(min_1(v) + min_1(dest) + \sum_{r \in R} min_1(r) + min_2(r)) \qquad (3.11)$$

Recent progress in learning-assisted tree search has shown that machine learning can be used to narrow search space in very large domains to allow for efficient solving (Silver et al., 2016; Silver et al., 2017). Promising parts of the search tree can be visited first in accordance with the suggestions of a learning-based model, unpromising parts then, only if time allows. However, in a context where finding an optimal solution is critical, the search cannot be directed in such a way, as proof of optimality is required. Consequently, we keep the GCN out of the tree search procedure. For a given instance, we use the GCN recursively in order to obtain a suggested order of visit of the mandatory nodes (§ 3.3.2.2), from which we build the associated solution path by concatenating the shortest paths. This is done in negligible time compared to the tree search. The cost of the solution path found in such manner is then used as an initial upper bound for the B&B algorithm (Figure 3.6) instead of $+\infty$.

### 3.3.5 Experiments

#### Benchmarks and Baselines

We run experiments to evaluate the impact of the GCN's upper bound on the B&B method. Since proof of optimality is necessary in our context, we focus only on small-scale problems for which optimal solving is possible in reasonable time. We consider four different graphs, $\mathcal{G}_1$, $\mathcal{G}_2$, $\mathcal{G}_3$ and $\mathcal{G}_4$, with respectively 15, 23, 22 and 23 nodes. These graphs represent realistic AUGV crisis areas in which aid has to be provided to key points in operational areas (Guettier, 2007). We generate 1512, 2928, 2712 and 2712 random instances for each graph respectively, using the instance generator defined in Algorithm 1. We generate instances by groups with an increasing

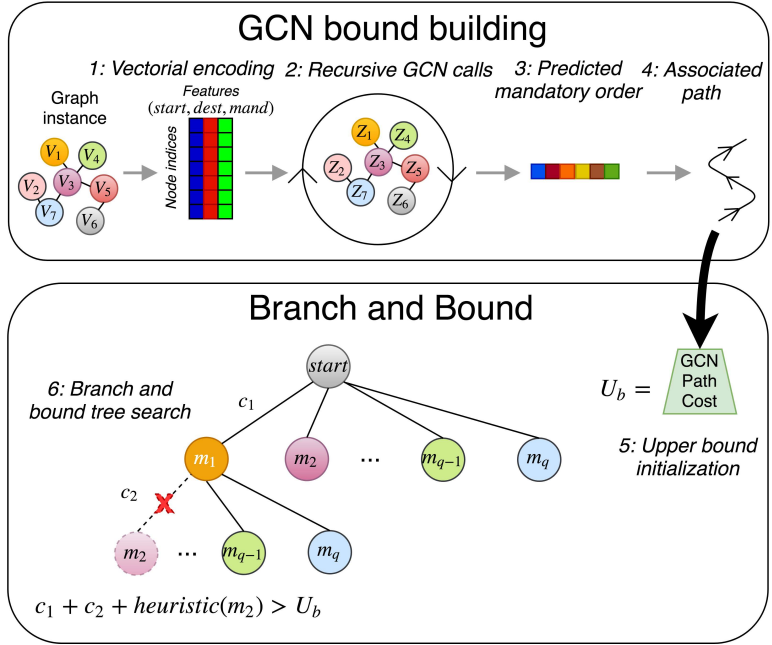

**Figure 3.6: GCN-assisted branch and bound algorithm pipeline.** The GCN is used recursively to build an order of visit for the mandatory nodes. The order is converted into a path using the previously computed shortest path pairs, and the cost of the path is used as an initial upper bound for the algorithm. Here, the upper bound of the GCN allows for a level 2 early cut.

cardinality for the set of mandatory nodes, ranging from 5 to 12.

We consider 4 different baseline solvers on the benchmark instances generated to compare solving performance. All solvers search for an optimal solution path. First, we use a solver based on dynamic programming (DP) which searches the mandatory search tree. Second, we run the B&B algorithm to search the mandatory search tree, both with and without the upper bound provided by the GCN. Lastly, we solve instances using forward A* applied on the planning domain described in section §3.3.3, with the minimum spanning tree (MST) as heuristic function. The MST heuristic is admissible and computed for an instance ($start$, $dest$, $M$) by considering the complete graph $\mathcal{G}'$, which comprises only the $start$, $dest$ and mandatory nodes $M$. All pairs of nodes ($v$, $v'$) in $\mathcal{G}'$ are connected by an edge which has a weight equal to the cost of the shortest path from $v$ to $v'$ in $\mathcal{G}$. The MST heuristic value is obtained by adding the following three values: the total weight of the MST of all mandatory nodes $M$ in $\mathcal{G}'$, the minimum edge weight in $\mathcal{G}'$ from the $start$ node to any node in the MST, and the minimum edge weight in $\mathcal{G}'$ from any node in the MST to the $dest$ node. We run experiments with a timeout of 5 minutes per instance.

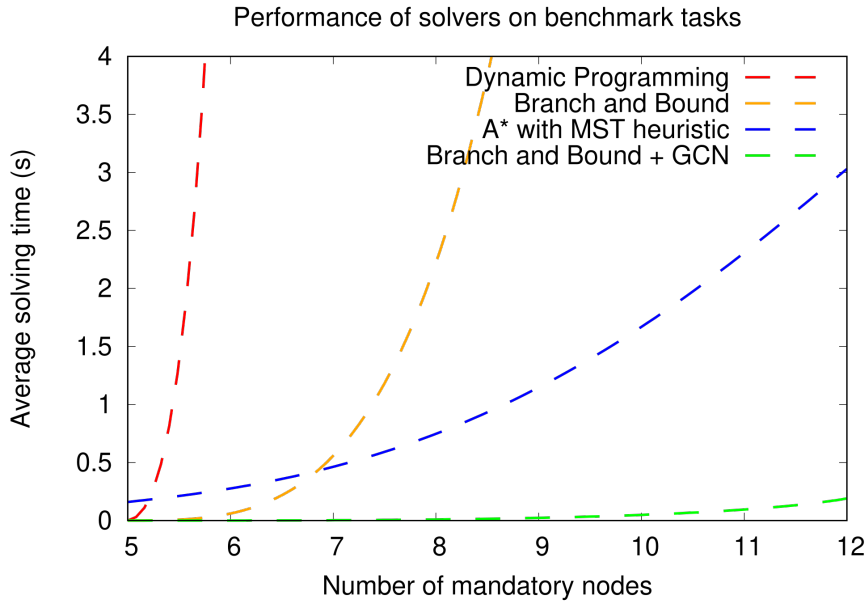

**Figure 3.7: Comparison of the performance of different solvers on benchmark instances generated for graph $\mathcal{G}_2$.** The $X$ axis represents the number of mandatory nodes of the instances, the $Y$ axis the average solving time. We limit the $Y$ axis to a range [0,4] to obtain a better comparison scale.

### Implementation Details

We choose TensorFlow to implement the GCN. We set the run-time of A* to 10 hours per graph for data generation. We use 3 graph convolutional layers of width 100. During training, we apply batch normalization (Ioffe and Szegedy, 2015) with decay of moving average $\varepsilon = 0.9$, dropout with *drop rate* of 0.1, and train the GCN with *Adam* (Kingma and Ba, 2015). We set the learning rate to $\eta = 10^{-4}$. We train models on a Tesla P100 GPU using over 1.5M training examples generated by A*, for ~5 hours. For each graph, we achieve relatively high cross-validation accuracy (94%, 90%, 91%, 89% for respectively $\mathcal{G}_1$, $\mathcal{G}_2$, $\mathcal{G}_3$ and $\mathcal{G}_4$). We conduct the benchmark tests in this section on a laptop with an Intel i5 processor and 8GB of RAM. We stress that the approach requires GPU only for faster training of the GCN, which is done offline. Problem instances can then be solved online on a CPU as is the case in our experiments.

**Table 3.5:** Experiments on graph $\mathcal{G}_3$. Legend: T/O = 5 minutes timeout.

| Mandatory #: | 5 | 7 | 9 | 11 | 12 |
|---|---|---|---|---|---|
| DP | | | | | |
| Avg. node visits: | 11.7K | 876K | 98.6M | T/O | T/O |
| Avg. time (s): | - | 0.26 | 30.52 | T/O | T/O |
| B&B | | | | | |
| Avg. node visits: | 418 | 4,94K | 146K | 11,1M | 120M |
| Avg. time (s): | - | - | 0.07 | 5.39 | 71.1 |
| A*, h=MST | | | | | |
| Avg. state visits: | 26.4 | 58.5 | 141 | 342 | 503 |
| Avg. time (s): | 0.14 | 0.29 | 0.73 | 1.84 | 2.70 |
| B&B + GCN | | | | | |
| Avg. node visits: | 148 | 1,24K | 10,8K | 161K | 642K |
| Avg. time (s): | - | - | 0.01 | 0.08 | 0.34 |

**Results**

We summarize results for $\mathcal{G}_2$ in Figure 3.7. We detail the experiments for graph $\mathcal{G}_3$ and $\mathcal{G}_4$ in Tables 3.5 and 3.6. We include solving time only if it is measurable by CPU clock time. Figures for all graphs show a similar trend. We note that for instances with seven mandatory nodes and more, best-first algorithms such as A* applied on the planning domain defined in §3.3.3 become more suited than depth-first algorithms such as B&B applied on the combinatorial search tree associated with the mandatory constraints. Although each planning state takes longer to compute in order to account for the specifics of the planning domain, overall significantly fewer planning states are visited than mandatory search tree nodes. This is because the planning domain takes advantage of the graph structure, which gives A* a significant edge over depth-first DP and B&B. We note, however, that when the upper bound of the GCN is used, the B&B algorithm is able to outperform A* on all instances, even the most complex ones, while scaling more smoothly with the number of mandatory nodes. In table 3.7 we provide additional insight on these results through information collected from the mandatory search tree for instances with 11 mandatory nodes. The average number of nodes processed in the subtree of each child node of the root node is given, as well as the average score of the best known solution after the subtree is processed.

**Table 3.6:** Experiments on graph $\mathcal{G}_4$. Legend: T/O = 5 minutes timeout.

| Mandatory #: | 5 | 7 | 9 | 11 | 12 |
|---|---|---|---|---|---|
| DP | | | | | |
| Avg. node visits: | 11.7K | 876K | 98.6M | T/O | T/O |
| Avg. time (s): | - | 0.24 | 30.53 | T/O | T/O |
| B&B | | | | | |
| Avg. node visits: | 545 | 7.58K | 193K | 11,8M | 122M |
| Avg. time (s): | - | - | 0.09 | 5.28 | 70.8 |
| A*, h=MST | | | | | |
| Avg. state visits: | 41.78 | 96.8 | 226 | 787 | 1,09K |
| Avg. time (s): | 0.15 | 0.44 | 0.96 | 3.18 | 4.36 |
| B&B + GCN | | | | | |
| Avg. node visits: | 233 | 2.46K | 29.1K | 235K | 859K |
| Avg. time (s): | - | - | 0.02 | 0.11 | 0.56 |

Since our B&B algorithm is depth-first, processing the entire subtree under the first child node of the root node when no initial upper bound is known is highly computationally expensive. Indeed, no cut can be made until a leaf node is reached, and even then, the identified solution is very likely to be costly compared to the optimal solution, thus the updated upper bound would still not allow for frequent cuts, until a good part of the subtree has been processed. On the other hand, if a good upper bound is known in advance, which is generally the case for the one given by the GCN in our experiments, the algorithm does not suffer from this issue, and early cuts can be made.

Lastly, we note that the GCN is trained on the data generated from the self-supervised algorithm detailed in § 3.3.3, for which instances have no more than 6 mandatory nodes and the data distribution relies on the A* planning domain. Still, it successfully generalizes to the experiment benchmark that is created with the instance generator defined in Algorithm 1 which concentrates instances with longer solution paths and more mandatory nodes.

**Table 3.7:** Information from the mandatory search tree.

| Root node child # | - | 1 | 2 | 3 | ... |
|---|---|---|---|---|---|
| B&B | | | | | |
| Avg. node visits | - | 11M | 50K | 30K | ... |
| Avg. best sol. score | ∞ | 10.35K | 9940 | 9781 | ... |
| B&B + GCN | | | | | |
| Avg. node visits | - | 74K | 11K | 7K | ... |
| Avg. best sol. score | 9890 | 9300 | 9264 | 9249 | ... |

**Discussion**

We introduced a method combining a GCN with a B&B tree search to solve the problem. We proposed a relevant self-supervised strategy based on A* to provide appropriate data for GCN training in the offline phase. Then, in the online phase, we accelerate optimal depth-first solving of the search tree associated with the mandatory constraints by leveraging the upper bound computed by the GCN. We showed that the GCN is able to successfully generalize to instances created with a different generator than the one used for its training. The use of the GCN's upper bound enables better scaling of the B&B algorithm onto more complex problems. We show that this speedup is significant and outperforms A* with a handcrafted heuristic function based on minimum spanning trees. The framework proposed here outperforms the first approach taken in § 3.2 and scales better as more mandatory nodes are imposed.

## 3.4  Conclusion

We formalized the first problem mathematically and introduced two approaches to solve it optimally. Each approach trains a GCN in the offline phase and uses it in the online phase to assist a different solver. In the first approach, we defined a CP-based framework to solve the problem. We used the CP solver to create learning data on which we trained a GCN to make, for each instance, a prediction on the first node to visit from the start node. The resulting solver, GCN + CP, was shown to be significantly faster than the baseline CP solver. Because of the way CP operates, we used GCN information only on the first node to visit (*i.e.* lower level information such as how to solve sub-instances of an instance is not included), arguably resulting in a performance limitation. We lifted this limitation in a second approach in which we had the GCN predict the next mandatory node to visit instead. By using recursive GCN calls, we were able to completely define an order of visit of mandatory nodes and its associated cost. We introduced a self-supervised training scheme based on A* to train the GCN. The cost of the GCN-induced order of visit of mandatory nodes is used as an initial upper bound in a B&B algorithm. The resulting framework, GCN + B&B, achieved significant performance gain versus the baseline B&B algorithm, and easily outperformed an A* solver with handcrafted heuristic. Furthermore, clear GCN generalization was achieved since gains were made on instances with more mandatory nodes and longer solution paths than the instances used for training. While GCN + B&B displays good performance and allows us to solve harder problems than GCN + CP, it still fails to solve instances with many mandatory nodes (*e.g.* $> 20$) in reasonable time. To address instances with a high number of mandatory nodes, we choose to explore approximate methods in the next chapter.

# Chapter 4

# Local Search with Approximation Methods

This chapter deals with the first problem again. However, unlike the previous chapter, we seek to provide a framework capable of solving the problem on a high scale (*i.e.* in large graphs, and with more mandatory nodes than experimented on in the previous chapter). Two approaches, inspired from Reinforcement Learning (RL) algorithms, are presented. The first one seeks to train a Graph Convolutional Network (GCN) with the value iteration algorithm. This approach does not yield satisfying results. The second approach, which trains a GCN via policy gradient by relying on a local search algorithm, achieves significantly better results.

## 4.1 Preliminaries

We refer the reader to the previous chapter (§3.1) for the mathematical definition of the problem tackled in this chapter. Methods presented in this section are also applicable to directed graphs (provided the spectral-based GCN architecture is replaced with a spatial-based one).

### Related Works

Related works using Graph Neural Networks (GNN) for approximate solving of planning tasks and combinatorial problems have emerged during this thesis and afterwards. (Li, Chen, and Koltun, 2018) present an approach which combines tree search, GNNs and a local search algorithm for approximate solving of some NP-hard problems such as the maximum independent set problem. Their approach substantially outperforms previous learning-based approaches, and performs on par with highly optimized state-of-the-art heuristic solvers. This approach is enhanced in (Xing and Tu, 2020) with Monte Carlo Tree Search (MCTS) and adapted to the Travelling Salesman Problem (TSP). A GNN approach for the decision version of the TSP is also presented in (Prates et al., 2019). The problem of graph colouring is recently studied in (Lemos et al., 2019) who propose a GNN-based approach which outperforms baselines on some graph distributions. As far as planning is concerned, (Ma et al., 2018) leverage a GNN for the selection of a planner inside a portfolio for STRIPS planning problems and outperform the previous leading learning-based approach based on a convolutional neural network (Katz et al., 2018). Lastly, (Silver et al., 2020) use a GNN to predict, for a given task in classical planning, which objects are most relevant to the task and require focus to find a plan more efficiently. The resulting approach is reportedly significantly faster than several baselines which include partial grounding strategies and lifted planners.

### Definitions and Notations

#### 2-OPT Algorithm

We give the pseudocode for the 2-OPT algorithm in Algorithm 8 in the appendix. 2-OPT is used in the following sections to optimize a given order of exploration of mandatory nodes for a given problem instance $I = (start, dest, M)$.

#### Notations

We use the same notations as in Chapter 3 except:

- $\pi$: A decision policy for the GCN, dependent on its parameters $\theta$.

## 4.2 Optimization with Graph Convolutional Networks and Value Iteration Reinforcement Learning

This section presents an approach which aims to train a GCN on a given graph in the offline phase by value iteration reinforcement learning. The idea is again to use the learned problem topology online for more efficient planning, but for bigger instances than previously. The intuition behind using value iteration is to propagate learned knowledge from small instances gradually to bigger ones, overcoming the previous challenge of being unable to solve big instances up front. We first introduce the approach chosen to solve problems, based on the A* algorithm. Second, we describe the architecture chosen for the GCN we use. Next, we detail our value iteration scheme for training the GCN on a graph, and explain how the GCN is coupled with A* to solve problems. Lastly, we discuss performance of this framework.

## Resolution Approach

To solve the problem, we consider the same planning domain as defined in §3.3.3. We remind the reader of the planning state transition rules in Table 4.1 for convenience.

**Table 4.1:** Transition rules to successors and predecessors for planning states.

| $s = (start, dest, M)$ | |
|---|---|
| Successor state $s'$ | Predecessor state $s'$ |
| $s' = (start', dest', M')$ | $s' = (start', dest', M')$ |
| $(start, start') \in \mathcal{E}$ | $(start', start) \in \mathcal{E}$ |
| $dest' = dest$ | $dest' = dest$ |
| $M' = M \backslash \{start'\}$ | $M' = M$ $\qquad \rhd\,{}^{1}*$ |
| - | or |
| - | $M' = M \cup \{start\}$ $\quad \rhd\,{}^{2}*$ |
| $s$ is not an end state | $s'$ is not an end state |
| Transition cost: $(start, start')$ | Transition cost: $(start', start)$ |

[1]$*$ only if $start' \notin M$
[2]$*$ only if $start \neq dest$

We remind the reader that in this planning domain, states are instances. Termination states or termination instances are denoted as $(v, v, \varnothing)$, $\forall v \in \mathcal{V}$. Transitions between states basically consist in 'moving' one step at a time in the graph $\mathcal{G}$ by incrementally changing the start node of the considered instance. More specifically, a transition from an instance $s$ corresponds to choosing one edge departing from the starting node of $s$. The cost of the transition is the cost of the edge. The resulting new state $s'$ is a sub-instance of $s$ for which the starting node is the node on the receiving end of the chosen edge; the destination node is the same as $s$; and the list of mandatory nodes is the same as $s$ minus the new starting node. To solve instance $I = (start, dest, M)$ optimally in the planning domain, one needs to find a plan which begins from the start state $(start, dest, M)$, ends at the goal state $(dest, dest, \varnothing)$, and for which the total cost is minimal. To compute a plan, we choose the A* algorithm. We aim to use the GCN as heuristic function $h$ for A*. The GCN heuristic is not admissible, causing loss of optimality guarantees. Nonetheless, we hope that use of the GCN as heuristic function could provide A* with guidance good enough to solve problems on a higher scale than with the previous optimal approaches.

## Graph Convolutional Network Architecture

Our GCN is comprised of graph convolutional layers from (Kipf and Welling, 2017). We refer the reader to §3.2.2 or §2.9 for more details on the convolution. We use the same encoding as in § 3.2.2.1 to encode a problem instance as a vector $\mathbf{x}$. We remind the reader that three features are used per node: *start, end, mandatory*.

Let $f$ be the function representing the GCN. Here, function $f$ takes as input an instance encoded as a vector $\mathbf{x}$ and stacks several graph convolutional layers interleaved with the ReLU$(\cdot) = \max(0, \cdot)$ piece-wise activation function. The last graph convolutional layer is followed by a fully-connected layer. The output is a single, real valued number $\hat{\mathbf{y}} \in \mathbb{R}$. We again refer to the parameters of $f$ as $\theta$: $f_\theta(\mathbf{x}) = \hat{\mathbf{y}}$. In this approach, we aim to teach the GCN to predict, for an input instance encoded as $\mathbf{x}$ the cost of the optimal path which solves this instance. To this end, the GCN is trained on labeled data, *i.e.* pairs $(x_i, y_i)$ obtained in a value iteration fashion described next. The *mean squared loss* $\mathcal{L}$ is used to train the neural network:

$$\mathcal{L}(\theta) = \frac{1}{m} \sum_{i=0}^{m} (\mathbf{y}_i - f_\theta(\mathbf{x}_i))^2 \tag{4.1}$$

### 4.2.1   Value Iteration Training Scheme

Here, we define a value iteration procedure to train the GCN on graph $\mathcal{G}$. First of all, we consider the planning domain defined previously in §4.2. We seek to teach the GCN to learn the value function $V$, which defines the total expected reward for a state $s$ under the optimal policy $\pi^*$ to solve problem instances. We take the following elements into consideration regarding the environment the agent is evolving in:

- Model-based: the graph is entirely known, therefore transitions are entirely defined.

- Full observability: All current states and transitions are fully visible.

- Deterministic actions: actions consist in moving from edges to edges in the graph, and they are entirely deterministic as knowledge of the graph is assumed to be correct.

- Rewards: We define penalties instead of rewards; the transition from a state $s$ to a state $s'$ leads to an immediate (positive) penalty equal to the weight of the edge linking the starting nodes of $s$ and $s'$.

By defining immediate penalties equal to the cost of edges taken by the agent, the purpose of the value iteration approach will be to minimize the total score of penalties obtained by the agent from a start instance $(start, dest, M)$ to the goal state $(dest, dest, \varnothing)$, *i.e.* minimize the total cost of the path in the graph associated with the plan.

In order to train the GCN, we keep a fixed size window $W$ in which we add training data of the form $(\mathbf{x}_i, \mathbf{y}_i)$; where $\mathbf{x}_i$ are instances and $\mathbf{y}_i$ the estimated associated V values. The size of the window being fixed, if maximum capacity is reached, the oldest data is removed. We define a new instance generator. This generator has higher odds of generating easy instances first (termination instances, or instances without, or with very few mandatory nodes) which are close to termination instances in the planning domain. Then, as the generator generates more and more instances, its odds of creating a higher percentage of hard instances increases. This is done by choosing a random number of mandatory nodes between a lower value and an upper value, and increasing these bounds every few instances generated.

When an instance $I$ is generated, we use the following value iteration equation to compute $V(s)$ where $s$ is the state associated to $I$:

$$V(s) = min_{s' \in succ(s)}(R(s, s') + \gamma f_\theta(s'))  \tag{4.2}$$

Where $succ(s)$ refers to the list of successors of state $s$ according to the rules defined in Table 4.1; $R(s, s')$ is the immediate penalty incurred from $s$ to $s'$, *i.e.* the cost of the edge linking the starting nodes in $s$ and $s'$; $\gamma \in ]0,1]$ a discount factor and $f_\theta(s')$ the estimation by the GCN of V(s'). If $s'$ is a termination instance, we replace $f_\theta(s')$ with 0 instead. Once $V(s)$ computed, we store the pair $(s, V(s))$ in W. Then, we perform a training step of the GCN on a batch of training examples $(\mathbf{x}_i, \mathbf{y}_i)$ taken randomly from $W$. The training step essentially performs a step of gradient descent on the mean squared loss for this random batch and updates the parameters $\theta$ so as to lower this loss. This *teaches* f to improve its predictions on the selected pairs $(\mathbf{x}_i, \mathbf{y}_i)$ that were randomly sampled from $W$. This process is repeated for every generated instance.

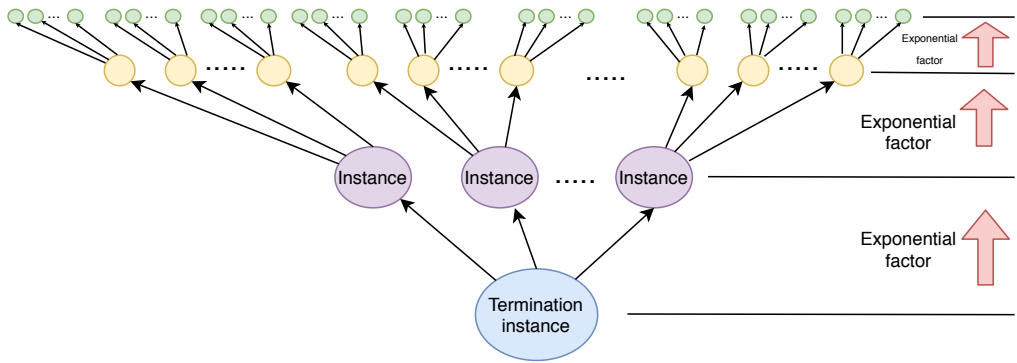

**Figure 4.1: Exponential factor of the planning domain.**

The fact that the generator first has a higher likelihood of generating easy instances (close to termination instances) allows the GCN to learn the V-value for those easy instances. This knowledge is then required for predecessor instances, as defined in Equation 4.2, which is why the generator slowly increases its chance of generating harder instances (farther from termination instances) as it generates more instances. The window training scheme, which allows the GCN to update its knowledge according to current and prior knowledge, all while discarding oldest, least optimal knowledge, is inspired from the replay memory in reinforcement learning applications such as (Mnih et al., 2013).

### 4.2.2 Experiments

We run the value iteration algorithm to train the GCN on a randomly generated graph $\mathcal{G}$ containing 100 nodes and 10% edge connectivity. After significant experimenting, we noticed that the GCN does not perform well on instances with more than 5 mandatory nodes. This leads the GCN to provide a heuristic $h$ to A* which is not only costly to compute at each state, but also provides highly inaccurate estimates of the optimal V-value for states.

To train the GCN, we tried several modifications to overcome this problem. We modified extensively the value of the $\gamma$ discount factor, originally set at $\gamma = 0.9$. We tried removing it as well (*i.e.* $\gamma = 1$) in order to avoid 'graph loops' with low cost which can induce the value iteration in error by looping continuously, all while discounting future V-values. We modified the instance generator extensively as well. We also tried to process instances in a reverse 'episode'-like fashion instead of using the previous generator: start with a termination instance and follow with predecessors recursively in order to allow better propagation of V-values. Despite our efforts, we were still unable to train the GCN so as to provide accurate V-values for instances with 5 or more mandatory nodes, resulting with a highly costly, unstable and inaccurate heuristic function for A*. We assume this problem comes from the inability of the GCN to generalize sufficiently to cope with the exponentially increasing number of states as we move farther from termination instances. This renders the value iteration algorithm unable to propagate information efficiently onto distant states from termination instances (as illustrated in Figure 4.1).

## 4.3 Optimization with Graph Convolutional Networks and Local Search via Policy Gradient Reinforcement Learning

In this section, we present an approach which relies on a local search algorithm to solve the problem. We devise a policy gradient reinforcement scheme reliant on the local search algorithm to train a GCN on a graph in the offline phase. The resulting framework uses the trained GCN to initialize a start point for the local search algorithm to solve problems in the online phase. This section is organized as follows. First, we present the local search algorithm and explain how it is used to solve problem instances. Second, we present the GCN architecture and explain how it is combined with the local search algorithm. Third, we detail the reinforcement learning algorithm used to train the GCN. Lastly, we conduct experiments to evaluate the impact of local search initialization by the GCN on solution compute time and solution quality.

### Resolution Approach

In this approach, we proceed similarly as in §3.3.1 to solve problem instances. We remind the reader the overall idea briefly. We start by computing all pairs of shortest paths between nodes and their associated cost in $\mathcal{G}$ using Dijkstra's algorithm (Algorithm 7 in the Appendix). Let $SPC : \mathcal{V}^2 \to \mathbb{R}$ be a function which returns the cost of the shortest path between two nodes. To solve any instance $(start, dest, M)$, we focus on optimizing the order of visit of the mandatory nodes in $M$. The optimized order, $M^*$, is used to rebuild the entire corresponding path by replacing each pair of nodes in the path $start - M^* - dest$ with the corresponding shortest path. Unlike previously, here we do not optimize $M$ with optimal branch and bound tree search. We use the local search algorithm 2-OPT instead, which has the advantage of being far more efficient at tackling complex problems where $M$ contains many nodes, at the cost of losing optimality guarantees. We explain next how 2-OPT is applied on instance $I = (start, dest, M)$. For complete pseudocode of how 2-OPT operates, we refer the reader to Algorithm 8 in the appendix.

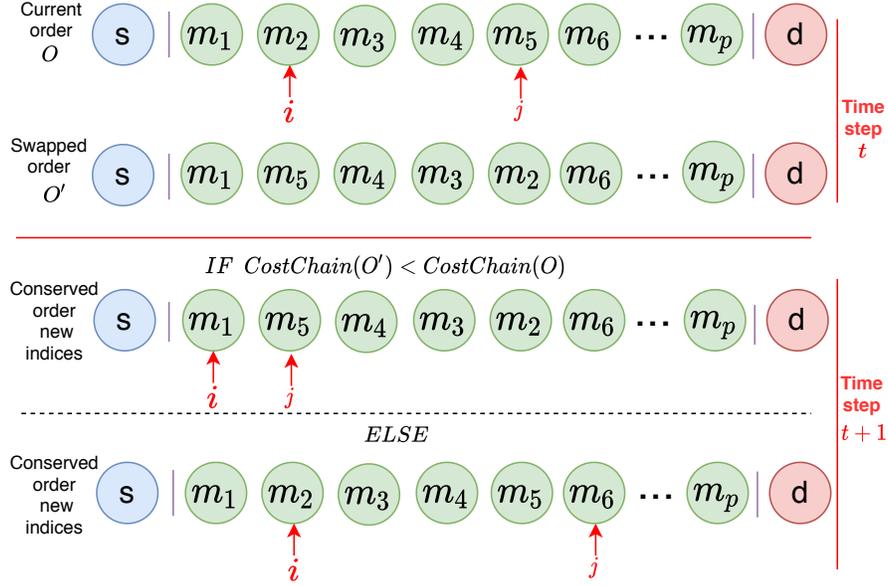

**Figure 4.2: Illustration of the 2-OPT algorithm on a problem instance.** Letter 's' in the blue circle is the abbreviation for start node, letter 'd' in the red circle for destination node. The '$m_i$' in the green circles are the mandatory nodes.

Suppose $M$ contains $p$ nodes. Let $M = (m_1, m_2, ..., m_p)$. We define the function $CostChain : \mathcal{V}^p \to \mathbb{R}$ as:

$$
\begin{aligned}
CostChain(m_1, m_2, ..., m_p) = & SPC(start, m_1) + \\
& SPC(m_p, dest) + \\
& SPC(m_1, m_2) + ... + SPC(m_{p-1}, m_p)
\end{aligned}
\tag{4.3}
$$

Essentially, 2-OPT works by swapping elements in $M = (m_1, m_2, ..., m_p)$ to decrease the cost provided by the *CostChain* function until a convergence criterion is met. More precisely, two indexes, $i$ and $j$, are used to create a nested loop. In this loop, $i$ starts at 0, $j$ at 1 (with indices starting from 0). At each iteration, elements between $i$ and $j$ are swapped symmetrically, and the new cost is compared to the old one with the *CostChain* function. Two possibilities exist:

- If the new cost is higher, the swap is undone, and $j \leftarrow j + 1$. If $j > p$, then $i \leftarrow i + 1$ and $j \leftarrow i + 1$. In the event that $i = p$, the algorithm has converged and is stopped.

- If the new cost is lower, the swap is kept and the new order conserved. Indices $i$ and $j$ are reset to 0 and 1 respectively.

### 4.3.1   Network Architecture and Solving Pipeline

We use graph convolutional layers from (Kipf and Welling, 2017) (§3.2.2, §2.9). We use the encoding in § 3.2.2.1 to encode a problem instance as a vector **x**, essentially using the following three features per node: *start, end, mandatory*. We refer as $f$ the function representing the GCN, $\theta$ its parameters. The main difference with the previous approach (§4.2) consists in the output of the GCN and what the GCN is taught to predict. Architecture-wise, this approach is the same as the one in §3.3.2. The GCN takes as input an instance **x**, stacks several graph convolutional layers (interleaved with the ReLU$(\cdot) = \max(0, \cdot)$ piece-wise activation function) and follows up with a fully-connected layer. The output is a vector $z \in \mathbb{R}^{|\mathcal{V}|}$. We use the *softmax* function (Equation 3.9) to convert $z$ into a list of probabilities $\hat{\mathbf{y}}$ for each node in $\mathcal{V}$. Here, the GCN is taught to predict, for an input instance encoded as **x**, which mandatory node should be visited first to lower the cost of the solution path. We refer to the GCN's decisions as the decision policy $\pi$ of the algorithm, which entirely depends on its parameters $\theta$. Training of the GCN is done in a policy gradient fashion, as explained in the next section, on pairs of labeled data $(x_i, y_i)$, where $x_i$ represents an instance and $y_i$ a corresponding mandatory node. GCN Training optimizes the cross-entropy loss defined in Equation 3.10.

Lastly, for any instance, we are able to define a complete order of visit of mandatory nodes by calling the GCN recursively: the first call defines the first mandatory node to visit and a sub-instance where this node is assumed to have been already visited, the second call (on the sub-instance) defines the second mandatory node to visit and another sub-instance, and so on. We refer the reader to §3.3.2.2 for more details. To solve any instance $I = (start, dest, M)$, our proposed framework proceeds in the following way. **(1)** Encode $I$ into a vector **x**. **(2)** Call the GCN recursively to obtain a suggested order of visit of mandatory nodes $M$. **(3 & 4)** Initialize and execute 2-OPT on the suggested order of visit. **(5 & 6)** Use the 2-OPT converged order of mandatory nodes to rebuild the complete associated path by replacing pairs of nodes with their shortest paths. Figure 4.3 illustrates the entire process.

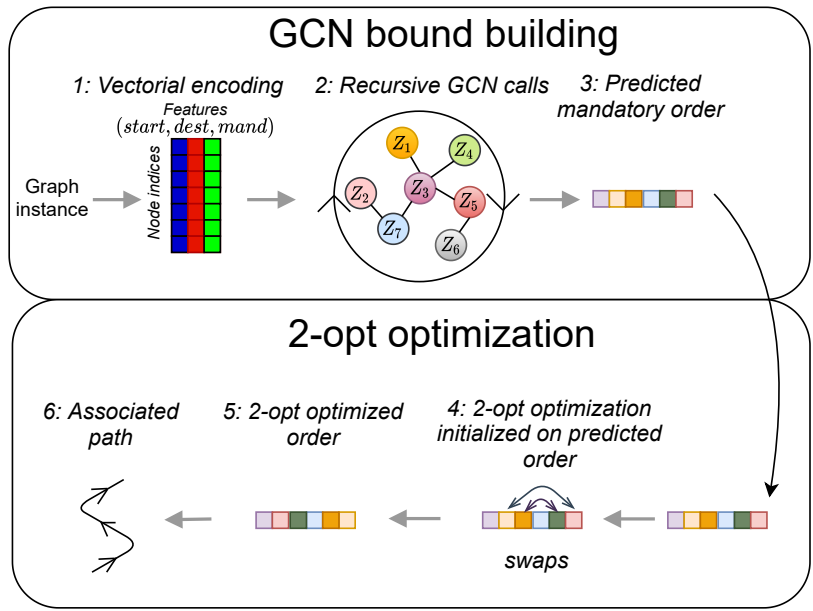

**Figure 4.3: Illustration of the GCN-initialized 2-OPT approach.** An instance is encoded into a vector. Recursive calls of the GCN define a suggested order of visit of mandatory nodes. We run 2-OPT on this suggested order until convergence. The converged order is used to build an associated path in the graph by linking shortest paths.

### 4.3.2 Policy Gradient Training Scheme

In this section, we present the policy gradient algorithm used to train the GCN on graph $\mathcal{G}$. The 2-OPT algorithm being deterministic (*i.e.* 2-OPT will always return the same converged order of mandatory nodes for a same input mandatory order), our framework's policy only depends on the parameters $\theta$ of the GCN. We would like to learn $\theta$ such that the output of the framework will be a solution path that can be obtained as quickly as possible and as optimal as possible.

The basic idea behind our training scheme is to allow the GCN to improve its knowledge by allowing it to interact with the 2-OPT algorithm. To this end, we consider two training 'directions'. First, we train the GCN in such a way as to output, for any input instance, an order of visit of the mandatory nodes that is as converged as possible in terms of 2-OPT iterations. This will make further 2-OPT optimization of this output faster. Second, we train the GCN to output, for any instance, an order with a cost as low as possible. Even if not optimal, this will increase the chances of further optimization by 2-OPT to converge into local minimas with lower cost.

We detail how the training scheme works. The following loop is repeated, until available time limit runs out. We generate an instance $I = (start, dest, M)$ randomly. We sample $start$, $dest$ and $M$ with a discrete uniform distribution (number of nodes in $M$ is also determined randomly). Next, we create two orders for nodes in $M$. The first order, $M_{rand}$, is obtained by shuffling nodes in $M$ randomly. The second order, $M_{GCN}$, is the order suggested by recursive calls of the GCN. We use 2-OPT on both $M_{rand}$ and $M_{GCN}$, and obtain converged orders $M_{rand}^*$ and $M_{GCN}^*$ respectively. We compare the cost of these two orders using the *CostChain* fonction. If $CostChain(M_{rand}^*) < CostChain(M_{GCN}^*)$, we select the order $M_{rand}^*$, otherwise we select the order $M_{GCN}^*$. The selected order of mandatory nodes, which we call $M_{selected}^*$, is broken down into instance - mandatory node pairs of training examples $(x_i, y_i)$ in a similar way as how the GCN is used recursively on an instance. More precisely, $x_0 \leftarrow encoding(I)$ and $y_0 \leftarrow M_{selected}^*[0]$, where $encoding$ designates the vector encoding of an instance. A sub-instance is created : $I' = (M_{selected}^*[0], dest, M \setminus \{M_{selected}^*[0]\})$, and $x_1 \leftarrow encoding(I')$, $y_1 \leftarrow M_{selected}^*[1]$. Another sub-instance is created $I'' = (M_{selected}^*[1], dest, M \setminus \{M_{selected}^*[0], M_{selected}^*[1]\})$ and $x_2 \leftarrow encoding(I'')$, $y_2 \leftarrow M_{selected}^*[2]$, and so on. We repeat the process until all pairs $(x_i, y_i)$ have been determined, and we add these pairs inside a window of training examples of fixed size (similar to the term *experience replay* often used in reinforcement learning). Oldest pairs of data are removed from the window if necessary to keep the size constant. Finally, $C_t(|M|)$ steps of gradient descent are performed to train the GCN, each step on a batch of training examples sampled randomly from the window. Function $C_t$ is a function that will return a higher number if there are many nodes in $M$ (since the number of pairs added to the dataset is equal to the number of nodes in $M$), lower if there are few nodes in $M$. This is to avoid over-training or under-training the GCN on the window, which is why we determine the number of training steps by looking at the number of training data added to the window. The whole process is illustrated in Figure 4.4, pseudocode is given in Algorithm 3.

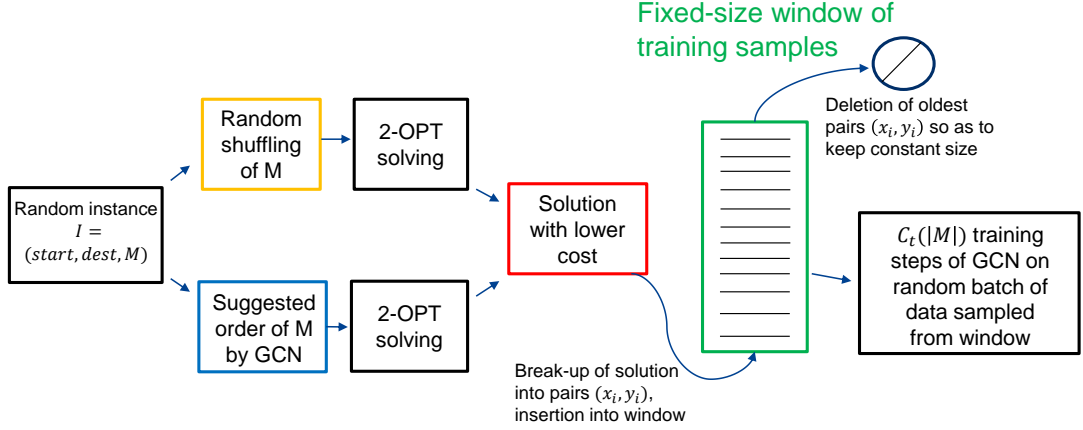

**Figure 4.4: Illustration of the policy-gradient training scheme.** Function $C_t$ returns a higher number of training steps the more there are nodes in $M$.

---

**Algorithm 3** Policy-gradient pseudocode

---

1: **function** PolicyGradient(Graph $\mathcal{G} = (\mathcal{V}, \mathcal{E})$, GCN $f$, Parameters $\theta$)
2:     $window \leftarrow \varnothing$
3:     **while** True **do**
4:         Generate random instance in $\mathcal{G}$ with uniform distribution $(start, dest, M)$
5:         $M_{GCN} \leftarrow GCNORDER((start, dest, M), \theta)$
6:         $M_{rand} \leftarrow Shuffle(M)$
7:         $I \leftarrow (start, dest, M_{GCN})$
8:         $I' \leftarrow (start, dest, M_{rand})$
9:         $(cost^*, M^*) \leftarrow 2OPT(I)$
10:       $(cost^*_{rand}, M^*_{rand}) \leftarrow 2OPT(I')$
11:       **if** $cost^* < cost^*_{rand}$ **then**
12:           $M_{better} \leftarrow M*$
13:       **else**
14:           $M_{better} \leftarrow M^*_{rand}$
15:       Add pairs <instance, mandatory node> to $window$ from $(start, dest, M)$ and $M_{better}$.
16:       $\theta \leftarrow AdamOptimizer(b, f, \theta)$             $\triangleright$ [1]*
17:     Return $\theta$

[1]* Train GCN on random batch $b$ selected from $window$ with Adam.

---

Intuitively, suppose that the policy of the GCN is $\theta_t$ at the start of the loop. If converged order $M^*_{rand}$ has a lower cost than converged order $M^*_{GCN}$, data from converged order $M^*_{rand}$ will be added to the window and learned by the GCN next. This will increase chances that the policy of the GCN at $\theta_{t+1}$ will allow it to output an order closer to $M^*_{rand}$ next time it encounters a similar instance to $I$. Thus, speed performance would be improved (as GCN output would be closer to a converged order) as well as solution quality (lower cost than if using $\theta_t$, which originally led to a converged order with higher cost). On the other hand, if converged order $M^*_{rand}$ has a higher cost than converged order $M^*_{GCN}$, data from converged order $M^*_{GCN}$ is added to the window and learned, improving only speed performance of policy $\theta_{t+1}$.

### 4.3.3  Experiments

In this section we evaluate the impact of the GCN on the performance of 2-OPT, both in terms of speed and solution cost. We generate 2 connected undirected graphs $\mathcal{G}_1$ and $\mathcal{G}_2$ with 1000 nodes and 10% edge connectivity (*i.e.* when creating the graph, each possible edge has a 10% probability of generation). Each edge is assigned a random integer weight between 1 and 100. On each graph, we generate a benchmark of instances categorized into groups with different number of mandatory nodes: 300, 500 and 700. Each group has 100 instances. Instances are created using the generator defined in Algorithm 1, differently from the uniform distribution used to sample instances in the training phase, limiting possible bias. We compare the performance of 2-OPT with random mandatory node initialization (baseline 2-OPT) against initialization given by the GCN (which we refer to as GCN+2-OPT). Tables 4.2 and 4.3 summarize experiments on $\mathcal{G}_1$ and $\mathcal{G}_2$ respectively.

We note similar trends in both graphs. The average cost of the solution found by GCN + 2-OPT is lower than the average cost of the solution found by baseline 2-OPT, and this translates into better solutions found by GCN + 2-OPT most of the time. Across all benchmark groups, for more than 90% instances, GCN + 2-OPT returns a solution with lower cost than baseline 2-OPT. Finally, computation times also significantly differ. For instances with 500 and 700 mandatory nodes, GCN + 2-OPT is able to compute a solution on average more than twice faster than baseline 2-OPT. Experiments highlight the capacity of the GCN to output orders which are close to 2-OPT convergence points (which directly translates into faster processing) with lower cost (hence, GCN + 2-OPT finds better solutions most of the time).

**Table 4.2:** Summary of experiments on graph $\mathcal{G}_1$. Costs and solving times are averages across all instances in a group. Method 2-OPT refers to using 2-OPT on a randomly initialized order of mandatory nodes, GCN+2-OPT on the GCN-initialized order.

| Benchmark | Method | Cost | Solving time (s) | Better Sol. found |
|-----------|--------|------|------------------|-------------------|
| $|M| = 300$ | 2-OPT | 1250 | 21 | 8% |
|  | GCN+2-OPT | 1198 | 14 | 92% |
| $|M| = 500$ | 2-OPT | 1839 | 115 | 3% |
|  | GCN+2-OPT | 1754 | 58 | 97% |
| $|M| = 700$ | 2-OPT | 2331 | 214 | 6% |
|  | GCN+2-OPT | 2268 | 98 | 94% |

**Table 4.3:** Summary of experiments on graph $\mathcal{G}_2$. Costs and solving times are averages across all instances in a group. Method 2-OPT refers to using 2-OPT on a randomly initialized order of mandatory nodes, GCN+2-OPT on the GCN-initialized order.

| Benchmark | Method | Cost | Solving time (s) | Better Sol. found |
|-----------|--------|------|------------------|-------------------|
| \|M\| = 300 | 2-OPT | 1614 | 14 | 4% |
| | GCN+2-OPT | 1488 | 11 | 96% |
| \|M\| = 500 | 2-OPT | 2273 | 85 | 5% |
| | GCN+2-OPT | 2156 | 41 | 95% |
| \|M\| = 700 | 2-OPT | 2841 | 173 | 5% |
| | GCN+2-OPT | 2704 | 84 | 95% |

**Implementation Details**

We select $C_t(x) = 10x$ and a window of size $50k$. We implement the GCN with TensorFlow. The GCN has 5 graph convolutional layers with 128 hidden features in each layer. We apply batch normalization (Ioffe and Szegedy, 2015) with decay of moving average $\varepsilon = 0.9$, dropout with drop rate set to 0.1, and we train the GCN using Adam optimizer (Kingma and Ba, 2015) with batches of size 64. Learning rate is set to $\eta = 10^{-4}$. We train a GCN for each graph by setting a training time limit of 1 week on a Tesla P100 GPU.

**Discussion**

We proposed a framework to solve problems with a high number of mandatory nodes approximately. The framework leverages a GCN to get a suggested order of visit of mandatory nodes for an input instance, and 2-OPT is called on the suggested order to further optimize it. The GCN is trained on a given graph $\mathcal{G}$ in the offline phase with a policy gradient reinforcement learning method which combines it jointly with 2-OPT to overcome the challenge of obtaining data on complex instances, the main barrier which prevented previous approaches from scaling. The proposed framework, with the trained GCN, can then be used in the online phase to solve instances more efficiently.

## 4.4   Conclusion

In order to solve instances of the first problem with a high number of mandatory nodes, we sought to train a GCN in the offline phase with reinforcement learning.

To this end, we explored two approaches: one based on value iteration, the other on policy gradient.

In the first approach based on value iteration, the GCN failed to generalize to instances with more than 5 mandatory nodes. We explain this failure by the fact that the exponential complexity linked to the number of mandatory nodes, which we were trying to avoid with reinforcement learning, ended up appearing in the planning domain used for value iteration learning, making it very hard to propagate information efficiently from easy instances to harder ones.

In the second approach, we used a local search algorithm, 2-OPT, to solve instances approximately. This effectively overcame the complexity issue, but instead brought the optimality issues of local search: the convergence point of 2-OPT depends on the initialization point. Therefore, we introduced a policy-gradient scheme to jointly train the GCN with 2-OPT. This step enables the GCN to provide a proper initialization point to 2-OPT. We showed in experiments that benefits provided by the GCN to 2-OPT are twofold. Firstly, the GCN allows 2-OPT to converge significantly faster: on the most complex instances, we obtained computation times more than twice faster for GCN + 2-OPT than baseline 2-OPT. Secondly, it allows convergence towards a solution that is significantly more likely to have a lower cost than baseline 2-OPT. This was the case more than 9 times out of 10 in our experiments, providing some relief to the drawback of local search algorithms which often get stuck in local optimas.

# Chapter 5

# Multi-Criteria Search with an Unknown Criterion

This chapter formalizes the second problem mathematically. A method is proposed to solve it. In this method, the autonomous feasibility, *i.e.* the unknown criterion, is simulated in the offline phase and learned with a machine learning model. In the online phase, predictions of the model are used to give an estimate of this criterion and perform multi-criteria optimization with Constraint Programming (CP). Experiments suggest that the proposed method achieves a good compromise between the two criteria to optimize.

## 5.1 Preliminaries

### Problem Formalization

We consider the second scenario problem presented in the context chapter. In this scenario, the considered vehicle is an Autonomous Ground Vehicle (AGV) and passengers are on board the vehicle. The AGV not being fully autonomous, one aim of the scenario consists in minimizing the number of required disengagements of autonomous driving so that passengers can focus on other tasks than driving. The AGV is also faced with typical itinerary computation which includes mandatory pass-by nodes, similar to the first problem. The objective consists in minimizing the total distance the AGV needs to travel to visit all mandatory nodes, as well as the number of times human driving intervention is required. This problem is a multi objective problem with mandatory pass-by point constraints. Previous related works include (Lust and Teghem, 2010) and (Manthey, 2012). We also refer to mandatory nodes as mandatory waypoints in this chapter.

We express the terrain as a weighted connected graph $\mathcal{G} = (\mathcal{V}, \mathcal{E})$. We assume the graph to be undirected, as AGV scenarios considered in experiments in this chapter correspond to undirected graphs. However, the resolution method proposed is applicable to both undirected and directed graphs. Each edge $e \in \mathcal{E}$ has a weight, which represents the distance of the corresponding path in the terrain. Additionally, edge $e$ also has an *autonomous feasibility* feature, equal to 0 if human intervention is required for the AGV to proceed on that path, 1 otherwise. This feature is unknown both in the offline phase and the online phase, and is linked to weather conditions. One objective is to learn to approximate this feature in the offline phase by simulation, in order to be able to predict it in the online phase and thus carry out multi-criteria optimization. Formally, the problem is mathematically defined as follows:

$$I = (start, dest, M)$$

- *start* is the start node the AGV begins at.

- *dest* is the destination node the AGV needs to arrive at.

- $M$ is the list of mandatory nodes the AGV needs to visit at least once along the way, in no specific order.

A solution path $\pi$ is an ordered list of nodes $\pi = \{start, v_1, ..., v_q, dest\}$, where $v_i$ are graph nodes, such that $M \subset \pi$. There is no limit to how many times a node or an edge can be included in a solution path. The associated total distance cost $D$ for $\pi$ is the sum of the distance cost of all edges included: $D = d_{start,v_1} + d_{v_1,v_2} + ... + d_{v_q,dest}$ where $d_{v,v'}$ is the weight of the edge linking $v$ and $v'$. The autonomous feasibility $P$ of $\pi$ is the sum of the autonomous feasibility for each edge included: $P = p_{start,v_1} + p_{v_1,v_2} + ... + p_{v_q,dest}$ where $p_{v,v'}$ is the autonomous feasibility of the edge linking $v$ and $v'$. The objective is to find a solution path that makes a compromise between minimizing $D$ and maximizing $P$. We seek a pareto-optimal solution, *i.e.* a solution for which neither $D$ nor $P$ can be improved without worsening the other. The problem is NP-hard (Lust and Teghem, 2010).

Lastly, we assume the autonomous feasibility criterion is mostly dependent on three parameters:

- Rainfall: heavy rain may cause new muddy areas to occur in the terrain, affecting

crossover duration between two points or increasing the risk of losing AGV platform control.

- Fog: heavy fog may cause a performance drop of the AGV's LIDAR, making it harder for the AGV to keep track of the surrounding environment.

- Wind: heavy winds may require moderate to high level of corrections in followed trajectories.

### Definitions and Notations

We refer the reader to §3.1 for definitions on constraint optimization and Dijkstra's algorithm. We use the same notations as in Chapter 3 with the following exceptions:

- $\mathbf{x}$: A vector of weather and terrain features for a given edge in the graph.

- $\hat{\mathbf{y}}$: The output of a multi-layer perceptron for a given input $\mathbf{x}$, valued between 0 and 1, representing the probability of the edge being autonomously feasible.

- $\mathbf{y}$: Autonomous feasibility of an edge in the graph, equal to either 0 or 1.

- $D$: Criterion representing the distance (to be minimized) for a solution path.

- $P$: Criterion representing the autonomous feasibility (to be maximized) for a solution path.

- $d_{vv'}$: Weight of the edge $(v, v')$, representing the distance.

- $p_{vv'}$: Autonomous feasibility of the edge $(v, v')$.

## 5.2 Multi-Criteria Constraint Optimization with Predicted Criterion

We propose a method to learn the autonomous feasibility of edges in the offline phase by examining weather and terrain conditions. Learned knowledge is then used in the online phase, making possible the search of a pareto-optimal solution which minimizes total distance and maximizes autonomous feasibility. To this purpose, we use a simulated AGV environment that includes a decision function for the autonomous feasibility as part of a higher-level environment representation. Computing the autonomous feasibility by running this function for online mission planning would require simulating the entire environment. This is impractical for real-time applications. Instead, in the offline phase, we train a Multi-Layer Perceptron (MLP) to approximate the autonomous feasibility. When planning the mission online, we use the trained MLP to make predictions of autonomous feasibility. We define a bi-criteria optimization strategy for a CP-based planner and use it to compute navigation plans. We conduct experiments on realistic scenarios and show that the proposed framework reduces the need for human interactions, trading with an acceptable increase of the total path distance.

### 5.2.1 Simulated Environment

First, we assign scales to the intensity of each weather parameter:

- $u_1 \in \{0, 1, ..., 10\}$ is the intensity of rainfall.

- $u_2 \in \{0, 1, ..., 10\}$ is the thickness of the fog.

- $u_3 \in \{0, 1, ..., 10\}$ is the strength of the wind.

A single weather configuration $\mathbf{x}_i$ is made of those three features, *i.e.* $(u_1, u_2, u_3)$.

The environment in which the AGV evolves is modelled in a 3D map. Vertices defined in $\mathcal{G}$ represent positions in the environment, while edges are represented by a set of continuous sub-positions (waypoints) linking vertices. Figure 5.1 shows a typical environment simulated by the 4d Virtualiz software in which the AGV proceeds. The simulation is realistic as it takes into account not only vehicle physics, but more importantly the vehicle's sensors, as well as core programs. Among such programs lie the environment-building functions, which enable the vehicle to build a state of its surrounding environment from its sensors. Key functions such as obstacle avoidance or waypoint-follow functions are also implemented in the simulator mimicking real life situations with high fidelity. We leverage the simulated environment to design and test our autonomous system in scenarios similar to real life conditions.

When a graph is defined, the simulator links it to the 3D map and several built-in functionalities become available. We can thus generate custom missions for the AGV by defining a start position, an end position, and a list of mandatory nodes. Additional environment parameters are available in such simulators and we consider a few representative ones in this work. In particular, we experiment with the rain variable $u_1$, the fog variable $u_2$ and the wind variable $u_3$. When sending the AGV on a defined mission $I = (start, dest, M)$ taking a path $\pi$, the simulating system will make the vehicle drive on a set of edges $e \in \pi$. Edges resulting in autonomous failure (vehicle getting stuck or taking longer than a set timeout threshold) $Q \subset \pi$ correspond to difficult sections of the path, which would require manual control of the vehicle. The criteria for autonomous feasibility depends on both current weather conditions and terrain structure and topology, as well as the vehicle's autonomous capabilities.

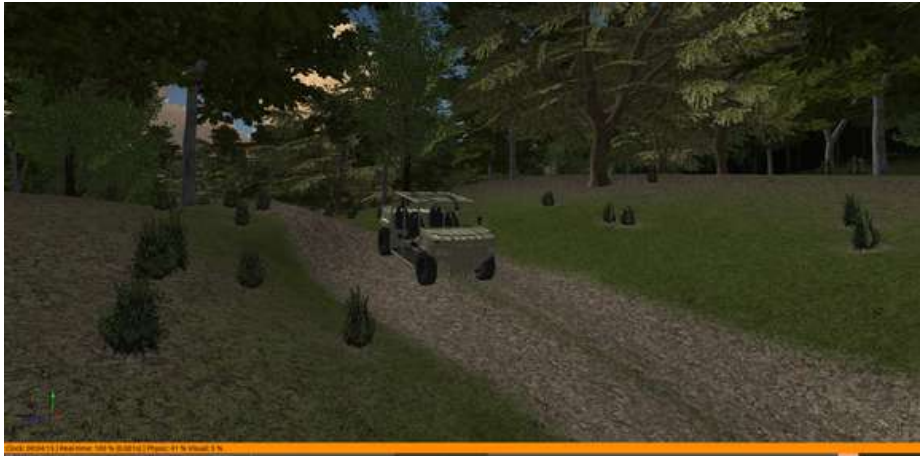

**Figure 5.1: An AGV evolving in an environment created by the 4d Virtualiz simulator.**

In order to learn to approximate the autonomous feasibility criterion, we use this simulated environment to create training data. To this end, we generate different weather conditions by changing the variables $u_1, u_2, u_3$ and send the vehicle on random missions. For a given set of values $u_1, u_2, u_3$ and a path $\pi$ that the vehicle has to follow, we retrieve the list of edges $Q \subset \pi$ which the simulator finds difficult for autonomous maneuvers. For each edge $e_i \in Q$, we store in a dataset a feature vector $\mathbf{x}_i = [u_1, u_2, u_3, \max(\text{slope}_{e_i}), \text{dist}_{e_i}]$ and $y_i = 0$. Variable $\max(\text{slope}_{e_i})$ is the maximum slope of edge $e_i$, $\text{dist}_{e_i}$ the distance of edge $e_i$ and $y_i$ is the preference label associated to $\mathbf{x}_i$. With $y_i = 0$, the edge should be avoided if possible when planning under these weather conditions. Similarly, for each edge $e_i \in \pi \backslash Q$, we store the value $\mathbf{x}_i$ and $y_i = 1$. These edges should be preferred when planning under these conditions. Regarding the structure of the terrain, we are unable to retrieve more information for an edge than only its distance and maximum slope. While this lack of information results in a limitation for learning performance, our experiments show that the performance of our framework remains satisfying.

### 5.2.2 Neural Network Training

In this section, we describe the machine learning model we train in the offline phase to simulate autonomous feasibility. More precisely, we train a MLP for identifying the difficult areas on the map using terrain and weather information and supervision from the decisions of the simulator concerning the respective sections on the map. Modern training practices for multi-layer networks usually allow reasonable approximations for a large variety of problems. With this in mind, we attempt to train a MLP to approximate the decisions of the 3D simulator over edges of the graph map.

Our MLP is made of 4 fully-connected layers (see Equations 2.1 and 2.2 for details) interleaved with the ReLU$(\cdot) = \max(0, \cdot)$ piece-wise activation function, except the last layer for which the sigmoid$(\cdot) = \frac{1}{1+e^{-\cdot}}$ activation is used. The MLP takes as input a vector $\mathbf{x}_i = [u_1, u_2, u_3, \max(\text{slope}_i), \text{dist}_i]$ and outputs a probability $\hat{y}_i$, valued between 0 and 1, of the edge $e_i$ being autonomously feasible or not. For predictions and path-planning, this probability is rounded to the closest integer 0 or 1 when classifying an edge under given weather and terrain conditions. We refer to our MLP as $f$ and its trainable parameters as $\theta$, *i.e.* $f_\theta(\mathbf{x}_i) = \hat{y}_i$

The simulator serves as *teacher* to the MLP, which learns here to mimic the simulator's decisions based on the path configuration and weather. Following the creation of the dataset in section (§ 5.2.1), we use the stored pairs of edges and labels $(\mathbf{x}_i, y_i)$ to train the MLP to correctly predict the label node $\hat{y}_i$. For supervision, we minimize the average binary cross-entropy loss $\mathcal{L}$ defined as follows, typically used for binary classification tasks:

$$\mathcal{L}(\theta) = \frac{1}{m} \sum_{i=0}^{m} [-\mathbf{y}_i \log(f_\theta(\mathbf{x}_i)) - (1 - \mathbf{y}_i) \log(1 - f_\theta(\mathbf{x}_i))] \tag{5.1}$$

where m is the number of training examples in the dataset, $\mathbf{y}_i$ the label (0 or 1) associated for a an edge in particular weather conditions, defined as $\mathbf{x}_i$. We train the MLP using stochastic gradient descent with momentum (Diederik P. and Jimmy, 2015). In our experiments, the MLP $f$ achieves an accuracy of 79% on the validation set. This is due to the lack of features which come into play in deciding the autonomous feasibility for an edge. In section (§5.2.4), this performance is tested and evaluated on new situations. In comparison, we also ran a logistic regression on the same training set, which achieved a validation accuracy of 71%. We believe some feature engineering may be necessary to provide slightly more relevant features for the logistic regression.

### 5.2.3 Constrained Multi-Criteria Optimization for Navigation and Maneuver Planning

We aim to compute an optimized navigation plan. In our approach, the navigation plan is represented as a path sequence of waypoints in a predefined graph. A "good" plan must minimize distance and maximize autonomous feasibility while satisfying mandatory waypoints.

Path planning is achieved using CP. The approach is similar to the CP method followed in §3.2.1, but adapted for multi-criteria optimization. We use a model-based constraint solving approach and cost objective functions (Equations 5.5 and 5.6). The problem is formulated in CP as a Constraint Optimization Problem (COP). Distance and autonomous feasibility are considered as primary and secondary cost objectives, respectively. We propose a multi-criteria optimization algorithm, based on global search, and adapted from branch and bound (B&B) techniques (Narendra and Fukunaga, 1977). As we proceeded for single-criterion optimization, we implement both COP formulation and search techniques with the CLP(FD) domain of SICStus Prolog library (Carlsson, 2015). The search technique is hybridized with a probing method (Guettier and Lucas, 2016), allowing automatic structuring of the global search tree. Here, probing focuses on learned and predicted autonomous feasibility in order to define an upper bound to the secondary metric. Probing takes as input predicted autonomous feasibility, builds up a heuristic sub-optimal path based on it as a preference, and lastly initializes the secondary cost criterion. The resulting algorithm is a Probe-based Constraint Multi-Criteria Optimizer denoted as PCMCO.

**Planning Model with Flow Constraints for Multi-Criteria Optimization**

The PCMCO elaborates a classical flow formulation with integrals, widely used in operation research (Gondran and Minoux, 1979). For a given path-planning problem $I = (start, dest, M)$, we use the same flow variables and equations constraints to model the instance as the single-criterion approach in §3.2.1.2 which we remind here for clarity. A set of flow variables $\varphi_e \in \{0, 1\}$, where $e \in \mathcal{E}$, models a possible path from $start \in \mathcal{V}$ to $dest \in \mathcal{V}$. A flow variable for an edge $e = (v, v')$ is denoted as $\varphi_{vv'}$. An edge $e$ belongs to the navigation plan if and only if $\varphi_e = 1$. The resulting navigation plan is represented as $\Phi = \{e \mid e \in \mathcal{E}, \varphi_e = 1\}$. From an initial position to a requested final one, path consistency is enforced by flow conservation equations, where $\omega^+(v) \subset \mathcal{E}$ and $\omega^-(v) \subset \mathcal{E}$ represent respectively outgoing and incoming edges from vertex $v \in \mathcal{V}$. Equation 5.2 ensures path connectivity and uniqueness while Equation 5.3 imposes limit conditions for starting the path at *start* and ending it at *dest*. Mandatory waypoints are imposed with Equation 5.4.

$$\sum_{e \in \omega^+(v)} \varphi_e = \sum_{e \in \omega^-(v)} \varphi_e \tag{5.2}$$

$$\sum_{e \in \omega^+(start)} \varphi_e = 1, \quad \sum_{e \in \omega^-(dest)} \varphi_e = 1, \tag{5.3}$$

$$\forall v \in M \sum_{e \in \omega^+(v)} \varphi_e \geq 1 \tag{5.4}$$

Here, contrary to the single-criterion approach, we consider two criteria. Path length is given by the metric (5.5), and we consider the path length as the primary optimization criterion $\mathcal{D}_{opt1} = \sum_{v \in \mathcal{V}} D_v$ to minimize. Likewise, the secondary criterion $\mathcal{P}_{opt2} = \sum_{v \in \mathcal{V}} P_v$ has the same formulation and is based on autonomous feasibility (5.6). Here, constants $p_{vv'}$ are edge preferences resulting from the predictions of the MLP $f$ on autonomous feasibility for each edge $e \in E$.

$$\forall v \in V, D_v = \sum_{v'v \in \omega^-(v)} \varphi_{v'v} d_{vv'} \tag{5.5}$$

$$\forall v \in V, P_v = \sum_{v'v \in \omega^-(v)} \varphi_{v'v} p_{vv'} \tag{5.6}$$

The global search technique underlying PCMCO guarantees completeness, as well as proof of completeness. It is based on the classical algorithmic components listed in §3.2.1.3 (variable filtering, tree search with backtracking and branch and bound). Within the branch and bound algorithm, the primary cost $\mathcal{D}_{opt1}$ drives the optimization loop. We extend the algorithm with preference optimization $\mathcal{P}_{opt2}$ to converge towards a pareto-optimal solution. At each iteration $k$, we impose that $\mathcal{P}_{opt2}^{k+1} \leq \mathcal{P}_{opt2}^k$ as a secondary optimization schema. This constraint is weaker than $\mathcal{D}_{opt1}^{k+1} < \mathcal{D}_{opt1}^k$, applied to the primary distance cost, which corresponds to the default operational semantic predicate *minimize* of the SICStus Prolog library. Value $P_{opt2}^0$ is initialized by probing with an arbitrary heuristic solution obtained with the Dijkstra algorithm. In this manner, branch and bound will favor learned preferences. This design demonstrates efficient pruning while ensuring search completeness. This in turn guarantees convergence towards a pareto-optimal solution.

**Table 5.1:** Multi-criteria experiments carried out on benchmark instances of graph $\mathcal{G}$. The first metric reported is the distance of the solution path, in meters. The second one is the number of human interventions required in the solution path. For both metrics, we compute the mean, median (med) and standard deviation (std) over all benchmark instances.

| Weather & Method: | Distance (m) | | | Interventions | | |
|---|---|---|---|---|---|---|
| | mean | med | std | mean | med | std |
| Fine weather | | | | | | |
| PCO | 4463 | 4246 | 840 | 1.6 | 2 | 1.1 |
| MLP + PCMCO | 4912 | 5016 | 954 | 0.1 | 0 | 0.3 |
| Moderate weather | | | | | | |
| PCO | 4166 | 4102 | 675 | 2.3 | 2 | 1.2 |
| MLP + PCMCO | 5431 | 5390 | 1003 | 0.4 | 0 | 0.6 |
| Difficult weather | | | | | | |
| PCO | 4207 | 4115 | 687 | 4.1 | 4 | 1.2 |
| MLP + PCMCO | 5153 | 5256 | 881 | 2.5 | 2 | 1.4 |

### 5.2.4 Experiments

For a given problem, minimizing the total distance of a solution path while maximizing autonomous feasibility are contradictory objectives requiring a compromise. This section carries two purposes. The first is to verify that the MLP $f$ is capable of making consistent predictions to avoid difficult edges. The second is to evaluate the compromise made by the CP-based solver described previously.

We generate 200 random benchmark instances associated to a connected undirected graph $\mathcal{G}$ (Guettier, 2007) that is representative of real scenarios for AGV search & rescue operations. Instances are generated with the generator described in Algorithm 1. The MLP is trained offline in simulations on this graph to predict the autonomous feasibility. We consider three different types of weather conditions: *fine*, *moderate* and *difficult*. For each weather type, we randomly select 50 instances, and we compare the solutions given by two solvers. The first solver is the reference probe-based constrained optimizer (PCO), which does not explore any preference criterion and only optimizes the distance. The second solver is the upgraded version with multi-criteria optimization (denoted as PCMCO). It takes into account the preference predictions of the MLP $f$ for current weather conditions. For each edge $e_i \in \mathcal{E}$, the preference prediction is obtained with a forward pass of the feature vector described in section (§5.2.2). We denote the resulting hybridization as MLP + PCMCO.

We study the influence of the edge preferences given by the MLP $f$ on the solution path. For each instance, we compute the solution path given by each solver. The total distance is then computed by summing the distances of all edges in the solution path. The solution path is also simulated in the 3D simulation environment to count the number of required human interventions. The human intervention count used in this section is a criterion which is opposite to the autonomous feasibility criterion, and should therefore be minimized. Results are averaged per instance and reported in table 5.1.

For fine weather conditions, the use of the MLP preferences enables MLP + PCMCO to almost never require human assistance in exchange for a 10% higher distance cost than PCO's. On the other hand, PCO requires more than 1 human intervention per instance on average. For moderate weather conditions, we see those gaps widening. A 30% higher distance cost allows MLP + PCMCO to require far less human interventions than PCO. Lastly, for difficult weather conditions, we observe that MLP + PCMCO incurs a 22% higher distance cost. While MLP + PCMCO requires far less human interventions than PCO, it still requires more than 2 human interventions on average. We assume this is due to the fact that difficult weather conditions cause a majority of edges to be difficult for autonomous driving. The solution path has no choice but to include some of those edges. This would also explain the lower distance cost increase than for moderate weather conditions.

Additionally, we run statistical tests to compare PCO and MLP+PCMCO and summarize them in table 5.2. Firstly, a paired sample t-test is done, for each weather condition, to compare the mean path distance given by PCO and MLP+PCMCO. The high t-values obtained, combined with very low p-values, indicate that the distance costs found by PCO and MLP+PCMCO differ significantly and that it is very unlikely to be due to coincidence. Secondly, a $\tilde{\chi}^2$ test is performed on the intervention count criterion for each weather condition. The high p-values observed validate the hypothesis that MLP+PCMCO acts independently of PCO in terms of autonomous feasibility.

These results highlight the fact that the MLP $f$ makes consistent predictions, and that MLP + PCMCO offers a good compromise between distance metric and autonomous feasibility.

### Implementation Details

We implement the MLP with TensorFlow. The MLP stacks 4 fully-connected layers. We apply batch normalization (Ioffe and Szegedy, 2015) with decay of moving average $\varepsilon = 0.9$, dropout with drop rate set to 0.1, and use Adam optimizer (Diederik P. and

**Table 5.2:** Statistical tests on benchmark results. The paired sample t-test is run on the distance criterion, while the $\tilde{\chi}^2$ test is run on the intervention count criterion.

| Test Method | Paired t-test | | $\tilde{\chi}^2$ test | |
| --- | --- | --- | --- | --- |
| | t-value | p-value | $\tilde{\chi}^2$-value | p-value |
| Fine weather | 7.28 | $10^{-9}$ | 19.1 | 0.99 |
| Moderate weather | 8.18 | $10^{-9}$ | 19.4 | 0.89 |
| Difficult weather | 6.51 | $10^{-7}$ | 9.37 | 0.99 |

Jimmy, 2015) with batches of size 64. Learning rate is set to $\eta = 10^{-4}$. All experiments are done without GPU, on a laptop with an Intel i5 processor and 8GB of RAM.

## 5.3 Conclusion

The proposed method successfully answers the issues raised in the second problem, where the distance criterion needs to be minimized, while the autonomous feasibility criterion, which is uncertain, has to be maximized. Our approach proposes offline learning into a model of autonomous feasibility using a simulation environment. We introduced a CP-based algorithm which takes into account the model's prediction of autonomous feasibility and compromises between both criteria. Experiments suggest the proposed framework is capable of finding in the online phase a good compromise which offers a higher autonomous feasibility for an acceptable increase in distance cost. AGV crew could benefit from such an approach in situations where their workload needs to be reduced.

# Chapter 6

# Scheduling under Uncertainty and Disjunctive Constraints

This chapter deals with the third problem. In this problem, online reactive scheduling is possible and the objective is to find a reactive scheduling strategy which takes into account the uncertain movements of the partner vehicle. We formalize the problem mathematically first and give necessary definitions. Next, we present a novel time-variant type of controllability for reactive scheduling, along with a method based on tree search and Graph Neural Networks (GNN) guidance to synthesize reactive scheduling strategies. In this particular problem, no offline preparation is possible since terrains and scenarios are not known ahead of time. Therefore, we provide a framework with a pre-trained GNN which can be used online in any terrain and circumstances. Experiments on known benchmarks suggest the framework is almost always capable of finding reactive strategies in the presented controllability type when state-of-the-art solvers can in Dynamic Controllability (DC), but significantly more efficiently.

## 6.1 Preliminaries

### Problem Formalization

We consider the third problem presented in the context chapter, where an Autonomous Unmanned Ground Vehicle (AUGV) needs to perform synchronized maneuvers with a partner manned vehicle. Maneuvers need to be executed in such a way as to guarantee convoy safety. Movements of the AUGV are controllable time variables $a_i \in A$, movements of the partner vehicle are uncontrollable time variables $u_i \in U$. Each variable $u_i$ is triggered in some interval of time after a given $a_i$, corresponding to the partner vehicle movement resuming after ensuring the AUGV is stopped. Additionally, time constraints binding $a_i$ and $u_i$ impose that the AUGV can only move in a given interval after the partner vehicle and ensure the partner vehicle is stopped when the AUGV is moving. Additionally, time constraints can bind $a_i$ to set time intervals for safety to ensure the AUGV is stopped at critical moments during the mission. Such constraints can incorporate disjunctions, as is the case with the scenario in the third problem. The AUGV is able to adapt its maneuvers over time as it observes how the partner vehicle moves, and needs to pre-compute a reactive scheduling strategy before the start of the maneuvers which guarantees all constraints will be satisfied if the strategy is followed. This problem is mathematically known as Disjunctive Temporal Network with Uncertainty (DTNU). Currently, the only solving approach for DTNUs in the reactive scheduling controllability type known as DC is based on Timed-Game Automatas (TGA) (Cimatti, Micheli, and Roveri, 2016; Cimatti et al., 2014). For complete background on DTNUs, existing controllability types of scheduling and related works, we refer the reader to §2.6. Here, we give a brief definition of DTNUs. A DTNU $\Gamma$ is defined as follows:

$$\Gamma = \{A, U, C, L\}$$

where:

- $A = (a_1, a_2, ..., a_n) \in \mathbb{R}^n$ is a set of real controllable timepoint variables, which can be scheduled at any moment in time.

- $U = (u_1, u_2, ..., u_q) \in \mathbb{R}^q$ is a set of uncontrollable timepoints variables.

- Each uncontrollable timepoint $u_j \in U$ is linked to exactly one controllable timepoint $a_i \in A$ by a contingency link $l \in L$: $l = (a_i, \vee_{k=1}^{q'} [x_k, y_k], u_j)$. This means that $u_j$ can occur on its own anywhere in the intervals of time $[x_k, y_k]$ after $a_i$ has been scheduled.

- $C$ is a set of free constraints, each of the form $\vee_{k=1}^{q''} v_{k,j} - v_{k,i} \in [x_k, y_k]$, where $v_{k,j}, v_{k,i} \in A \cup U$. All constraints need to be satisfied.

## Notations

We use the following main notations.

- $a_i$: A controllable timepoint; $A$ is the set of all controllable timepoints.

- $u_i$: An uncontrollable timepoint; $U$ is the set of all uncontrollable timepoints.

- $v$: A timepoint, either controllable or not; $V$ is the set of all timepoints.

- $C$: A set of time constraints binding timepoints, called free constraints.

- $L$: A set of contingency links, linking possible occurrence times of uncontrollable timepoints to scheduling times of controllable ones.

- $[x, y]$: A time interval for which the lower bound is $x$ and upper bound $y$.

- $\gamma, \Gamma$: Used to refer to DTNUs.

- $\Delta_t$: A duration of time which depends on current time step $t$.

- $\mathcal{G} = (\mathcal{K}, \mathcal{E})$: Graph conversion of a DTNU.

- $X_\kappa$, $X_\epsilon$, $X_\rho$: Respectively matrices of node features, adjacency and edge features of $\mathcal{G}$.

## 6.2   Tree Search of a Time-based Dynamic Scheduling Strategy

We study strategy synthesization of DTNUs as a search problem, express states as graphs, and use GNNs known as Message Passing Neural Networks (MPNN) to learn heuristics based on solved DTNUs to guide search. MPNNs (§ 2.13) are a spatial-based GNN architecture, in opposition to the GCN architecture used previously which is spectral-based. Fundamentally, we motivate this choice primarily because expressing DTNUs as graphs requires taking into account multiple edge features, which spectral-based approaches do not support. The key contributions of this approach are the following. **(1)** We introduce a Time-based form of Dynamic Controllability (TDC & R-TDC), time discretization rules, and a tree search approach to identify R-TDC strategies. **(2)** We describe a MPNN architecture for handling DTNU scheduling problems and use it as heuristic for guidance in the tree search. Moreover, we define a self-supervised learning scheme to train the MPNN to solve randomly generated DTNUs with short timeouts to limit search duration. **(3)** We introduce constraint propagation rules which enable us to enforce time domain restrictions for variables in order to ensure soundness of strategies found. We carry out experiments comparing the ability of the tree search algorithm to find R-TDC strategies against the ability of the state of the art DC solver (PYDC-SMT from (Cimatti, Micheli, and Roveri, 2016)) to find DC strategies. We show that the tree search is almost always able to find a R-TDC strategy when PYDC-SMT finds a DC strategy. Moreover, it finds the R-TDC strategy significantly more efficiently, resulting in 50% more DTNU instances solved in the same time budget on a known benchmark. Lastly, we show that the learned MPNN heuristic considerably improves the tree search on harder DTNUs: performance gains with the heuristic go up to 11 times more instances solved within the same time frame than the baseline tree search. Results also highlight that the MPNN, which is trained on a set of solved DTNUs, is able to generalize to larger DTNUs.

### 6.2.1 Problem and Controllability Definitions

Complexity of DC checking for DTNUs is *PSPACE*-complete (Bhargava and Williams, 2019). To alleviate this, we introduce R-TDC, a new form of reactive scheduling. There are multiple reasons behind this choice. R-TDC is based on time discretization, and we assumed that proper discretization of time can lessen the complexity of the search space. Furthermore, R-TDC allows to consider the search space as a tree structure, which makes it easier to use learning algorithms for improved efficiency. We start by giving necessary definitions for this work and introducing TDC and R-TDC.

**Definition 1** (DTNU and variants). *A DTNU $\Gamma$ is a tuple {A,U,C,L}, where: A is a set of controllable timepoints; U a set of uncontrollable timepoints; C a set of free constraints, each of the form $\vee_{k=1}^{q} v_{k,j} - v_{k,i} \in [x_k, y_k]$, for some $v_{k,j}, v_{k,i} \in V = A \cup U$, $x_k, y_k \in \mathbb{R} \cup \{-\infty, +\infty\}$ and $q \in \mathbb{Z}^+$; L a set of contingency links, each of the form $\langle a_i, \vee_{k=1}^{q'} [x'_k, y'_k], u_j \rangle$ where $a_i \in A$, $u_j \in U$, $x'_k, y'_k \in \mathbb{R} \cup \{-\infty, +\infty\}$, $0 \leq x_k \leq y_k \leq x_{k+1} \leq y_{k+1} \, \forall k = 1, 2, ..., q' - 1$ and $q' \in \mathbb{Z}^+$, indicating possible occurrence time intervals of $u_j$ after $a_i$. A DTN is a DTNU without uncontrollable timepoints, an STNU a DTNU without disjunctions in free constraints and contingency links, an STN an STNU without uncontrollable timepoints.*

**Definition 2** (DC & TDC). *DC is a reactive form of scheduling which incorporates occurrences of uncontrollable events as they unfold and adapts to them. A problem is DC if and only if it admits a valid dynamic strategy expressed as a map from partial schedules to Real-Time Execution Decisions (RTEDs) (Cimatti, Micheli, and Roveri, 2016). A partial schedule represents the current scheduling state, i.e. the set of timepoints that have been scheduled or occurred so far and their timing. RTEDs allow for two possible actions: **(1)** The wait action, i.e. wait for an uncontrollable timepoint to occur. **(2)** The $(t, \mathcal{X})$ action, i.e. if nothing happens* *before time t, schedule the controllable timepoints in $\mathcal{X}$ at t. A strategy is valid if, for every possible occurrence of the uncontrollable timepoints, controllable timepoints get scheduled in a way that all free constraints are satisfied. A **TDC** strategy is a representation of a DC strategy as a timed tree,* i.e. *a map from tree nodes to children nodes. Tree nodes represent partial schedules, and their children lead to the execution of one of the following actions: **(1)** Schedule a set of controllable timepoints at current time; **(2)** Wait a period of time or until an uncontrollable timepoint occurs, whichever happens first.*

**Definition 3** (R-TDC). ***R-TDC** is a subset of TDC. A R-TDC strategy is a finite tree. In particular, actions associated to a partial schedule in R-TDC are: **(1)** Schedule a set of controllable timepoints at current time; **(2)** Wait an **uninterruptible** period of time, the wait duration being defined by time discretization rules in §* 6.2.3.2.

A TDC strategy can fully express a DC strategy which has an infinite number of mappings from partial schedules to RTEDs, given an infinite tree. In this work, we restrict TDC to a finite search space, R-TDC, and weigh how loss of search completeness results in increased efficiency. The restrictions in R-TDC stem from the uninterruptible waits: occurrence times of uncontrollable timepoints happening in waits are only bounded. Thus, partial schedules in R-TDC tree nodes do not carry exact occurrence times for uncontrollable timepoints which already occurred but only occurrence intervals. Moreover, time discretization rules are used in R-TDC to inspect a partial schedule in order to define a wait duration. The aim is to maximize the duration to speed up strategy search while limiting loss of possible strategies. Lastly, in order to improve completeness, we augment R-TDC waits with the possibility of instantaneous reactive executions **during strategy execution**. These are requests made to the waiting controller agent in charge of executing the strategy to immediately execute some controllable timepoint(s) when it observes an uncontrollable timepoint occur, as described in § 6.2.3.3. Associated waits remain uninterruptible during strategy search however, resulting in only bounded and not exact scheduling times of controllable timepoints which are executed in such fashion.

**Definition 4** (R-TDC strategy structure). *A R-TDC strategy is a tree. This tree is comprised of a list of nodes $(N_1, N_2, ..., N_{q-1}, N_q)$. Each $N_i$ is of the form:*

$$N_i = (N_j, O_{ji}, E_i, \langle s_i, e_i, R_i \rangle)$$

*where:*

- $N_j$ *is the parent node of $N_i$ in the tree.*

- *Time $s_i$ is the start time of the wait in node $N_i$.*

- *Time $e_i$ is the end time of the wait in node $N_i$.*

- *$O_{ji}$ is the list of uncontrollable timepoints assumed to occur during the wait in node $N_j$. There exist as many $O_{ji}$ as the number of combinations of uncontrollable timepoints that may occur during the wait in node $N_j$. Therefore, in a R-TDC strategy, node $N_j$ will have exactly the same number of children nodes to account for all possible outcomes of uncontrollable timepoints.*

- *$R_i$ is a mapping which can associate to any uncontrollable timepoint that may occur in the wait $\langle s_i, e_i \rangle$ a set of controllable timepoints to reactively execute by the agent during strategy execution. The associated wait remains uninterrupted during strategy search even if some uncontrollable timepoints are assumed to occur in the wait.*

- *$E_i$ is a set of controllable timepoints to schedule at $s_i$.*

*Each path from the root node of a R-TDC strategy to any leaf node satisfies the following properties:*

- *It covers the time horizon entirely (each new wait starts at the same time as the end of the previous wait).*

- *It represents a unique outcome of the occurrence possibilities of uncontrollable timepoints. Each uncontrollable timepoint is bounded in an occurrence interval $\langle s_i, e_i \rangle$. All possible outcomes are included in the strategy.*

- *It assigns to every controllable timepoint a given time of scheduling (or time interval for those reactively executed in response to uncontrollable timepoints).*

- *All constraints are satisfied given the scheduling time, scheduling time intervals and occurrence intervals of all timepoints.*

We explain next how a R-TDC strategy is executed.

#### 6.2.1.0.1 R-TDC Strategy Execution.

A R-TDC strategy is executed in the following way by a controller agent. The agent starts at the root R-TDC node. For each current node $N_i = (N_j, O_{ji}, E_i, \langle s_i, e_i, R_i \rangle)$, it executes at $t = s_i$ the timepoints in $E_i$, and waits from time $s_i$ to $e_i$ with the reactive strategy $R_i$, *i.e.* if $R_i$ stipulates it, the agent will immediately execute some controllable timepoints in response to some uncontrollable timepoints that may occur during the wait, as soon as they do. At the end of the wait, the agent deduces from the list of uncontrollable timepoints that occurred which child $N_i'$ of node $N_i$ it transitioned to. It moves to $N_i'$ and repeats the same process. Those guidelines are followed recursively until all constraints are satisfied.

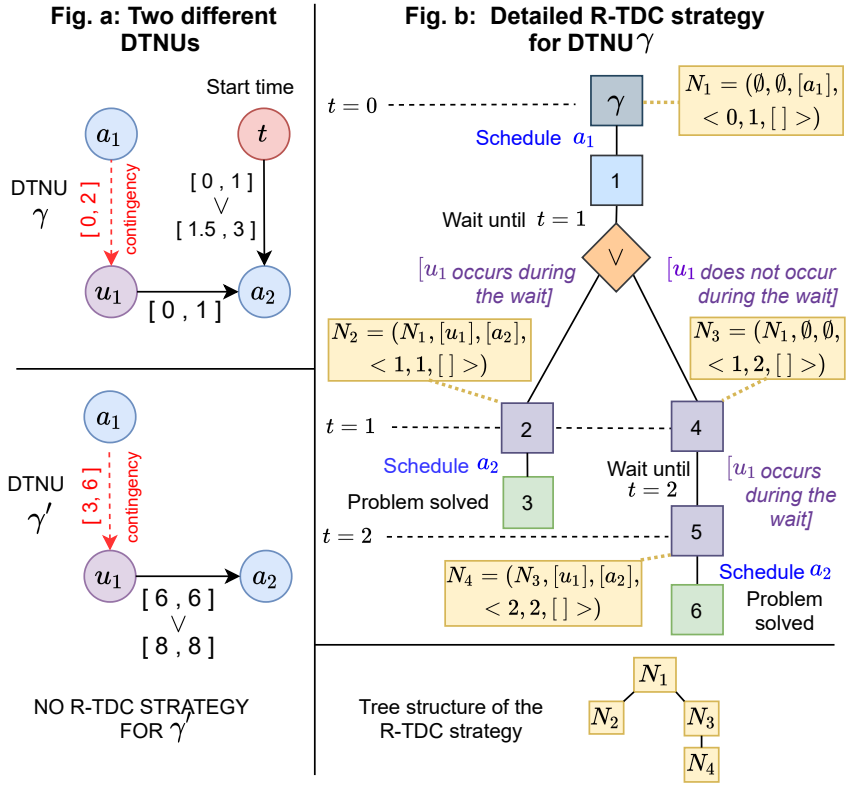

**Figure 6.1: Two example DTNUs $\gamma$ and $\gamma'$.** Timepoints $a_1$ and $a_2$ are controllable; $u_1$ is uncontrollable. Black arrows and their intervals represent time constraints between timepoints; red arrows and their intervals contingency links. A detailed R-TDC strategy is displayed for $\gamma$. Squares below $\gamma$ are sub-DTNUs; the $\vee$ sign lists transitional possibilities. Nodes $N_i$ are R-TDC strategy nodes as defined in the definition section.

We give a simple example of a R-TDC strategy for a DTNU $\gamma$ in Figure 6.1. DTNU $\gamma'$ on the other hand is an example of a DTNU which is DC but not R-TDC. More precisely, it shows a clear limitation of R-TDC: when a controllable timepoint $a$ absolutely has to be scheduled a set time after an uncontrollable timepoint $u$ occurs: $a - u \in [x, x], x \in \mathbb{R}^+$. This is impossible in R-TDC as occurrence time of $u$ can only be bounded during strategy search and not exact, because any wait interval, however small, in which $u$ is assumed to occur is bounded. Finally, in Figure 6.2, we give a simplified example of a R-TDC strategy for the example DTNU from (Cimatti,

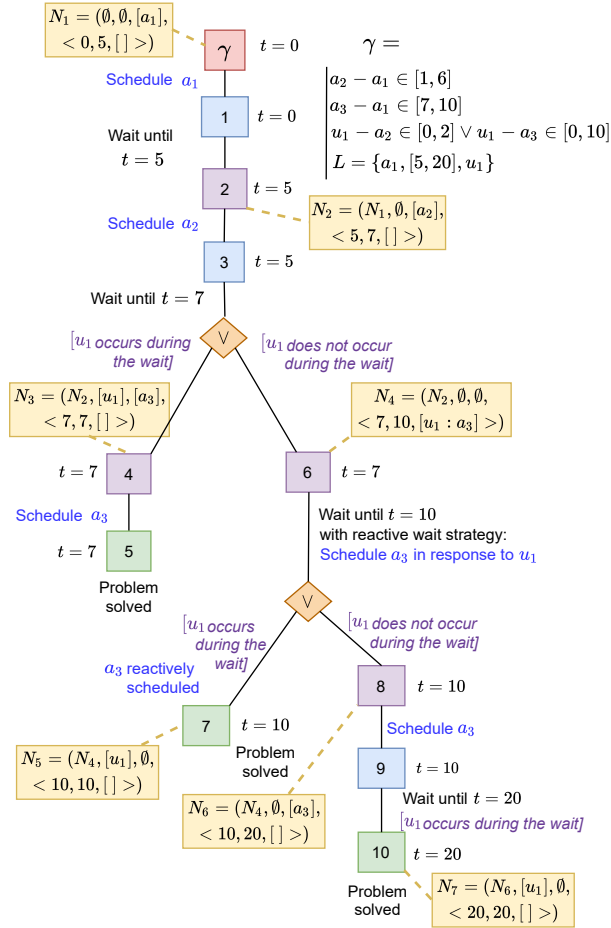

**Figure 6.2: Detailed R-TDC strategy of a DTNU $\Gamma$.** The red square $\gamma$ is the original DTNU. Other squares are sub-DTNUs, except the $\vee$ signs which list transitional possibilities. Nodes $N_i$ represent the R-TDC strategy nodes as explained in the definition section.

### 6.2.2 Tree Search Preliminaries

We introduce here the tree search algorithm. The root of the search tree built by the algorithm is a DTNU, and other tree nodes are either sub-DTNUs or logical nodes (*OR, AND*) which respectively represent decisions that can be made and how uncontrollable timepoints can unfold. At a given DTNU tree node, decisions such as scheduling a controllable timepoint at current time or waiting for a period of time develop children DTNU nodes for which these decisions are propagated to constraints. In this tree, only one timepoint can be scheduled per branch, rather than a set of timepoints, simply for compatibility reasons with the heuristic function used for guidance. The R-TDC controllability of a *leaf* DTNU node, *i.e.* a sub-DTNU for which all controllable timepoints have been scheduled and uncontrollable timepoints are assumed to have occurred in specific intervals, indicates whether or not this sub-DTNU has been solved at the end of the scheduling process. We also refer to the R-TDC controllability of a DTNU node in the search tree as its *truth attribute*. Lastly, the search logically combines R-TDC controllability of children nodes to determine R-TDC controllability for parent nodes.

### 6.2.2.1   Tree Structure

Let $\Gamma = \{A, U, C, L\}$ be a DTNU, where $A$ is the list of controllable timepoints, $U$ the list of uncontrollable timepoints, $C$ the list of constraints and $L$ the list of contingency links. The root node of the search tree is $\Gamma$. There are four different types of nodes in the tree and each node has a *truth* attribute (see §6.2.3.4) which is initialized to *unknown* and can be set to either *true* or *false*. The different types of tree nodes are listed below and shown in Figure 6.3.

**DTNU nodes.**   Any DTNU node other than the original problem $\Gamma$ corresponds to a sub-problem of $\Gamma$ at a given point in time $t$, for which some controllable timepoints may have already been scheduled in upper branches of the tree, some amount of time may have passed, and some uncontrollable timepoints are assumed to have occurred. A DTNU node is made of the same timepoints $A$ and $U$, constraints $C$ and contingency links $L$ as DTNU $\Gamma$. It also carries an schedule memory $S$ of what exact time, or during what time interval, timepoints were scheduled or occurred during previous decisions in the tree. Lastly, the node also keeps track of the activation time intervals of activated uncontrollable timepoints $B$. The schedule memory $S$ is used to create an updated list of constraints $C'$ resulting from the propagation of the scheduling time or occurrence time interval of timepoints in constraints $C$ as described in §6.2.3.5. A non-terminal DTNU node, *i.e.* a DTNU node for which all timepoints have not been scheduled, has exactly one child node: a *d-OR* node.

**OR nodes.**   When a choice can be made at time $t$, this transition control is represented by an *OR* node. We distinguish two types of such nodes, *d-OR* and *w-OR* . For *d-OR* nodes, the first type of choice available is which controllable timepoint $a_i$ to schedule. This leads to a DTNU node. The other type of choice is to wait a period of time (§6.2.3.1) which leads to a *WAIT* node. *w-OR* nodes can be used for reactive wait strategies, *i.e.* to stipulate that some controllable timepoints will be executed reactively in waits during strategy execution (§6.2.3.3). The parent of a *w-OR* node is therefore a *WAIT* node and its children are *AND* nodes, described below.

**WAIT nodes.** These nodes are used after a decision to wait a certain period of time $\Delta_t$. The parent of a *WAIT* node is a *d-OR* node. A *WAIT* node has exactly one child: a *w-OR* node, which has the purpose of exploring different reactive wait strategies. The uncertainty management related to uncontrollable timepoints is handled by *AND* nodes.

**AND nodes.** Such nodes are used after a wait decision is taken and a reactive wait strategy is decided, represented respectively by a *WAIT* and *w-OR* node. Each child node of the *AND* node is a DTNU node at time $t + \Delta_t$, $t$ being the time before the wait and $\Delta_t$ the wait duration. Each child node represents an outcome of how uncontrollable timepoints may unfold and is built from the set of *activated* uncontrollable timepoints (uncontrollable timepoints that have been triggered by the scheduling of their controllable timepoint) whose occurrence time interval overlaps the wait. If there are $l$ activated uncontrollable timepoints, then there are at most $2^l$ *AND* node children, representing each element of the power set of activated uncontrollable timepoints (§6.2.3.1).

Figure 6.3 illustrates how a sub-problem of $\Gamma$, referred to as $DTNU_{O,P,t}$, is developed. Here, $O \subset A$ is the set of controllable timepoints that have already been scheduled, $P \subset U$ the set of uncontrollable timepoints which have occurred, and $t$ the time. This root node transitions into a *d-OR* node. The *d-OR* node in turn is developed into several children nodes $DTNU_{O\cup\{a_i\},P,t}$ and a *WAIT* node. Each node $DTNU_{O\cup\{a_i\},P,t}$ corresponds to a sub-problem which is obtained from the scheduling of controllable timepoint $a_i$ at time $t$. The *WAIT* node refers to the process of waiting a given period of time, $\Delta_t$ in the figure, before making the next decision. The *WAIT* node leads directly to a *w-OR* node which lists different wait strategies $R_i$. If there are $l$ activated uncontrollable timepoints, there are $2^l$ subsets of uncontrollable timepoints $\Lambda_i$ that could occur. Each $AND_{R_j}$ node has one sub-problem DTNU for each $\Lambda_i$. Each sub-problem $DTNU_{O_i,P\cup\Lambda_i,t+\Delta_t}$ of the node $AND_{R_j}$ is a DTNU at time $t + \Delta_t$ for which all uncontrollable timepoints in $\Lambda_i$ are assumed to have happened during the wait period, *i.e.* in the time interval $[t, t + \Delta_t]$. Additionally, some controllable timepoints may have been reactively executed during the wait and may now be included in the set of executed controllable timepoints $O_i$. Otherwise, $O_i = O$.

Two types of leaf nodes exist in the tree. The first type is a node $DTNU_{A,U,t}$ for which all controllable timepoints $a_i \in A$ have been scheduled and all uncontrollable timepoints $u_i \in U$ have occurred. The second type is a node $DTNU_{A \setminus A', U, t}$ for which all uncontrollable timepoints $u_i \in U$ have occurred, but some controllable timepoints $a_i \in A'$ have not been scheduled. The constraint satisfiability test of the former type of leaf node is straightforward: scheduling times and occurrence time intervals of all timepoints are propagated to constraints in the same fashion as in §6.2.3.5. The leaf node's truth attribute is set to *true* if all constraints are satisfied, *false* otherwise. For the latter type, we propagate the occurrence time intervals of all uncontrollable timepoints as well as the scheduling time of all scheduled controllable timepoints in the same way, and obtain an updated set of constraint $C'$. This leaf node, $DTNU_{A \setminus A', U, t}$, is therefore characterized as $\{A', \varnothing, C', \varnothing\}$ and is a DTN. We add the constraints $a_i' \geq t, \forall a_i' \in A'$ and use a mixed integer linear programming solver (Cplex, 2009) to solve the DTN. If a solution is found, the time values for each $a_i' \in A'$ are stored and the leaf node's truth value is set to *true*. Otherwise, it is set to *false*. After a truth value is assigned to the leaf node, the truth propagation function defined in §6.2.3.4 is called to logically infer truth value properties for parent nodes. A *true* value reaching the root node of the tree means a R-TDC strategy has been found. The R-TDC strategy is a subtree of the search tree obtained by selecting recursively from the root, for each *d-OR* and *w-OR* nodes, the child with the *true* attribute, and for each *AND* node, all children nodes (which are necessarily *true*). A *false* attribute reaching the root means there is no existing R-TDC strategy. Lastly, the search algorithm explores the tree in a depth-first manner. At each *d-OR* , *w-OR* and *AND* node, children nodes are visited in the order they are created. Once a child node is selected, its entire subtree will be processed by the algorithm before the other children are explored. We give pseudocode for the tree search in Algorithm 4.

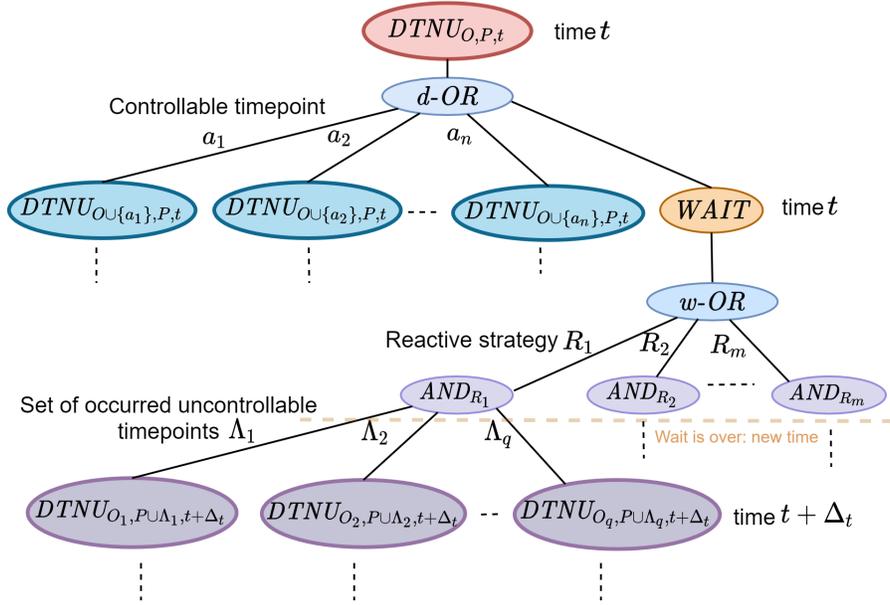

**Figure 6.3: Basic structure of the search tree describing how a DTNU node $DTNU_{O,P,t}$ is developed.** $DTNU_{O,P,t}$ (placed at the root of the tree) refers to a DTNU where $O$ is the set of controllable timepoints that have already been scheduled, $P$ the set of uncontrollable timepoints that have occurred, and $t$ the time. Each branch $a_i$ refers to a controllable timepoint $a_i$, $R_i$ to a reactive strategy during the wait, and $\Lambda_i$ to a combination of uncontrollable timepoints which can occur during the wait.

Note: In Algorithm 4, function $isLeaf(x)$ sets the truth value of $x$ to *true* and returns *true* if all constraints are satisfied. It sets the truth value to *false* and returns *true* if a constraint is violated. If no truth value can be inferred at this stage with the updated constraints, a second check is run to determine if all uncontrollable timepoints have occurred. If so, the corresponding DTN is solved, the truth value of $x$ is updated accordingly, and the function returns *true* . Otherwise, no logical outcome can be inferred for the current state of the constraints because there remains at least one uncontrollable timepoint and this function returns *false* .

---

**Algorithm 4** Tree Search

---

1: **function** explore(TreeNode $\psi$)
2:     **if** $parent(\psi).truth \neq unknown$ **then**
3:         **return**
4:     **if** isDTNU($\psi$) **then**
5:         updateConstraints($\psi$)                         ▷ [1]*
6:         **if** IsLeaf($\psi$) **then**                       ▷ [2]*
7:             propagateTruth($\psi$)
8:             **return**
9:         Create *d-OR* child $\psi'$
10:        explore($\psi'$)
11:     **if** isOR($\psi$) **then**
12:         Create list of all children $\Psi'$              ▷ [3]*
13:         **for** $\psi' \in \Psi'$ **do**
14:             explore($\psi'$)
15:     **if** isAND($\psi$) **then**
16:         Create list of all children $\Psi'$              ▷ [4]*
17:         **for** $\psi' \in \Psi'$ **do**
18:             explore($\psi'$)
19:     **if** isWAIT($\psi$) **then**
20:         create *w-OR* child $\psi'$
21:         explore ($\psi'$)
22: **function** main(DTNU $\gamma$)
23:     explore($\gamma$)
24:     **if** $\gamma.truth = True$ **then**
25:         **return** *True*
26:     **else**
27:         **return** *False*

---

[1]* updateConstraints($x$): Updates the constraints of DTNU node $x$.

[2]* isLeaf($x$): described in main text for space reasons.

[3]* If this is a *d-OR* node, the list $\Psi'$ contains all the children DTNU nodes resulting from either the decision of scheduling a controllable timepoint, or the *WAIT* node resulting from a wait if available. If this is a *w-OR* node, $\Psi'$ contains all $AND_{R_j}$ nodes, each of which possess a reactive wait strategy $R_j$

[4]* Here, the list $\Psi'$ contains all DTNUs resulting from all possible combinations $\Lambda_1, \Lambda_2, ..., \Lambda_q$ of uncontrollable timepoints which have the potential to occur during the current wait.

### 6.2.3 Tree Search Characteristics

#### 6.2.3.1 Wait Action

When a wait decision of duration $\Delta_t$ is taken at time $t$ for a DTNU node, two categories of uncontrollable timepoints are considered to account for all transitional possibilities:

- $Z = \{\zeta_1, \zeta_2, ..., \zeta_l\}$ is a set of timepoints that could either happen during the wait, or afterwards, *i.e.* the end of the activation time interval for each $\zeta_i$ is greater than $t + \Delta_t$.

- $H = \{\eta_1, \eta_2, ..., \eta_m\}$ is a set of timepoints that are certain to happen during the wait, *i.e.* the end of the activation time interval for each $\eta_i$ is less than or equal to $t + \Delta_t$.

There are $q = 2^l$ number of different possible combinations (empty set included) $Y_1, Y_2, ..., Y_q$ of elements taken from $Z$. For each combination $Y_i$, the set $\Lambda_i = H \cup Y_i$ is created. The union $\bigcup\limits_{i=1}^{q} \Lambda_i$ refers to all possible combinations of uncontrollable timepoints which can occur by $t + \Delta_t$. In Figure 6.3, for each *AND* node, the combination $\Lambda_i$ leads to a DTNU sub-problem $DTNU_{O_i, P \cup \Lambda_i, t + \Delta_t}$ for which the uncontrollable timepoints in $\Lambda_i$ are considered to have occurred between $t$ and $t + \Delta_t$ in the schedule memory $S$. In addition, any potential controllable timepoint $\phi$ planned to be instantly executed in a reactive wait strategy $R_i$ in response to an uncontrollable timepoint $u$ in $\Lambda_i$ will also be considered to have been executed between $t$ and $t + \Delta_t$ in $S$. The only exception is when checking constraint satisfiability for the conjunct $u - \phi \in [0, y]$ which required the reactive execution, for which we assume $\phi$ executed at the same time as $u$, thus the conjunct is considered satisfied.

#### 6.2.3.2 Wait Eligibility and Period

The way time is discretized holds direct implications on the search space explored and the capability of the algorithm to find strategies. Longer waits make the search space smaller, but carry the risk of missing key moments where a decision is needed. On the other hand, smaller waits can make the search space too large to explore. We explain when the wait action is eligible, and how its duration is computed.

**Wait Eligibility**   At least one of these two criteria has to be met for a *WAIT* node to be added as child of a *d-OR* node. **(1)** There is at least one activated uncontrollable timepoint for the parent DTNU node. **(2)** There is at least one conjunct of the form $v \in [x, y]$, where $v$ is a timepoint, in the constraints of the parent DTNU node. These criteria ensure that the search tree will not develop branches below *WAIT* nodes when waiting is not relevant, *i.e.* when a controllable timepoint necessarily needs to be scheduled. It also prevents the tree search from getting stuck in infinite *WAIT* loop cycles.

**Wait Period**   We define the wait duration $\Delta_t$ at a given *d-OR* node eligible for a wait dynamically by examining the updated constraint list $C'$ of the parent DTNU and the activation time intervals $B$ of its activated uncontrollable timepoints. Let $t$ be the current time for this DTNU node. The wait duration is defined by comparing $t$ to elements in $C'$ and $B$ to look for a minimum positive value defined by the following three rules. **(1)** For each activated time interval $u \in [x, y]$ in $B$, we select $x - t$ or $y - t$, whichever is smaller and positive, and we keep the smallest value $\delta_1$ found over all activated time intervals. **(2)** For each conjunct $v \in [x, y]$ in $C'$, where $v$ is a timepoint, we select $x - t$ or $y - t$, whichever is smaller and positive, and we keep the smallest value $\delta_2$ found over all conjuncts. **(3)** We determine timepoints which need to be scheduled ahead of time by chaining constraints together. Intuitively, when a conjunct $v \in [x, y]$ is in $C'$, it means $v$ has to be scheduled when $t \in [x, y]$ to satisfy this conjunct. However, $v$ could be linked to other timepoints by constraints which require them to happen before $v$. These timepoints could in turn be linked to yet other timepoints in the same way, and so on. The third rule consists in chaining backwards to identify potential timepoints which start this chain and potential time intervals in which they need to be scheduled. The following mechanism is used: for each conjunct $v \in [x, y]$ in $C'$ found in (2), we apply a recursive backward chain function to both $(v, x)$ and $(v, y)$. We detail here how it is applied to $(v, x)$, the process being the same for $(v, y)$. Conjuncts of the form $v - v' \in [x', y'], x' \geq 0$ in $C'$ are searched for. For each conjunct found, we add to a list two elements, $(v', x - x')$ and $(v', x - y')$. We select $x - x' - t$ or $x - y' - t$, whichever is smaller and positive, as potential minimum candidate. The backward chain function is called recursively on each element of the list, proceeding the same way. We keep the smallest candidate $\delta_3$. Figure 6.4 illustrates an application of this process. Finally, we set $\Delta_t = \min(\delta_1, \delta_2, \delta_3)$ as the wait duration. This duration is stored inside the *WAIT* node.

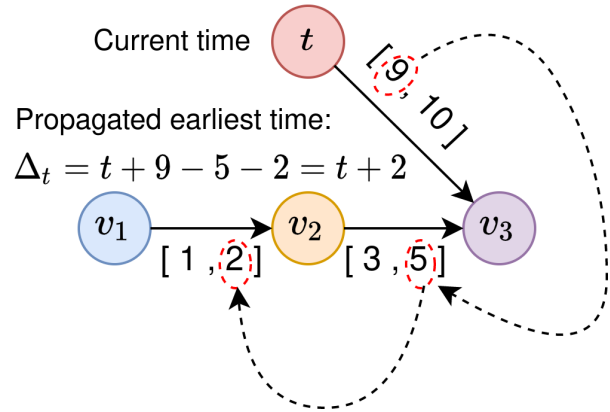

**Figure 6.4: Application of the $3^{rd}$ wait discretization rule to determine a wait duration.** Current time is $t$. Variables $v_1$, $v_2$ and $v_3$ are timepoints. Here, $v_2$ is constrained to be scheduled in the time interval $[1, 2]$ after $v_1$, $v_3$ in $[3, 5]$ after $v_2$ as well as in $[t + 9, t + 10]$. The rule suggests not to wait longer than 2 units of time at $t$: scheduling $v_1$ at $t + 2$, followed by scheduling $v_2$ at $t + 4$ opens a window of opportunity for $v_3$ to be scheduled at $t + 9$.

### 6.2.3.3 Reactive Executions during Waits

Scheduling of a controllable timepoint may be necessary in some situations at the exact same time as when an uncontrollable timepoint occurs to satisfy a constraint. Therefore, different reactive wait strategies are considered and listed as children of a *w-OR* node after a wait decision, before the start of the wait itself. We designate as a *conjunct* a constraint relationship of the form $v_i - v_j \in [x, y]$ or $v_i \in [x, y]$, where $v_i, v_j$ are timepoints and $x, y, \in \mathbb{R}$. We refer to a constraint where several conjuncts are linked by $\vee$ operators as a *disjunct*. If at any given DTNU node in the tree there is an activated uncontrollable timepoint $u$ with the potential to occur during the next wait and there is at least one unscheduled controllable timepoint $a$ such that a conjunct of the form $u - a \in [0, y], y \geq 0$ is present in the constraints, a reactive wait strategy is available that will set $a$ to be executed as soon as $u$ occurs in the strategy execution.

If there are $s$ controllable timepoints that may be set to be reactively executed, there are $2^s$ different reactive wait strategies $R_i$, each of which is embedded in an *AND* child of the *w-OR* node. Let $\Phi = \{\phi_1, \phi_2, ..., \phi_s\} \subset A$ be the complete set of unscheduled controllable timepoints for which there are conjunct clauses $u - \phi_i \in [0, y]$. We denote as $R_1, R_2, ..., R_m$ all possible combinations of elements taken from $\Phi$, including the empty set. The child node $AND_{R_i}$ of the *w-OR* node resulting from the combination $R_i$ has a reactive wait strategy for which all controllable timepoints in $R_i$ will be immediately executed during strategy execution at the moment $u$ occurs during the wait, if it occurs.

#### 6.2.3.4 Truth Value Propagation

In this section, we describe how truth attributes of nodes are related to each other. The truth attribute of a tree node represents its R-TDC controllability, and the relationships shared between nodes make it possible to define sound strategies. When a leaf node is assigned a truth attribute $\beta$, the tree search is momentarily stopped and $\beta$ is propagated onto upper parent nodes. To this end, a parent node $\omega$ is selected recursively and we distinguish the following cases:

- The parent $\omega$ is a DTNU or *WAIT* node: $\omega$ is assigned $\beta$.

- The parent $\omega$ is a *d-OR* or *w-OR* node: If $\beta = true$, then $\omega$ is assigned *true* . If $\beta = false$ and all children nodes of $\omega$ have *false* attributes, $\omega$ is assigned *false* . Otherwise, the propagation stops.

- The parent $\omega$ is an *AND* node: If $\beta = false$, then $\omega$ is assigned *false* . If $\beta = true$ and all children nodes of $\omega$ have *true* attributes, $\omega$ is assigned *true* . Otherwise, the propagation stops.

After the propagation finishes, the tree search algorithm resumes where it was stopped. The pseudocode for the propagation algorithm is given in Algorithm 5.

---

**Algorithm 5** Truth Value Propagation

---

1: **function** propagateTruth(TreeNode $\psi$)
2:      $\omega \leftarrow$ parent($\psi$)                                            ▷ $^1*$
3:      **if** $\omega = null$ **then**
4:          **return**
5:      **if** isDTNU($\omega$) or isWAIT($\omega$) **then**                     ▷ $^2*$
6:          $\omega.truth \leftarrow \psi.truth$
7:          propagateTruth($\omega$)
8:      **else if** isOR($\omega$) **then**                               ▷ $^3*$
9:          **if** $\psi.truth = True$ **then**
10:             $\omega.truth \leftarrow True$
11:             propagateTruth($\omega$)
12:          **else**
13:             **if** $\forall \sigma_i, \sigma_i.truth = False$ **then**          ▷ $^4*$
14:                 $\omega.truth \leftarrow False$
15:                 propagateTruth($\omega$)
16:      **else if** isAND($\omega$) **then**                           ▷ $^5*$
17:          **if** $\psi.truth = False$ **then**
18:             $\omega.truth \leftarrow False$
19:             propagateTruth($\omega$)
20:          **else**
21:             **if** $\forall \sigma_i, \sigma_i.truth = True$ **then**           ▷ $^4*$
22:                 $\omega.truth \leftarrow True$
23:                 propagateTruth($\omega$)

---

$^1*$ parent($x$): Returns the parent node of $x$, *null* if none.
$^2*$ isDTNU($x$): Returns *True* if $x$ is a DTNU node, *False* otherwise; isWait($x$): Returns *True* if $x$ is a *WAIT* node, *False* otherwise.
$^3*$ isOR($x$): Returns *True* if $x$ is an *d-OR* or *w-OR* node, *False* otherwise.
$^4*$ $\sigma_i$: Child number $i$ of $\omega$. For a *d-OR* or *w-OR* node, in the case where $\psi$ is *false* but not all other children of $\omega$ are *false* the propagation stops. Likewise, for an *AND* node and in the case where $\psi$ is *true* but not all other children of $\omega$ are *true*, the propagation stops.
$^5*$ isAnd($x$): Returns *True* if $x$ is an *AND* node, *False* otherwise.

---

### 6.2.3.5   Constraint Propagation

Decisions taken in the tree define when controllable timepoints are scheduled and also bear consequences on the occurrence time of uncontrollable timepoints. We explain here how these decisions are propagated into constraints, as well as the concept of '*tight bound*'. Let $C'$ be the list of updated constraints for a DTNU node $\psi$ for which the parent node is $\omega$. We distinguish two cases. Either $\omega$ is a *d-OR* node and $\psi$ results from the scheduling of a controllable timepoint $a_i$, or $\omega$ is an *AND* node and $\psi$ results from a wait of $\Delta_t$ time units. In the first case, let $t$ be the scheduling time of $a_i$. The updated list $C'$ is built from the constraints of the parent DTNU of $\psi$ in the tree. If a conjunct contains $a_i$ and is of the form $a_i \in [x, y]$, this conjunct is replaced with *true* if $t \in [x, y]$, *false* otherwise. If the conjunct is of the form $v_j - a_i \in [x, y]$, we replace the conjunct with $v_j \in [t + x, t + y]$. The other possibility is that $\psi$ results from a wait of $\Delta_t$ time units at time $t$, with a reactive wait strategy $R$. In this case, the new time is $t + \Delta_t$ for $\psi$. As a result of the wait, some uncontrollable timepoints $u_i \in \Lambda$ are assumed to have occurred, and some controllable timepoints $a_i \in A_R$ may be executed reactively during the wait. Let $v_i \in \Lambda \cup A_R$ be these timepoints occurring during the wait. The occurrence time of these timepoints is assumed to be in $[t, t + \Delta_t]$. For uncontrollable timepoints $u_i' \in \Lambda' \subset \Lambda$ for which the activation time ends at $t + \Delta_{t_i}' < t + \Delta_t$, and potential controllable timepoints $a_i'$ instantly reacting to these uncontrollable timepoints, the execution time is further reduced and considered to be in $[t, t + \Delta_{t_i}']$. We define a concept of *tight bound* to update constraints which restricts time intervals in order to account for all possible values $v_i$ can take between $t$ and $t + \Delta_t$. For all conjuncts $v_j - v_i \in [x, y]$, we replace the conjunct with $v_j \in [t + \Delta_t + x, t + y]$. Intuitively, this means that since $v_i$ can happen at the latest at $t + \Delta_t$, $v_j$ can not be allowed to happen before $t + \Delta_t + x$. Likewise, since $v_i$ can happen at the earliest at $t$, $v_j$ can not be allowed to happen after $t + y$. Finally, if $t + \Delta_t + x > t + y$, the conjunct is replaced with *false* . Also, the process can be applied recursively in the event that $v_j$ is also a timepoint that occurred during the wait, in which case the conjunct would be replaced by *true* or *false*. In any case, any conjunct obtained of the form $a_j \in [x', y']$ is replaced with *false* if $t + \Delta_t > y'$. Finally, if all conjuncts inside a disjunct are set to *false* by this process, the constraint is violated and the DTNU is no longer satisfiable.

### 6.2.3.6 Optimization Rules

The following rules are added to make branch cuts when possible.

**Constraint Check.** When a DTNU node is explored and the updated list of constraints $C'$ is built according to §6.2.3.5, if a disjunct is found to be *false* , $C'$ will no longer be satisfiable. All the subtree which can be developed from the DTNU will only have leaf nodes for which this is the case as well. The search algorithm will not develop this subtree.

***Symmetrical Subtrees.*** Some situations can lead to the development of the exact same subtrees. A trivial example, for a given DTNU node at a time $t$, is the order in which a given combination of controllable timepoints $a_1, a_2, ..., a_k$ is taken before taking a wait decision. Regardless of what order these timepoints are explored in the tree before moving to a *WAIT* node, they will be considered scheduled at time $t$. When taking a wait decision, it is thus checked that all preceding controllable timepoints scheduled before the previous wait are a combination of timepoints that has not been tested yet.

***Truth Checks.*** Before exploring a new node for which the truth attribute is set to *unknown*, the truth attribute of the parent node is also checked. The node is only developed if the parent node's truth attribute is set to *unknown*. In this manner, when children of a tree node are being explored (depth-first) and the exploration of a child node leads to the assignment of a truth value to the tree node, the remaining unexplored children can be left unexplored.

### 6.2.4 Learning-based Heuristic

We present our learning model and explain how it provides tree search heuristic guidance. Our learning architecture is the MPNN architecture from (Gilmer et al., 2017). It uses message passing rules allowing neural networks to process graph-structured inputs where both vertices and edges possess features. This architecture was originally designed for node classification in quantum chemistry and achieved state-of-the-art results on a molecular property prediction benchmark. We refer the reader to §2.3 and Equations 2.12, 2.13 and 2.14 for more details. Here, we define a way of converting DTNUs into graph data. Then, we process the graph data with a fixed MPNN architecture and use the output to guide the tree search.

Let $\Gamma = \{A, U, C, L\}$ be a DTNU. We explain how we turn $\Gamma$ into a graph $\mathcal{G} = (\mathcal{K}, \mathcal{E})$. First, we convert all time values from absolute to relative with the assumption the current time for $\Gamma$ is $t = 0$. We search all converted time intervals $[x_i, y_i]$ in $C$ and $L$ for the highest interval bound value $d_{max}$, *i.e.* the farthest point in time. We proceed to normalize every time value in $C$ and $L$ by dividing them by $d_{max}$. As a result, every time value becomes a real number between 0 and 1. Next, we convert each controllable timepoint $a \in A$ and uncontrollable timepoint $u \in U$ into graph nodes with corresponding *controllable* or *uncontrollable* node features. The time constraints in $C$ and contingency links in $L$ are expressed as edges between nodes with 10 different edge distance classes $(0 : [0, 0.1], 1 : [0.1, 0.2], ..., 9 : [0.9, 1])$. We also use additional edge features to account for edge types (constraint, disjunction, contingency link, direction sign for lower and upper bounds). Moreover, intermediary nodes are used with a distinct node feature in order to map possible disjunctions in constraints and contingency links. We add a *WAIT* node with a distinct node feature which implicitly designates the act of waiting a period of time.

The graph conversion of DTNU $\gamma$ is characterized by three elements: the matrix of all node features $X_\kappa$, the adjacency matrix of the graph $X_\epsilon$ and the matrix of all edge features $X_\rho$. These features are processed by a fixed number of consecutive message passing layers from (Gilmer et al., 2017) which make the MPNN. Each layer takes an input graph, consists of a phase during which messages are passed between nodes, and returns the same graph with new node features. Edge features remain the same. The overall process for a layer is the following. For each node $\kappa_i$ in the input graph, a *feature aggregation* phase is applied and creates new features for $\kappa_i$ from current features of neighboring nodes and edges. In detail, for each neighbor node $\kappa_j$, a small neural network (termed multi-layer perceptron, or MLP) takes as input the features of the edge connecting $\kappa_i$ and $\kappa_j$ and returns a matrix which is then multiplied by the features of $\kappa_j$ to obtain a feature vector. The sum of these vectors for the entire neighborhood defines the new features for $\kappa_i$. The output of the message passing layer consists of the graph updated with the new node features. In each message passing layer, the same MLP is used to process every node, so it can be applied to input graphs of any size, *i.e.* the MPNN architecture can take as input DTNUs of any size as long as the same set of input features per node is used (which is the case for any DTNU). Moreover, each message passing layer uses a different MLP architecture and can thus be trained to learn a different feature aggregation scheme.

Let $f$ be the mathematical function for our MPNN and $\theta$ its set of parameters. Our function $f$ stacks 5 message passing layers coupled with the ReLU$(\cdot) = \max(0, \cdot)$ piece-wise activation function after each layer, except the last one. The sigmoid function $\sigma(\cdot) = \frac{1}{1+\exp(-\cdot)}$ is used after the last layer to obtain a list of probabilities $\pi$ over all nodes in $\mathcal{G} : f_\theta(X_\kappa, X_\epsilon, X_\rho) = \pi$. The probability of each node $\kappa$ in $\pi$ corresponds to the likelihood of transitioning into a R-TDC DTNU from the original DTNU $\Gamma$ by taking the action corresponding to $\kappa$. If $\kappa$ represents a controllable timepoint $a$ in $\Gamma$, its corresponding probability in $\pi$ is the likelihood of the sub-DTNU resulting from the scheduling of $a$ at current time being R-TDC. If $\kappa$ represents a *WAIT* decision, its probability refers to the likelihood of the *WAIT* node having a *true* attribute, *i.e.* the likelihood of all children DTNUs resulting from the wait being R-TDC (with the wait duration rules set in §6.2.3.2). We call these two types of nodes *active* nodes. Otherwise, if $\kappa$ is another type of node, its probability is not relevant to the problem and ignored. Our MPNN is trained on DTNUs generated and solved in §6.2.5 only on active nodes by minimizing the multi-class binary cross-entropy loss:

$$\frac{1}{m} \sum_{i=1}^{m} \sum_{j=1}^{q} -Y_{ij} \log(f_\theta(X_i)_j) - (1 - Y_{ij}) \log(1 - f_\theta(X_i)_j)$$

Here $X_i = (X_{i_\kappa}, X_{i_\epsilon}, X_{i_\rho})$ is DTNU number $i$ among a training set of $m$ examples, $Y_{ij}$ is the R-TDC controllability (1 or 0) of active node number $j$ for DTNU number $i$.

Lastly, the MPNN heuristic is used in the following way in the tree search. Once a *d-OR* node is reached, its parent DTNU node is converted into a graph and the MPNN $f$ is called upon the corresponding graph elements $X_\kappa, X_\epsilon, X_\rho$. Active nodes in output probabilities $\pi$ are then ordered by highest values first, and the tree search visits the corresponding children tree nodes in the suggested order, preferring children with higher likelihood of being R-TDC first.

### 6.2.4.1 Learning and Implementation Details

We choose Pytorch for learning purposes. We use message passing layers that take as input a graph where nodes and edges possess features and return the graph with new node features. We detail pseudocode of a message passing layer applied to a graph $\mathcal{G} = (\mathcal{K}, \mathcal{E})$ in Algorithm 6.

Our MPNN architecture is made of 5 graph convolutional layers from (Gilmer et al., 2017). Each layer has a residual skip connection to the preceding layer (He et al., 2016), 32 abstract node features and a different two-layer MLP (multi-layer perceptron) which has 128 neurons in its hidden layer. In addition, we use batch normalization after each graph layer and apply the ReLU$(\cdot) = \max(0, \cdot)$ activation function. We add a dropout regularization layer with a *keep rate* 0.9 before the output layer to reduce overfitting. The input of the MPNN is the graph conversion of a DTNU. Figure 6.5 illustrates an example of graph conversion. We use 10 edge distance classes: $0 : [0, 0.1)$, $1 : [0.1, 0.2)$, ..., $9 : [0.9, 1]$. Training is done with the *adagrad* optimizer (Duchi, Hazan, and Singer, 2011) and an initial learning rate $10^{-4}$ on a dataset comprised of $30K$ instances generated as described in §6.2.5. We split the data into a training set comprised of $25K$ instances and a cross-validation set comprised of $5K$ instances on which we achieve 84% accuracy. We use a laptop with the following specifications for experiments: $9^{th}$ gen. Intel Core i7, 16GB RAM and nvidia GTX 1660 Ti.

---

**Algorithm 6** Message Passing Layer

---

1: **function** MsgPass(Graph $\langle (\mathcal{K}, \mathcal{E}), (H_\kappa, X_\epsilon, X_\rho) \rangle$)                                        ▷ [1]*
2:     $H'_\kappa(\cdot, \cdot) \leftarrow 0$ // *Initialize new node features matrix*
3:     **for all** $\kappa_i \in \mathcal{K}$ **do**
4:         $h'_i \leftarrow 0$ // *Initialize new features for* $\kappa_i$
5:         **for all** $\kappa_j \in \mathcal{K}$ **do**
6:             **if** $X_\epsilon(\kappa_i, \kappa_j) = 1$ **then**
7:                 $\alpha \leftarrow X_\rho(\kappa_i, \kappa_j)$
8:                 $h \leftarrow H_\kappa(\kappa_j, \cdot)$
9:                 $h'_i \leftarrow h'_i + MLP(\alpha)h$                                        ▷ [2]*
10:        $H'_\kappa(\kappa_i, \cdot) \leftarrow h'_i$ // *Assign new features for* $\kappa_i$
11:    **return** $\langle (\mathcal{K}, \mathcal{E}), (H'_\kappa, X_\epsilon, X_\rho) \rangle$

---

[1]* $H_\kappa(\kappa_i, \cdot)$ returns a vector of current features for node $\kappa_i$; $X_\epsilon(\kappa_i, \kappa_j)$ returns 1 if $(\kappa_i, \kappa_j) \in \mathcal{E}$, 0 otherwise; $X_\rho(\kappa_i, \kappa_j)$ returns a vector of current features for edge $(\kappa_i, \kappa_j)$.
[2]* MLP represents a multi-layer perceptron mapping input edge features to a matrix of dimension num-output-node-features x num-input-node-features. Moreover, $h$ is of dimension num-input-node-features x 1. The matrix multiplication therefore results in a vector of size num-output-node-features.

---

### 6.2.5 Randomized Simulations for Heuristic Training

We leverage a learning-based heuristic to guide the tree search. A key component in learning-based methods is the annotated training data. We generate such data in automatic manner by using a DTNU generator to create random DTNU problems and solving them with a modified version of the tree search. We store results and use them for training the MPNN. We detail now our data generation strategy.

We create DTNUs with a number of controllable timepoints ranging from 10 to 20 and uncontrollable timepoints ranging from 1 to 3. The generation process is the following. For interval bounds of constraint conjuncts or contingency links, we randomly generate real numbers within $[0, 100]$. We restrict the number of conjuncts inside a disjunct to 5 at most. A random number $n_1 \in [10, 20]$ of controllable timepoints and $n_2 \in [1, 3]$ of uncontrollable timepoints are selected. Each uncontrollable timepoint is randomly linked to a different controllable timepoint with a contingency link. Next, we iterate over the list of timepoints, and for each timepoint $v_i$ not appearing in constraints or contingency links, we add in the constraints a disjunct for which at least one conjunct constrains $v_i$. The type of conjunct is selected randomly from either a *distance* conjunct $v_i - v_j \in [x, y]$ or a *bounded* conjunct $v_i \in [x, y]$ (50% probabilities). On the other hand, if $v_i$ was already present in the constraints or

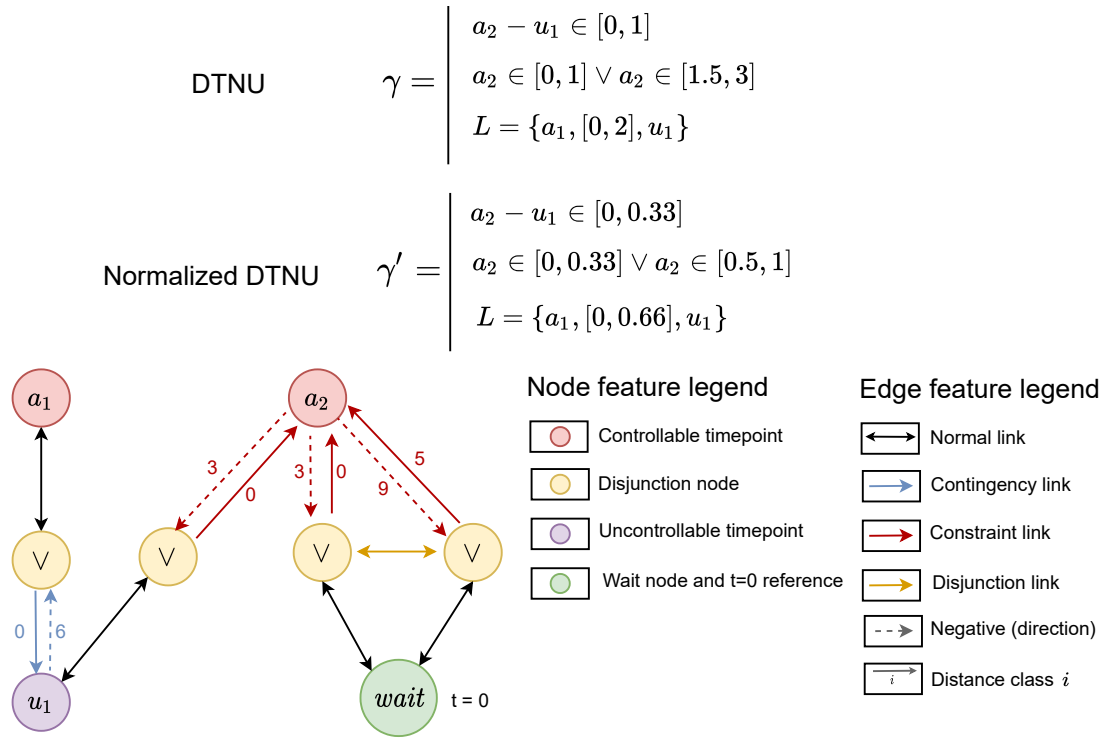

$$\text{DTNU} \qquad \gamma = \begin{vmatrix} a_2 - u_1 \in [0, 1] \\ a_2 \in [0, 1] \lor a_2 \in [1.5, 3] \\ L = \{a_1, [0, 2], u_1\} \end{vmatrix}$$

$$\text{Normalized DTNU} \qquad \gamma' = \begin{vmatrix} a_2 - u_1 \in [0, 0.33] \\ a_2 \in [0, 0.33] \lor a_2 \in [0.5, 1] \\ L = \{a_1, [0, 0.66], u_1\} \end{vmatrix}$$

**Figure 6.5: Conversion of a DTNU $\gamma$ into a graph.** $\gamma'$ is the normalized DTNU. Edge distances are expressed as distance classes. To distinguish between lower and upper bounds in intervals, we introduce an additional *negative directional sign* feature.

contingency links, we add a disjunct constraining $v_i$ with only a 20% probability.

In order to solve these DTNUs, we modify the tree search as follows. For a DTNU $\Gamma$, the first *d-OR* child node is developed as well as its children $\psi_1, \psi_2, ..., \psi_n \in \Psi$. The modified tree search explores each $\psi_i$ multiple times ($\nu$ times at most), each time with a timeout of $\tau$ seconds. We set $\nu = 25$ and $\tau = 3$. For each exploration of $\psi_i$, children nodes of any *d-OR* node encountered in the corresponding subtree are explored randomly each time. If $\psi_i$ is proved to be either R-TDC or non-R-TDC during an exploration, the next explorations of the same child $\psi_i$ are called off and the truth attribute $\beta_i$ of $\psi_i$ is updated accordingly. The active node number $k$, corresponding to the decision leading to $\psi_i$ from DTNU $\Gamma$'s *d-OR* node, is updated with the same value, *i.e.* $Y_k = \beta_i$ (1 for *true*, 0 for *false*). If every exploration times out, $\psi_i$ is assumed non-R-TDC and $Y_k$ is set to *false*. Once each $\psi_i$ has been explored, the pair $\langle G(\Gamma), (Y_1, Y_2, ..., Y_n) \rangle$ is stored in the training set, where $G(\Gamma)$ is the graph conversion of $\Gamma$ described in §6.2.4. Data related to solved sub-DTNUs of $\Gamma$ are not stored in the training set as it was found to cause bias issues and overall decrease generalization in MPNN predictions.

The assumption of non-R-TDC controllability for children nodes for which all explorations time out is acceptable in the sense that the heuristic used is not admissible and does not need to be. The output of the MPNN is a probability for each child node of the *d-OR* node, creating a preferential order of visit by highest probabilities first. Even in the event the suggested order first recommends visiting children nodes which will be found to be non-R-TDC, the algorithm will continue to explore the remaining children nodes until one is found to be R-TDC.

### 6.2.6  Experiments

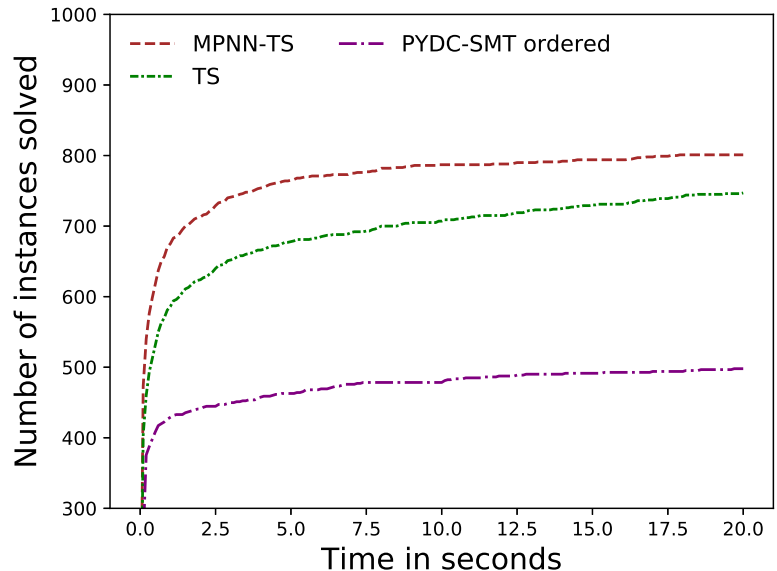

**Figure 6.6: Experiments on (Cimatti, Micheli, and Roveri, 2016)'s benchmark.** The X-axis represents the allocated time in seconds and the Y-axis the number of instances in the benchmark each solver can solve within the corresponding allocated time. Timeout is set to 20 seconds per instance.

We evaluate experimentally the efficiency of the tree search approach and the effect of the MPNN's guidance. We also compare them with a DC solver from (Cimatti, Micheli, and Roveri, 2016) on a same computer. R-TDC is a subset of DC and more restrictive: non-R-TDC controllability does not imply non-DC controllability. A R-TDC solver can thus be expected to offer better performance than a DC one while potentially being unable to find a strategy when a DC algorithm would. In this section, we refer to the tree search algorithm as TS, the tree search algorithm guided by the trained MPNN up to the $15^{th}$ (respectively $X^{th}$) $d$-OR node depth-wise in the tree as MPNN-TS (respectively MPNN-TS-X) and the most efficient DC solver from (Cimatti, Micheli, and Roveri, 2016) as PYDC-SMT ordered.

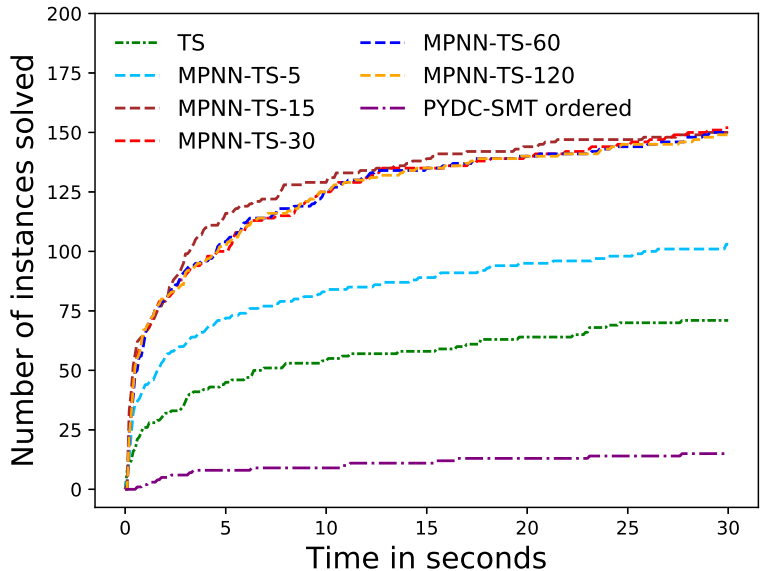

**Figure 6.7: Experiments on benchmark $B_1$.** Axes are the same as in Figure 6.6. Timeout is set to 30 seconds per instance.

First, we use the benchmark in the experiments of (Cimatti, Micheli, and Roveri, 2016) from which we remove DTNs and STNs. We compare TS, MPNN-TS and PYDC-SMT on the resulting benchmark which is comprised of 290 DTNUs and 1042 STNUs. Here, Limiting the maximum depth use of the MPNN to 15 offers a good trade off between guidance gain and cost of calling the heuristic. Results are given in Figure 6.6.

We observe that TS solves roughly 50% more problem instances than PYDC-SMT within the allocated time (20 seconds). In addition, TS solves 56% of all instances while the remaining ones time out. Among solved instances, a strategy is found for 89% and the remaining 11% are proved non-R-TDC. On the other hand, PYDC-SMT solves 37% of all instances. A strategy is found for 85% of PYDC-SMT's solved instances while the remaining 15% are proved non-DC. Finally, out of all instances PYDC-SMT solves, TS solves 97% accurately with the same conclusion, *i.e.* R-TDC when DC and non-R-TDC when non-DC. The use of the heuristic leads to an additional +6% problems solved within the allocated time. We argue this small increase is essentially due to the fact that most problems solved in the benchmark are small-sized problems with few timepoints which are solved quickly. Despite this fact, the heuristic still provides performance boost on a benchmark generated with another DTNU generator, suggesting the bias introduced by our DTNU generator remains limited and the MPNN is able to generalize to DTNUs created with a different approach.

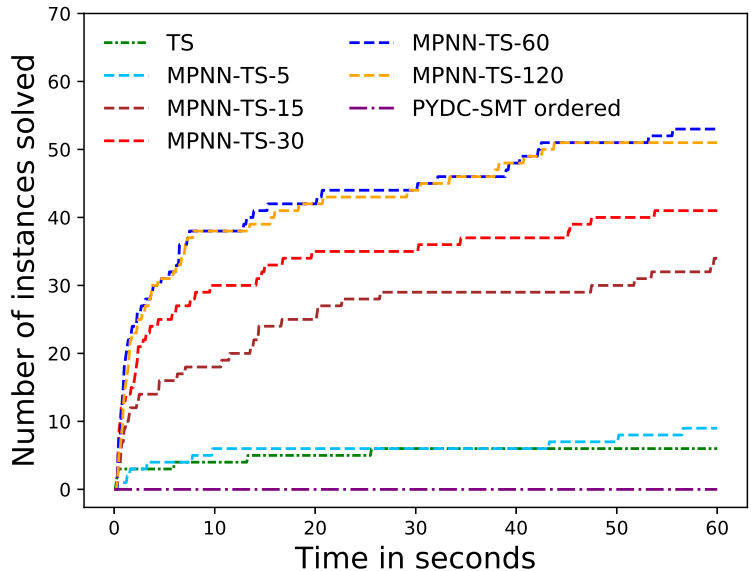

**Figure 6.8: Experiments on benchmark $B_2$.** Axes are the same as in Figure 6.6. Timeout is set to 60 seconds per instance.

For further evaluation of the heuristic, we create new benchmarks using the DTNU generator from §6.2.5 with varying number of timepoints. These benchmarks have fewer quick to solve DTNUs and harder ones instead. Each benchmark contains 500 randomly generated DTNUs which have 1 to 3 uncontrollable timepoints. Moreover, each DTNU has 10 to 20 controllable timepoints in the first benchmark $B_1$, 20 to 25 in the second benchmark $B_2$ and 25 to 30 in the last benchmark $B_3$. Each disjunct in the constraints of any DTNU contains up to 5 conjuncts. Experiments on $B_1$, $B_2$ and $B_3$ are respectively shown in Figure 6.7, 6.8 and 6.9. We note that for all three benchmarks no solver ever proves non-R-TDC or non-DC controllability before timing out due to the larger size of these problems.

PYDC-SMT performs poorly on $B_1$ and cannot solve any instance on $B_2$ and $B_3$. TS underperforms on $B_2$ and only solves 2 instances on $B_3$. However, we see a significantly higher gain from the use of the MPNN, varying with the maximum depth use. At best depth use, the gain is +91% instances solved for $B_1$, +980% for $B_2$ and +1150% for $B_3$. The more timepoints instances have, the more worthwhile heuristic guidance appears to be. Indeed, the optimal maximum depth use of the MPNN in the tree increases with the problem size: 15 for $B_1$, 60 for $B_2$ and 120 for $B_3$. We argue this is due to the fact that more timepoints results in a wider search tree overall, including in deeper sections where heuristic use was not necessarily worth its cost for smaller problems. Furthermore, the MPNN is trained on randomly generated DTNUs which have 10 to 20 controllable timepoints. The promising gains shown by experiments on $B_2$ and $B_3$ suggest generalization of the MPNN to bigger problems than it is trained on.

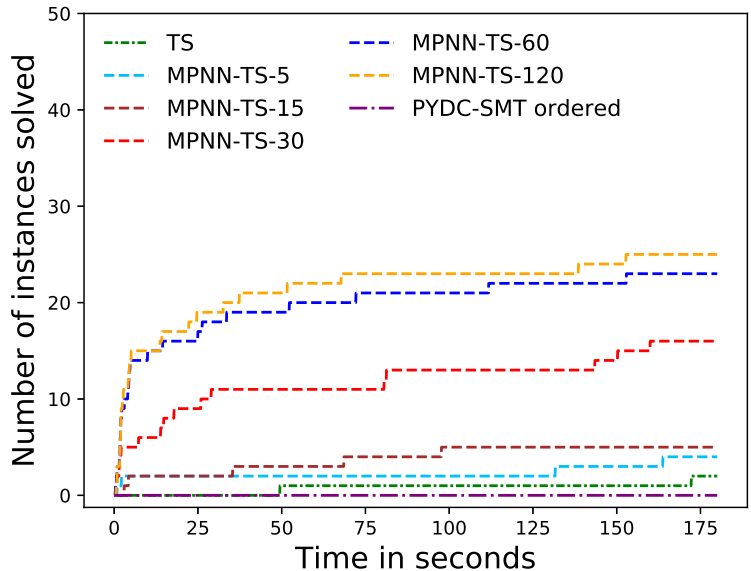

**Figure 6.9: Experiments on benchmark $B_3$.** Axes are the same as in Figure 6.6. Timeout is set to 180 seconds.

The proposed tree search approach presents a good trade off between search completeness and effectiveness: almost all examples solved by PYDC-SMT from (Cimatti, Micheli, and Roveri, 2016)'s benchmark are solved with the same conclusion, and many more which could not be solved are. Moreover, the R-TDC approach scales up better to problems with more timepoints, and the tree structure allows the use of learning-based heuristics. Although these heuristics are not key to solving problems of big scales, our experiments suggest they can still provide a high increase in efficiency.

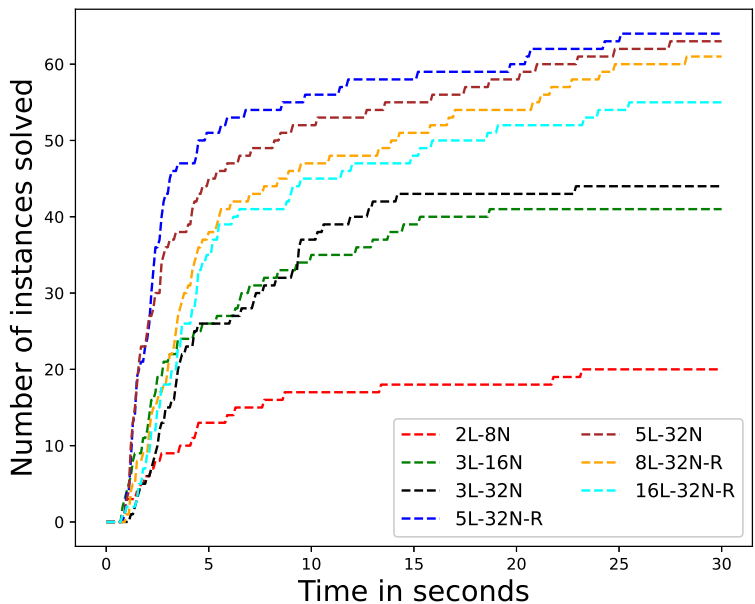

**Figure 6.10: Comparison of different MPNN architectures on performance.** The notation XL-YN refers to an MPNN with X layers and Y abstract node features per layer. The "-R" tag refers to the presence of residual layers. Experiments are done on a DTNU benchmark containing 400 instances with 20 to 25 controllable timepoints and up to 3 uncontrollable timepoints per DTNU. Timeout is set to 30 seconds per DTNU instance.

### 6.2.6.1 MPNN Architecture Comparison

We study the impact of the design choices of the MPNN architecture on performance. To this end we compare different architectures of MPNN by varying depth and width (number of abstract node features per layer) and train them on the training set created in §6.2.5. We also assess the added value of residual skip connections to preceding layers. We create a benchmark of 400 DTNU instances, each of which has 20 to 25 controllable timepoints and up to 3 uncontrollable timepoints (similarly to $B_2$). We solve them using the tree search guided by each of these MPNN architectures. We limit the use of the MPNN architectures to a maximal depth of 50 (*d-OR* node-wise). Results are shown in Figure 6.10. We note the smallest network is too small to learn efficiently and performs poorly. Three-layer networks perform better. Wider networks perform slightly better for the same depth, black network 32 vs. green network 16. Overall, medium-depth networks of 5 layers work best. Residual connections lead to slight but steady gains. Interestingly, deeper networks (8+ layers) display lower scores compared to more shallower variants (5 layers), suggesting depth performance saturation. The quantity of training data can however be a limiting factor: we assume the optimal architecture to be actually deeper.

## 6.3  Application to Perroquet Maneuver Scenario

We use MPNN-TS to solve scenarios from the third problem. We found that MPNN-TS solved such scenarios under 1 second if planning for less than 20 maneuvers ahead (10 controllable + 10 uncontrollable timepoints, *i.e.* 20 timepoint DTNUs), 2.5 seconds for 40, 6 seconds for 60. These results are amply satisfying for perroquet maneuvers. We decide to plan 40 maneuvers ahead maximum to avoid computation time of over 3 seconds in online scenarios. We note MPNN-TS performs notably better on these DTNUs than DTNUs of the same size from the experiment section in § 6.2.6. This is explained by two reasons: constraints of these maneuver scenarios are considerably easier, and their uncontrollable timepoints can only be activated one at a time (due to the characteristics of the perroquet maneuver), limiting the number of children nodes for *AND* nodes to only 2 maximum.

The following listing shows the strategy found by MPNN-TS on the scenario from Figure 1.6. Computation time is 0.24s. We selected the following values: $\delta = 10s$, $\Delta_1 = 25s$, $\Delta_2 = 65s$, $D_{max} = 40s$, $H_{max} = 40s$ and we set $t = 0$ at the beginning of the scenario for simplicity. In the listing, the strategy found is presented in a tree structure. Indices 0, 1 and 2 in the listing respectively refer to controllable timepoints $a_1$, $a_2$ and $a_3$. Indices 3, 4 and 5 in the listing refer respectively to uncontrollable timepoints $u_1$, $u_2$ and $u_3$. In the proposed R-TDC strategy, MPNN-TS figured out that the only way to overcome the $\Delta_2 - \Delta_1 = 40s$ exposure gap is to have the AUGV finish its first maneuver exactly at $t = \Delta_1 = 25s$ (*i.e.* it schedules $a_1$ at $t = 15s$ and the associated maneuver ends at $t = 25s$ given maneuver duration is $\delta = 10s$). In this way, even if the PV moves as soon as it can (*i.e.* $u_1$ gets executed at $t = 25s$), since the AUGV can wait at most 40 seconds after observing the start of the PV's maneuver before it has to move again, it can wait at most until $t = 65s = \Delta_2$, which is just enough not to violate the constraints requiring the AUGV to be still during the exposure interval.

**Listing 6.1:** R-TDC strategy computed by MPNN-TS on the scenario in Figure 1.6.

```
1  Strategy found
2  Compute time: 0.24 s
3
4  Wait 5 units at current time t = 0.00 with reactive strategy: {},
5  If these points occurred: []
```

```
 6   Wait 10 units at current time t = 5.00 with reactive strategy: {},
 7   If these points occurred: []
 8   Schedule [0] at current time t = 15.00,
 9   Wait 10 units at current time t = 15.00 with reactive strategy: {},
10    If these points occurred: []
11     Wait 20 units at current time t = 25.00 with reactive strategy: {},
12     If these points occurred: []
13     Wait 10 units at current time t = 45.00 with reactive strategy: {},
14      If these points occurred: [3]
15       Wait 10 units at current time t = 55.00 with reactive strategy: {},
16        If these points occurred: []
17         Wait 20 units at current time t = 65.00 with reactive strategy: {},
18          If these points occurred: []
19          Schedule [1] at current time t = 85.00,
20          Wait 10 units at current time t = 85.00 with reactive strategy: {},
21           If these points occurred: []
22            Wait 20 units at current time t = 95.00 with reactive strategy: {},
23            If these points occurred: []
24             Wait 10 units at current time t = 115.00 with reactive strategy: {},
25              If these points occurred: [4]
26               Schedule 2 at given time: [Decimal('155'), Decimal('155')]
27               Problem solved
28             If these points occurred: [4]
29              Schedule 2 at given time: [Decimal('135'), Decimal('135')]
30              Problem solved
31      If these points occurred: [3]
32       Wait 10 units at current time t = 45.00 with reactive strategy: {},
33        If these points occurred: []
34         Wait 10 units at current time t = 55.00 with reactive strategy: {},
35          If these points occurred: []
36          Schedule [1] at current time t = 65.00,
37          Wait 10 units at current time t = 65.00 with reactive strategy: {},
38           If these points occurred: []
39            Wait 20 units at current time t = 75.00 with reactive strategy: {},
40            If these points occurred: []
41             Wait 10 units at current time t = 95.00 with reactive strategy: {},
42              If these points occurred: [4]
43               Schedule 2 at given time: [Decimal('135'), Decimal('135')]
44               Problem solved
45             If these points occurred: [4]
46              Schedule 2 at given time: [Decimal('115'), Decimal('115')]
47              Problem solved
```

## 6.4   Conclusion

To solve the third problem, we introduced a new type of controllability for reactive scheduling (TDC & R-TDC) and a tree search approach for solving DTNUs in R-TDC. Strategies are built by discretizing time and exploring different decisions which can be taken at different key points, as well as anticipating how uncontrollable timepoints can unfold. We defined constraint propagation rules which ensure soundness of strategies found. We showed that the tree search approach is able to solve DTNUs in R-TDC more efficiently than the state-of-the-art DC solver can in DC, with almost always the same conclusion, despite R-TDC's lack of completeness compared to DC. Lastly, we created MPNN-TS, a solver which combines the tree search with a heuristic function based on MPNNs for guidance. The MPNN is trained with a self-supervised strategy and enables steady improvements of the tree search on harder DTNU problems, notably on DTNUs of bigger size than those used for training the MPNN. A significant advantage of MPNN-TS is that the same trained MPNN is applicable to graphs (DTNUs) of any size, *i.e.* our MPNN-TS framework is ready to be used in the online phase for the third problem in any scenario.

# Chapter 7

# Conclusion

First, we summarize all contributions presented in the thesis. Next, we conclude by exploring possible further works.

## 7.1   Contributions

The objective of this thesis consisted in investigating benefits machine learning algorithms could bring for planning and scheduling purposes, especially in applications involving Autonomous Unmanned Ground Vehicles (AUGV) and Autonomous Ground Vehicles (AGV). We studied different scenarios listed in 3 main problems. In the first problem, we defined an objective of minimal distance itinerary computation for an AUGV with mandatory pass-by node constraints. The second problem was similar to the first, but we added an additional criterion to the total itinerary distance called autonomous feasibility, which was unknown. This criterion needed to be maximized while the distance needed to be minimized. The third problem featured an AUGV performing synchronized maneuvers with a manned Partner Vehicle (PV) among a military convoy. In this last problem, the AUGV was required to calculate reactive strategies allowing it to synchronize its maneuvers with the PV in a way which guarantees the PV's safety from environmental dangers.

As we studied the first problem, we took note that we were aiming to perform path-planning tasks in a graph environment. While we originally intended to study how Convolutional Neural Networks (CNN) could be of use for the aforementioned problems, we turned our attention to a more recent variant of CNNs adapted to graphs, namely Graph Neural Networks (GNN). These networks seemed ideal because of the graph environment we were working in. Furthermore, their usefulness for combinatorial search problems also led us to use them for the third problem which had no direct graph representation, but for which we expressed the scheduling problem as a search problem. To summarize, we led the following research in this thesis:

1. We studied the first problem and proposed a mathematical representation for it. We proposed a vectorial encoding of problem instances, as well as a GNN architecture for processing and obtaining information of interest for search purposes. We introduced a Constraint Programming (CP) approach to solve instances optimally, and a self-supervised approach based on CP to train the GNN architecture. We showed in experiments that after training, GNN-derived information allowed the CP solver to perform far better than the baseline. Next, we proposed another approach based on branch & bound tree search to solve problem instances optimally. We introduced a planning domain and transition rules to solve the problem with planning algorithms such as A*. We modified the GNN architecture to make it suited for tree search, and introduced a self-supervised learning approach based on an algorithm derived from A* to train the GNN. The framework combining the GNN and the tree search exhibited significantly accelerated performance versus the baseline in experiments. It also outperformed by a large margin different other solvers, notably A* with a handcrafted heuristic tailored to the problem in question. Finally, we introduced an approach based on local search to solve complex, bigger instances approximately. In this approach, we used a GNN to provide an initialization point from which to begin local search. We presented a reinforcement learning approach to train the GNN. Experiments showed the GNN allowed local search to converge significantly faster to a solution which was better than the baseline more than 9 times out of 10.

2. We studied the second problem and formalized it mathematically as a multi-criteria optimization problem with constraints. We overcame the issue of not knowing one of the two criteria, the autonomous feasibility, by using a high-fidelity simulator offline. In simulations, we used a Multi-Layer Perceptron (MLP) to learn this unknown criterion using terrain and weather information. Next, we modified the CP solver used in the first problem in order to handle multi-criteria optimization with pareto-optimality guarantees. In experiments, we showed that the framework which uses MLP predictions on autonomous feasibility and the multi-criteria CP solver was able to find solution itineraries with limited increase in total distance cost compared to the single-criterion baseline, but with much higher autonomous feasibility.

3. We studied the third problem and expressed it mathematically, *i.e.* as a Disjunctive Temporal Network with Uncertainty (DTNU). We investigated the problem of reactive scheduling for DTNUs as a search problem and introduced a new form of controllability, Time-based Dynamic Controllablity (TDC & R-TDC). We presented a tree search approach capable of searching for R-TDC strategies. We defined a way to convert DTNUs into graphs, and a GNN architecture for processing those graphs in order to get relevant search information. The GNN was trained with a self-supervised training scheme which uses a randomized version of the tree search. The GNN was then paired with the tree search to provide search guidance. We carried out experiments in which the tree search significantly outperformed the state-of-the art solver. Additionally, we showed that GNN guidance allows the tree search to solve more complex DTNUs: up to 11 times more complex instances were solved within the same time budget compared to the tree search baseline.

To conclude, we successfully leveraged machine learning algorithms for planning and scheduling purposes in this thesis. Especially, recent advances in graph representation learning with GNNs proved particularly efficient for our applications.

## 7.2 Future Works

We propose the following ideas for future works:

1. Potential future work for problems 1 and 2 would include more types of constraints in addition to mandatory nodes to visit (such as nodes to exclude for instance or a maximum allowed number of crossing per edge). The proposed GNN architecture would need to include these additional types of constraints. While this is feasible from a technical point of view, each new constraint type results in additional features on nodes and potentially also on edges, making the search space possibly much wider. This would likely result in more difficult training of the GNN. Further works could investigate how the addition of other constraint types negatively affects the performance of the GNN, and how much more training data is needed to cope with it.

2. More future works for problem 1 would include the generalization to different graphs of a single GNN architecture. While this is entirely feasible from a GNN architecture point of view, the fact that each graph potentially uses very different real numbers for each edge weight makes it hard for a GNN to generalize to different graphs here. A potential solution could be to create a fixed number of distance classes, as we proceeded in our solving method for problem 3, and convert each edge weight to its corresponding class before using the GNN. However, while this way of proceeding seemed to perform relatively well for scheduling problems, there is no guarantee it will work as well for other types of problems such as path-planning. Further work may require investigating a more suited approach in order to avoid loss of information when converting those real numbers to classes.

3. A potential future work for problem 3 would be to try a form of 'incremental' training. This would consist in first training a GNN capable of dealing with DTNUs which have up to, for example, 30 timepoints, exactly as we proceeded in our described approach. Then, this GNN, which we call GNN-30, would be used only on DTNUs which have less than 30 timepoints. In the next training phase, we would generate DTNUs with 30 to 60 timepoints: when the tree search is being carried out on these DTNUs, we would use GNN-30 once we reach search nodes (sub-DTNUs) which have less than 30 timepoints. For search nodes which have more than 30 timepoints, we would use the randomized exploration as explained in our proposed method. With the new data obtained on 30 to 60 timepoint DTNUs, we would train a second GNN, namely GNN-60. This GNN would be used in the tree search for nodes which have between 30 to 60 timepoints. This cycle can then be repeated to train yet another GNN, and so on. This incremental training scheme is more complex to perform overall. Potential work would investigate how it could increase the performance of the overall framework on complex DTNUs.

# Appendix A

# Dijkstra's Algorithm

We give the pseudocode for Dijkstra's algorithm in the next page. The first function *Dijkstra* is used to find, for each node in the graph, an optimal predecessor in an optimal path coming from the source node $v_i$, until the optimal predecessor is found for the destination node $v_j$. A second function *RebuildPath* is used to build the shortest path from $v_i$ to $v_j$.

---

**Algorithm 7** Dijkstra's algorithm

---

1: **function** *Dijkstra*(Graph $\mathcal{G} = (\mathcal{V}, \mathcal{E})$, source $v_i$, destination $v_j$)
2:     Vertex Set $Q \leftarrow \varnothing$
3:     **for all** $v \in \mathcal{V}$ **do**
4:         $dist(v) \leftarrow \infty$                                       $\triangleright^1*$
5:         $prev(v) \leftarrow UNDEFINED$                    $\triangleright^2*$
6:         $Q \leftarrow Q \cup \{v\}$
7:     $dist(v_i) \leftarrow 0$
8:     **while** $Q \neq \varnothing$ **do**
9:         $u \leftarrow$ vertex in Q with minimal $dist(u)$
10:        $Q \leftarrow Q - \{u\}$
11:        **if** $u = v_j$ **then**
12:            return $dist()$, $prev()$
13:        **for all** $v \in \mathcal{V}$ s.t $(u, v) \in \mathcal{E}$ and $v \in Q$ **do**
14:            $alt \leftarrow dist(u) + length(u, v)$              $\triangleright^3*$
15:            **if** $alt < dist(v)$ **then**
16:                $dist(v) \leftarrow alt$
17:                $prev(v) \leftarrow u$
18:     Return $dist()$, $prev()$
19: **function** *RebuildPath* (Graph $\mathcal{G} = (\mathcal{V}, \mathcal{E})$, source $v_i$, destination $v_j$, $dist()$, $prev()$)
20:     $S \leftarrow [\,]$
21:     $u \leftarrow v_j$
22:     **if** $prev(u)$ is defined or $u = v_i$ **then**
23:         **while** $u$ is defined **do**
24:            insert $u$ at the beginning of $S$
25:            $u \leftarrow prev(u)$
26:     Return $S$

$^1*$ $dist(v)$ is a function that returns the minimum distance found from $v_i$ to $v$
$^2*$ $prev(v)$ is a function that returns the predecessor associated with the minimum distance found from $v_i$ to $v$
$^3*$ $length(v, v')$ is a function that returns the cost of the edge from $v$ to $v'$

---

# Appendix B

# 2-OPT Algorithm

We give the pseudocode for 2-OPT algorithm in the next page. The $Swap$ function is used inside the $2OPT$ function to swap nodes (symmetrically) for a given ordering of nodes only inside the range of two bounds.

**Algorithm 8** 2-OPT algorithm

1: // indices start at 0
2: **function** $Swap(ordering, i, k)$
3:     $l \leftarrow length(ordering)$
4:     $newOrdering \leftarrow []$
5:     **for** $j = 0, 1, ..., i - 1$ **do**
6:         $newOrdering[j] \leftarrow ordering[j]$   // copy in same order
7:     **for** $j = i, i + 1, ..., k$ **do**
8:         $newOrdering[j] \leftarrow ordering[k - (j - i)]$   // copy in reverse order
9:     **for** $j = k, k + 1, ..., l - 1$ **do**
10:         $newOrdering[j] \leftarrow ordering[j]$   // copy in same order
11:     **return** newOrdering
12: **function** $2OPT(ordering)$
13:     $l \leftarrow length(ordering)$
14:     $i \leftarrow 0$
15:     **while** $i \neq l - 1$ **do**
16:         $improve \leftarrow False$
17:         **for** $j = i + 1, i + 2, ..., l - 1$ **do**
18:             $newOrder \leftarrow Swap(ordering, i, j)$
19:             **if** $Cost(newOrder) < Cost(ordering)$ **then**             ▷ [1]*
20:                 $ordering \leftarrow newOrder$
21:                 $i \leftarrow 0$
22:                 $improve \leftarrow True$
23:                 **break** // leaves for loop
24:         **if** improve = false **then**
25:             $i \leftarrow i + 1$
26:     **return ordering**

[1]* $Cost(order)$ returns the total cost corresponding to the ordering of the nodes in *order*.

# Glossary

**Anytime planning** is a type of planning which returns valid plans quickly, and improves them in the remaining time.

**Autonomous feasibility** of an off-road path represents whether or not an autonomous vehicle can drive on the path without human assistance.

**Branch and bound tree search** is a tree exploration algorithm for combinatorial problems in which cuts are made when possible to avoid uninteresting parts of the tree.

**Constraint programming** is a technology used to find solutions to scheduling and combinatorial optimization problems.

**Contingency link** is a trigger which defines a time window when an uncontrollable timepoint variable can occur after a controllable one.

**Controllable timepoint** is a variable which the scheduling system can execute at any time $t \in \mathbb{R}$.

**Disjunctive temporal network with uncertainty** is a type of scheduling problem featuring controllable and uncontrollable timepoints, constraints with disjunctions, contingency links with disjunctions.

**Dynamic controllability** is a type of controllability for temporal problems with uncertainty in which online reactive scheduling is possible.

**Global search** is carried out in the entire search space. Given enough time and the right conditions, a global search algorithm will find an optimal solution.

**Graph convolutional network** is a type of graph neural network.

**Graph neural network** is a machine learning architecture which processes graph-structured inputs.

**Incremental planning** is a type of planning capable of repairing plans to cope with observed changes.

**Local search** is carried out locally. As opposed to global search, local search moves from solution to solution by making local changes, and can therefore remain stuck in local optimums.

**Message passing neural network** is a type of graph neural network.

**Multi-layer perceptron** is a type of neural network made of fully-connected layers.

**Offline phase** is a long phase where an autonomous vehicle is being prepared for a particular terrain.

**Online phase** is a phase where an autonomous vehicle is on a terrain and needs to compute either paths or maneuver strategies on the fly, as missions are defined.

**Pareto optimality** is a situation in multi-criteria optimization in which no criterion can be improved without worsening another one.

**Path-planning** is a specific form of planning in which plans are paths.

**Perroquet maneuver** is a type of military maneuver in which two vehicles are moving side by side, one at a time.

**Planning** consists in conceiving plans in order to achieve a desired goal.

**Self-supervised learning** is a method which automatically annotates unlabelled training examples, yielding a training set on which supervised learning is performed to train a machine learning model.

**Simple temporal network** is a type of scheduling problem featuring controllable timepoints and constraints.

**Simple temporal network with uncertainty** is a type of scheduling problem featuring controllable and uncontrollable timepoints, constraints, contingency links.

**Supervised learning** is a method used to train a machine learning model on a given set of training examples (training set).

**Training example** is a sample of data, *i.e.* a pair $(\mathbf{x}_i, \mathbf{y}_i)$, since data is labelled in the thesis.

**Training set** is a set of training examples used to train a machine learning model.

**Uncontrollable timepoint** is a variable which the scheduling system cannot execute. It occurs on its own sometime in a bounded time interval after a controllable timepoint has been executed.

# Acronyms

**AGV** Autonomous Ground Vehicle.
**AUGV** Autonomous Unmanned Ground Vehicle.

**B&B** Branch & Bound.

**CNN** Convolutional Neural Network.
**CP** Constraint Programming.

**DC** Dynamic Controllability.
**DP** Dynamic Programming.
**DTNU** Disjunctive Temporal Network with Uncertainty.

**GCN** Graph Convolutional Network.
**GNN** Graph Neural Network.

**MLP** Multi-Layer Perceptron.

**MPNN** Message Passing Neural Network.

**NN** Neural Network.

**PV** Partner Vehicle.

**RL** Reinforcement Learning.

**SC** Strong Controllability.
**STN** Simple Temporal Network.
**STNU** Simple Temporal Network with Uncertainty.

**TDC** Time-based Dynamic Controllability.

**WC** Weak Controllability.

# Index

## RÉSUMÉ

Cette thèse étudie l'apport que peut présenter l'apprentissage automatique pour des tâches de planification dans le cadre de véhicules autonomes en situation off-road. Nous nous intéressons plus particulièrement à des problèmes de planification d'itinéraire sous contraintes, ainsi que de calcul de stratégies d'exécution de manœuvres synchronisées avec d'autres véhicules. Nous présentons une série d'heuristiques basées sur de l'apprentissage ayant pour but d'aider des planificateurs d'itinéraire. Nous montrons que ces heuristiques permettent un gain significatif de performance pour des planificateurs optimaux. Nous montrons également dans le cas de la planification approximative un gain qui ne se limite pas à la performance uniquement mais s'étend également à la qualité de l'itinéraire trouvé, ce dernier étant presque toujours meilleur. Enfin, afin de calculer des stratégies d'exécution de manœuvres synchronisées, nous proposons une nouvelle forme de contrôlabilité dynamique de planification ainsi qu'un algorithme assisté par apprentissage. La technique proposée permet une nette amélioration sur des benchmarks connus dans cette forme de contrôlabilité vis à vis des travaux de l'état de l'art qui portent sur une forme de contrôlabilité analogue. Elle permet aussi de trouver des stratégies pour des problèmes de planification difficiles que les travaux précédents ne peuvent résoudre.

## ABSTRACT

This thesis explores the benefits machine learning algorithms can bring to online planning and scheduling for autonomous vehicles in off-road situations. Mainly, we focus on typical problems of interest which include computing itineraries that meet certain objectives, as well as computing scheduling strategies to execute synchronized maneuvers with other vehicles. We present a range of learning-based heuristics to assist different itinerary planners. We show that these heuristics allow a significant increase in performance for optimal planners. Furthermore, in the case of approximate planning, we show that not only does the running time decrease, the quality of the itinerary found also becomes almost always better. Finally, in order to synthesize strategies to execute synchronized maneuvers, we propose a novel type of scheduling controllability and a learning-assisted algorithm. The proposed framework achieves significant improvement on known benchmarks in this controllability type over the performance of state-of-the-art works in a related controllability type. Moreover, it is able to find strategies on complex scheduling problems for which previous works fail to do so.

## KEYWORDS

Artificial intelligence, Automated Planning, Scheduling under Uncertainty, Machine Learning, Reinforcement Learning, Graph Neural Networks.